\documentclass[10pt,journal,compsoc]{IEEEtran}

\usepackage{ifpdf}
\usepackage[nocompress]{cite}
\usepackage[pdftex]{graphicx}
\usepackage{amsmath}
\usepackage{amssymb}
\usepackage{algorithm}
\usepackage{algorithmicx}
\usepackage{algpseudocode}
\usepackage{booktabs}

\usepackage{makeidx}  % allows for indexgeneration
\usepackage{latexsym}
\usepackage{marginnote}
\usepackage{lipsum}

\usepackage{url}
\usepackage{multirow}
\usepackage{tabularx}
\usepackage{mathtools}
\usepackage{array, booktabs}
\usepackage{graphicx}
\usepackage{xcolor}         % colors
\usepackage{marginnote}
\usepackage[colorlinks,
            linkcolor=red,       %%修改此处为你想要的颜色
            anchorcolor=green,  %%修改此处为你想要的颜色
            citecolor=blue,        %%修改此处为你想要的颜色，例如修改blue为red
            ]{hyperref}
\usepackage{graphicx}  % 加载 graphicx 包，用于插入图片
\usepackage{subcaption} % 加载 subcaption 包，用于创建子图

\usepackage{multirow} % 表格
\usepackage{colortbl} % 表格
\usepackage{bm}
\usepackage{ulem}
\usepackage{xspace}

\newcommand{\mixup}{Mixup\xspace}

\newcounter{modification}[section]

\newcolumntype{C}[1]{>{\centering\arraybackslash}m{#1}} % for attention map

\begin{document}

\title{Towards Natural Machine Unlearning}

\author{Zhengbao He, Tao Li, Xinwen Cheng, Zhehao Huang, Xiaolin Huang,~\IEEEmembership{Senior Member,~IEEE}
\IEEEcompsocitemizethanks{
\IEEEcompsocthanksitem Corresponding author: Xiaolin Huang.
\IEEEcompsocthanksitem Z. He, T. Li, X. Cheng, Z. Huang, and  X. Huang are with the
Institute of Image Processing and Pattern Recognition, 
School of Automation and Intelligent Sensing, 
Shanghai Jiao Tong University, Shanghai 200240, P.R. China, and MoE Key Laboratory of System Control and Information Processing, Shanghai 200240, P.R. China.
E-mail: \{lstefanie, li.tao, xinwencheng, kinght\_h,  xiaolinhuang\}@sjtu.edu.cn
\IEEEcompsocthanksitem Code is available at \url{https://github.com/ZhengbaoHe/NatMU}.}}

\IEEEtitleabstractindextext{
\begin{abstract}
Machine unlearning (MU) aims to eliminate information  that has been learned from specific training data, namely forgetting data, from a pre-trained model. 
  Currently, the mainstream of  relabeling-based MU methods involves modifying  the forgetting data with incorrect labels and subsequently fine-tuning the model. 
  While learning such incorrect information can indeed remove knowledge, the process is quite unnatural as the unlearning process undesirably reinforces the incorrect information and leads to over-forgetting. 
  Towards more \textit{natural} machine unlearning, we inject correct information from the remaining data to the forgetting samples when changing their labels. 
  Through pairing these adjusted samples with their labels, the model  tends to use the injected correct information and naturally suppress the information meant to be forgotten. 
  Albeit straightforward, such a first step towards natural machine unlearning can significantly outperform current state-of-the-art approaches. In particular, our method substantially reduces the over-forgetting and leads to strong robustness across different unlearning tasks, making it a promising candidate for practical machine unlearning.
\end{abstract}

\begin{IEEEkeywords}
machine unlearning, natural unlearning, over-forgetting
\end{IEEEkeywords}}

% \maketitle

\maketitle

% is not selected <OR> if conference mode is selected - because compsoc
% conference papers position the abstract like regular (non-compsoc)
\IEEEdisplaynontitleabstractindextext

% \fi
\IEEEpeerreviewmaketitle

\ifCLASSOPTIONcompsoc
\IEEEraisesectionheading{\section{Introduction}\label{sec:introduction}}
\else
\section{Introduction}
\label{sec:introduction}

\fi

\IEEEPARstart{M}{odern} machine learning (ML) models are essential in various applications~\cite{DBLP:journals/pami/BadrinarayananK17, DBLP:journals/pami/CroitoruHIS23, DBLP:journals/pami/HongZLLLYYLGJPGBC24,DBLP:journals/pami/BaltrusaitisAM19}. However, their heavy reliance on extensive data for training raises significant privacy concerns. The General Data Protection Regulations (GDPR)~\cite{hoofnagle2019european} emphasizes individuals' rights to request the deletion of their private data, leading to a surge of interests in machine unlearning (MU). MU aims to remove the influence of specific data in the training set from a well-trained model. Recently, this field has garnered considerable attention not only for its contributions to privacy protection~\cite{GinartGVZ19, GuoGHM20, UllahM0RA21,goel2022towards} but also for its capacity to eliminate erroneous and sensitive data~\cite{ChundawatTMK23,assd, puma}.

Current popular MU methods are primarily based on optimization, achieving unlearning by fine-tuning the original model with manually crafted data with a proposed unlearning objective.  Among these, relabeling-based approaches play a key role, as they are easy to implement and provide stable training. 
For instance, Amnesiac~\cite{amnesiac} optimizes the model using randomly labeled forgetting samples  along with other remaining data.
BadTeacher~\cite{badteacher} relabels the forgetting samples with the predictions of a randomly initialized model as a  ``bad'' teacher. 
\begin{figure}[ht!]
\centering
\includegraphics[width=0.45\textwidth]{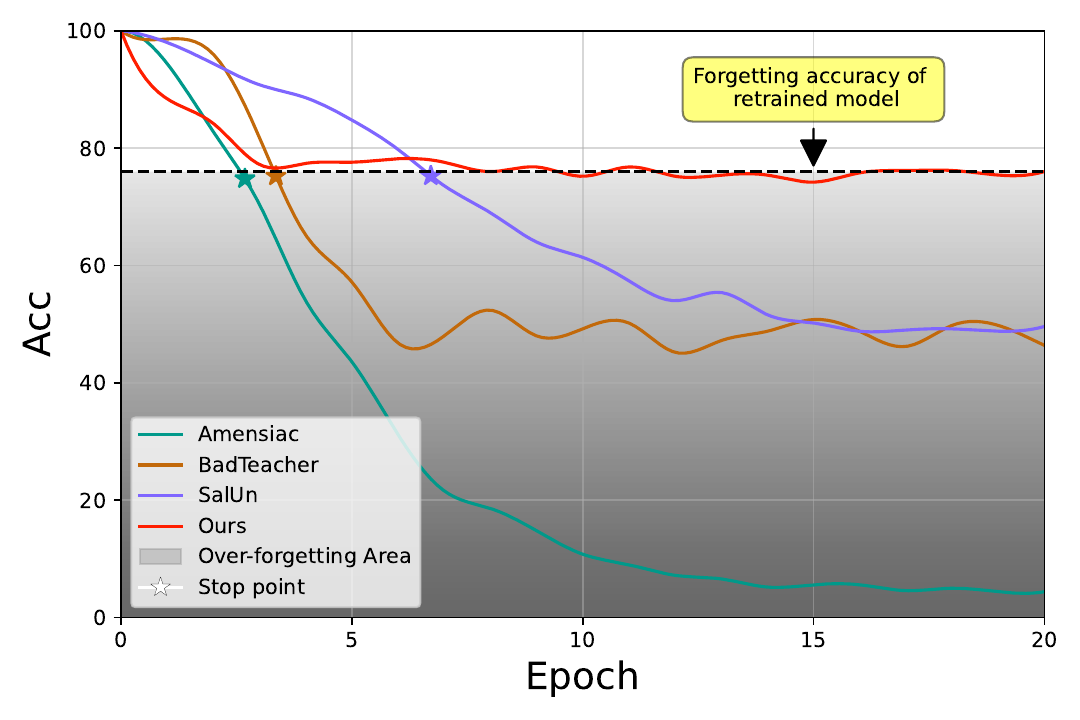}
\caption{Accuracy comparison of different methods on  
    forgetting samples. The experiments are taken on CIFAR-100 using ResNet-18 under 1\% sample-wise unlearning setting.
    % Accuracy on forgetting samples of different methods in random-subset unlearning on CIFAR-100 using ResNet18. 
    The dash line is the natural forgetting accuracy of retrained model, to which a smaller gap indicates better MU. The forgetting accuracy of other methods continuously decreases after crossing over the dash line, while ours can converge to that of the retrained model. Hence, to obtain a good MU performance, other methods may need to carefully stop training at a middle point, denoted as \text{\Large$\star$}.}
\label{fig:training-compare}
\end{figure}

The name of ``bad'' teacher hits the essence of the above methods: they create \textit{incorrect} information with the forgetting samples to fine-tune the model, compelling it to forget the correct information previously learned.
However, the incorrect information could be undesirably reinforced during the fine-tuning process.
An obvious observation, illustrated in Fig.~\ref{fig:training-compare}, is the so-called ``over-forgetting'', where after sufficient training, the accuracy of classifying the forgetting samples as their original labels (denoted as forgetting accuracy) is significantly lower than expected levels.
This occurs because the incorrect information in these wrongly relabeled instances\footnote{
In this paper, we consider a training point as a ``sample-label'' pair, also referred to as an ``instance''. Please see Sec.~\ref{subsection:preliminaries} for details.
} 
appears quite unnatural in the view of remaining data~\cite{memorylocated}. The conflicts between the relabeled instances and  remaining data cause the unlearned  model to  remember these forgetting samples even more firmly. 
This over-forgetting problem is further evidenced by substantial variations in the prediction entropy of the forgetting samples, which pose significant risks of privacy leakage and will be discussed in Sec.~\ref{subsection:revisiting}.
To avoid over-forgetting, one can restrict the mobility scale by identifying a parameter mask~\cite{salun} or by carefully selecting a stopping point. 
However, this requires meticulous hyperparameter tuning, which is impractical in real-world applications.

To address the problem of over-forgetting, we contend that the unlearning process should be \textit{natural}, minimizing the conflicts within the fine-tuning data.
The most natural MU process is to retrain the model from scratch with the training data excluding the forgetting ones, which is the golden standard for MU~\cite{DBLP:conf/uss/ThudiJSP22, unrolling-sgd, DBLP:journals/corr/abs-2308-07061}. 
This method ensures that all information used is correct, eliminating concerns about over-forgetting since the forgetting samples are excluded from the training process. 
However, retraining requires significant computational resources,  making it impractical in many cases,
especially in large-scale ML scenarios like CLIP~\cite{CLIP} or GPT~\cite{gpt}.
The challenge, therefore, lies in balancing the inclusion of forgetting samples to enhance unlearning efficiency and the exclusion of incorrect information to maintain the process's naturalness.

In this paper, we introduce a novel relabeling-based machine unlearning method, NatMU, which aims to facilitate a more natural unlearning by injecting correct information into the forgetting samples.
Specifically, such correct information is extracted from the remaining data and then injected into forgetting samples to create hybrid samples
using a modified \mixup~\cite{mixup} technique.
The hybrid samples are subsequently assigned categories consistent with the injected information. 
Each hybrid sample consists of two distinct types of information: one from the forgetting sample, and the other from the remaining sample. 
By learning to pair the hybrid sample with the reassigned label, the connection between injected information and corresponding label is reinforced.
This naturally suppresses model's response to the former type of information which is to be forgotten, thereby  achieving effective unlearning.
Since such reinforced connection inherently exists within the remaining set, NatMU achieves a more natural machine unlearning process.

NatMU takes the first step towards natural machine unlearning with the injection of correct information, which significantly reduces the  conflicts within the finetuning data.
The resulting model can maintain a natural generalization on forgetting samples, consistent with the predictions of retrained model. Moreover, NatMU prevents the output distribution shifts on forgetting samples, with similar behavior to the retrained model when facing membership inference attack (MIA). 
%As a result, NatMU can narrow the forgetting accuracy's gap with the retrained model from 39.44\% to 1.72\% when unlearning sub-class ``Vehicle2'' on CIFAR-20.
The natural property also eliminates the need to stop training at unstable points for good performance.
Consequently, NatMU demonstrates considerable robustness to various unlearning tasks, enabling the application of transferred hyperparameters across unknown settings.  
For instance, when unlearning the most 10\% difficult-to-learn samples,  NatMU maintains an average performance gap of only 2.75\% while other methods exhibit a gap over 10\%.

Our contributions can be summarized as follows: 

$\bullet$ We first point out the unnatural property of the previous relabeling-based MU methods, which leads to issues such as over-forgetting and impracticality.

$\bullet$ We propose an effective and efficient MU approach towards natural machine unlearning, named NatMU, which successfully addresses previous issues by injecting correct information into forgetting samples to  remove the information to be forgotten.

$\bullet$  Extensive experiments on various datasets and multiple machine unlearning scenarios demonstrate that NatMU significantly narrows the performance gap with the retrained model compared to previous approaches. Moreover, NatMU exhibits considerable robustness across unlearning scenarios.

\section{Related Work}

Machine unlearning aims to update a pretrained ML model to remove the influence of a subset of training set~\cite{first}.
The ML model can take various forms, such as K-means~\cite{GinartGVZ19}, linear classifier~\cite{GuoGHM20, puma}, or neural networks~\cite{UllahM0RA21, goel2022towards, ChundawatTMK23} as focused on in this paper.
Although retraining from scratch with the remaining data can achieve perfect MU, its significant demand for resources and time is unacceptable in the current deep learning context,
where DNNs with billions of parameters are commonly used in large-scale applications~\cite{CLIP,gpt}.
Therefore, numerous works have emerged that aim to approximate the retrained model, especially in the output space.

\textbf{Optimization-free unlearning} focuses on  removing the influence of the forgetting data by directly modifying model weights. Influence function is first introduced by \cite{KohL17}, and \cite{GuoGHM20} adopts it  for certified data removal in $L_2$-regularized linear models. The following work~\cite{DBLP:conf/nips/TannoPNL22,IzzoSCZ21,MehtaPSR22,DBLP:conf/ndss/WarneckePWR23} is dedicated to reducing the computational burdens introduced by  Hessian inversion. \cite{IzzoSCZ21} adopts infinitesimal jackknife with Newton methods to reduce the number of calculations while \cite{MehtaPSR22} reduces the overhead of one calculation by selecting only important parameters. 
FisherForgetting~\cite{fisherforgetting} introduces a weight scrubbing method by injecting noise to specific parameters to clean the information about the forgetting data with fisher information matrix \cite{fim}. SSD~\cite{ssd} and ASSD~\cite{assd} addresses its computational overhead and generalization  decrease through a fast but stringent parameter selection.
Besides the influence function-based methods, \cite{amnesiac} proposes to delete the gradient of mini-batches relevant to the forgetting data. \cite{unrolling-sgd} decomposes the SGD training process and performs a gradient ascent on the forgetting data. Although these methods can remove the influence of forgetting data, they may significantly impair the generalization capability of the MU model, especially in the context of modern deep learning.

\textbf{Optimization-based unlearning} re-optimizes the original model on a carefully crafted dataset with a proposed unlearning objective. \textbf{Relabeling-based} methods relabels training data and finetune the original model on them. Amnesiac~\cite{amnesiac} finetunes the model with randomly labeled forgetting samples along with unchanged remaining data. BadTeacher~\cite{badteacher}  relabels the remaining data with predictions of the original model and relabels the forgetting data with  predictions of a randomly initialized model, i.e., the ``bad'' teacher.
SalUn~\cite{salun} improves Amnesiac by constraining the model parameters' update with a weight saliency mask. 
$\delta\text{-targeted}^\lambda$~\cite{targeted} reassigns the top non-ground-truth class to forgetting samples as the forgetting label, and  proposes a strategy for dynamically re-weighting the cross-entropy loss to prevent overfitting on these relabeled data.  
LTU~\cite{ltu} introduces a meta-learning framework to shield the remaining data from the negative effects of learning randomly-labeled forgetting samples, and incorporates feedback from membership inference models as a form of forgetting guidance. 
These works aim to mitigate the negative impact of random labeling while realizing effective unlearning. However, these methods still face the over-forgetting problem, revealing the inherent flaws of the random labeling technique.
Rather than modifying only the labels, NatMU modifies the input samples to inject correct information, enabling a more natural and effective unlearning process. This shift from label-level to input-level modification distinguishes NatMU from prior approaches and contributes to both improved performance and robustness. 
\textbf{Relabeling-free} methods usually do not relabel the forgetting data~\cite{scrub, cf-k, sparse}. Executing gradient ascent on the forgetting data is a common technique to achieve unlearning, but it could lead to an unstable training. For example, SCRUB~\cite{scrub} also adopts the teacher-student framework like BadTeacher, but forces the unlearned model disobey the  original model, rather than obeying a random model.
Recently, some studies have explored the scenario where no remaining data are available, called ``zero-shot'' machine unlearning.
% \cite{} generates adversarial samples from forgetting samples and ensures that the predictions of the MU model on these samples match those of the original model, thereby maintaining model performance in the absence of remaining data. 
MU-Mis~\cite{mu-mis} reveals the link between sample's contribution to learning process and model's sensitivity to it, subsequently proposing  an algorithm by minimizing input sensitivity.
JiT~\cite{zero-shot-lips} and \cite{learningtounlearn} perturbs forgetting samples with random or adversarial perturbation, and ensures the predictions of perturbed versions match reference predictions to maintain model performance. However, these zero-shot unlearning methods cannot preserve a good model utility with effective unlearning, especially under sample-wise unlearning scenarios. 

\section{The Over-forgetting Issue}

\subsection{Preliminaries}\label{subsection:preliminaries}

Let $\mathcal{D} = \{(\boldsymbol{x}_i, y_i)\}_{i=1}^N$ be the training set containing inputs sample $\boldsymbol{x}_i \in \mathbb{R}^d$ with corresponding label $y_i \in \{1,2,...,K\}$. We denote a sample-label pair as an ``instance'' in this paper. Let forgetting set $\mathcal{D}^f \subset \mathcal{D}$ be the set of forgetting data and remaining set $\mathcal{D}^r = \mathcal{D} \setminus \mathcal{D}^f$ be the set of remaining data. 
The machine learning model can be represented as a parameterized function $f_\theta(\cdot) : \mathbb{R}^d \rightarrow  \mathbb{R}^K $ with parameter $\theta$. Let $\theta_o$ be the parameter of the original model, $\theta_u$ be the parameter of  the unlearned model and $\theta_r$ be the parameter of  the retrained model.
The objective of approximate MU is to 
\textbf
{narrow the gap  between the unlearned and retrained model} in output space. The gap could be measured by multiple metrics, such as forgetting accuracy and MIA ratio, which will be discussed in Section~\ref{setup}.

Mainstream machine unlearning work primarily focuses on three unlearning scenarios: 
(1)~\emph{full-class unlearning}, where $\mathcal{D}^f = \{ (\boldsymbol{x}_i, y_i) \in \mathcal{D}, y_i=k\}$ contains all samples with label $y_i=k$. 
(2)~\emph{sub-class unlearning}, where $\mathcal{D}^f \subset\{ (\boldsymbol{x}_i, y_i) \in \mathcal{D}, y_i=k\} $ contains 
a related subset of the samples labeled as $k$. For instance, this scenario includes unlearning the sub-class ``Shark'' within class ``Fish'' on CIFAR-20 dataset~\cite{cifar}. These two forgetting scenarios are also termed as \emph{class-wise unlearning}.
(3)~\emph{sample-wise unlearning}, where $\mathcal{D}^f$ is a random subset of $\mathcal{D}$. The size of $\mathcal{D}^f$ is determined by the forgetting ratio. For example,  1\%  forgetting ratio implies that 1\% of the samples from $\mathcal{D}$ are  selected to form $\mathcal{D}^f$. 
Among these unlearning scenarios, sample-wise unlearning presents the greatest challenge due to the disconnection  within $\mathcal{D}^f$ and the intricate intertwining between $\mathcal{D}^f$ and $\mathcal{D}^r$.

% \subsection{Current popular MU methods}
\subsection{Revisiting Relabeling-based MU Methods}\label{subsection:revisiting}
Relabeling-based MU algorithms 
%demonstrate exceptional performance across various scenarios, especially in random-subset unlearning. These methods 
require finetuning the original model on a carefully crafted finetuning dataset, typically comprising two parts: one built from the remaining data to maintain generalization ability and the other from the forgetting data to facilitate  unlearning. We name the latter part as \textbf{unlearning dataset}, and instances in the latter part as \textbf{unlearning instances}.
In Amnesiac~\cite{amnesiac}, the finetuning dataset consists of the original remaining data and forgetting samples with random incorrect labels: 
\begin{equation}
\begin{aligned}
    & D_\mathrm{AMN} = \mathcal{D}^r \cup \mathcal{D}^f_\mathrm{RL},  \\ 
    & \mathcal{D}^f_\mathrm{RL} = \{(\boldsymbol{x}_i, y_i^{\mathrm{rand}}), (\boldsymbol{x}_i, y_i) \in \mathcal{D}^f, y_i^{\mathrm{rand}} \neq y_i\}.
\end{aligned}
\end{equation}
BadTeacher~\cite{badteacher} utilizes the original model $f_{\theta_o}$ and a randomly initialized model $f_{\theta_\mathrm{bad}}$ to construct softly labeled finetuning dataset from $\mathcal{D}^r$ and $\mathcal{D}^f$ respectively:
\begin{equation}
\begin{aligned}
    & D_\mathrm{BT} = \mathcal{D}^r_{\mathrm{ori}} \cup \mathcal{D}^f_{\mathrm{bad}}, \\ 
    & \mathcal{D}^r_{\mathrm{ori}} = \{(\boldsymbol{x}_i,f_{\theta_o}(\boldsymbol{x}_i)), (\boldsymbol{x}_i, y_i) \in \mathcal{D}^r \},\\
    & \mathcal{D}^f_{\mathrm{bad}} = \{(\boldsymbol{x}_i,f_{\theta_\mathrm{bad}}(\boldsymbol{x}_i)), (\boldsymbol{x}_i, y_i) \in \mathcal{D}^f\}.
\end{aligned}
\end{equation}
SalUn~\cite{salun} improves Amnesiac by freezing certain model parameters according to a gradient-based weight saliency map. This comes from the intuition that some parameters have greater influence on forgetting samples than remaining sample. However, it still adopts the same loss function with Amnesiac and cannot  not fully address the problems of it.

To approximate the retrained model in the output space, the ideal approach would be to relabel the forgetting samples with their predictions on the retrained model as the ground-truth labels. However, in real-world unlearning scenarios, these ground-truth labels cannot be obtained.
Therefore, to ensure the unlearning process can effectively remove the learned knowledge, previous works assign incorrect labels to these samples as a secondary priority.
Learning on such modified data with incorrect labels can effectively remove the knowledge learned from them, since these modified instances cannot provide any useful information for classification~\cite{rethinking} and even introduce harmful knowledge.

As illustrated in Fig.~\ref{fig:training-compare}, these methods exhibit an ``over-forgetting'' issue, where the accuracy of classifying the forgetting samples as their original labels is significantly lower than that of the retrained model. The core challenge lies in that incorrect information, i.e., the forgetting samples with their incorrect labels, is undesirably reinforced during finetuning. This unlearning process is unnatural, as it forces the model to assimilate incorrect information, which actually conflicts with the remaining data. Consequently, this adversely impacts the model’s ability to generalize these forgetting samples, a capability that the retrained model retains. For instance, in sample-wise unlearning, most forgetting samples can still be correctly classified by the retrained model, even without being included in the training. Essentially, this unlearning approach modifies learned knowledge into incorrect knowledge rather than removing it. Although we can early stop the finetuning with meticulous hyperparameter tuning, such hyperparameters are not robust and cannot be generalized to  different sets of forgetting samples. This hyperparameter tuning is also impractical in real-world applications, as we cannot get the retrained model to determine where to stop the finetuning.

\begin{figure}[t]
\centering
\includegraphics[width=0.48\textwidth]{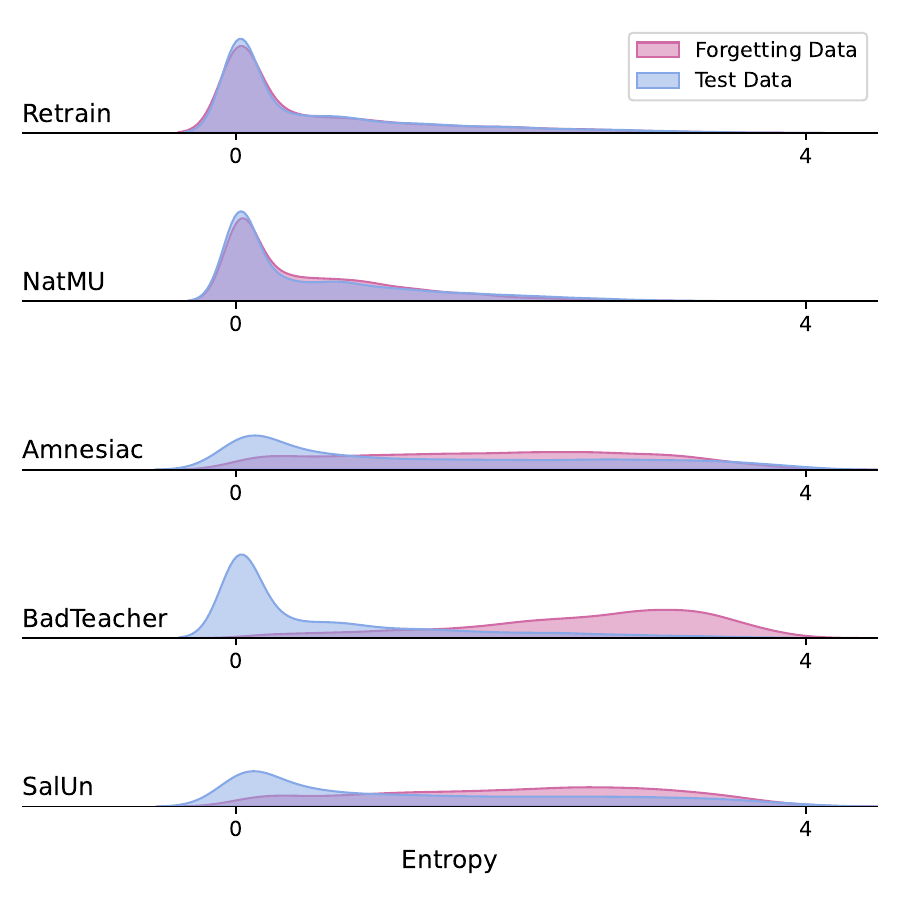}
\caption{Entropy distribution of different unlearned models' output on forgetting and test samples using kernel density estimation. The output distribution of NatMU is similar to Retrain on both forgetting samples and test samples. In contrast, other methods significantly change the entropy distribution of forgetting samples, thereby enabling the separation of forgetting samples from test samples and posing a serious risk of privacy leakage.  The experiments are conducted on CIFAR-100 using ResNet-18 under 10\% random-sample-wise unlearning setting. The number of unlearning epochs is set to 5 and the hyperparameters are selected following the setup in Section \ref{random-sample-wise-unlearning}. } 
\label{fig:entropy}
\end{figure}

Accuracy is only an index for over-forgetting. In fact, when incorrect information is injected, the output distribution will be largely changed.   Fig.~\ref{fig:entropy} demonstrates the entropy distributions of the test samples and forgetting samples. Ideally, the distribution for an unlearning model should be similar to the retrained model. However, the existing methods significantly alter the entropy distribution of the forgetting samples, even when hyper-parameters are  carefully tuned according to accuracy. This discrepancy allows an attacker to easily identify the forgetting samples, posing a serious risk of privacy leakage.

Therefore, it is crucial to realize a more natural machine unlearning while ensuring its efficiency. 
On one hand, the forgetting samples should be included to enhance unlearning efficiency. On the other hand, excluding incorrect information ensures a natural machine unlearning process. 
In this paper, we propose a novel method, named NatMU, towards natural machine unlearning. The core idea of NatMU is to inject information extracted from the remaining data into the forgetting samples to reduce the conflicts in each modified instance.

\section{Method}
\begin{figure*}[t!]
    \centering

    \begin{subfigure}[b]{0.72\textwidth}
        \resizebox{\textwidth}{!}{
            \setlength{\tabcolsep}{1pt}
            \begin{tabular}{
m{0.2\textwidth}<{\centering}  
m{0.2\textwidth}<{\centering}
m{0.2\textwidth}<{\centering}  
m{0.2\textwidth} <{\centering} 
m{0.2\textwidth}<{\centering} 
m{0.2\textwidth} <{\centering} 
}
    \toprule

    $\boldsymbol{x}^f$ 
    & $\boldsymbol{x}^f \circ \boldsymbol{m}_j$
    & $\boldsymbol{x}^r \circ (\boldsymbol{1}_d -\boldsymbol{m}_j)$
    & $\mathcal{T}_{\boldsymbol{m}_j}(\boldsymbol{x}^r, \boldsymbol{x}^r)$
    &  Before
    &  After \\ 
    \midrule

    % \multicolumn{6}{l}{\includegraphics[width=1.2\textwidth]{figs/total.png}}
    \includegraphics[width=0.2\textwidth]{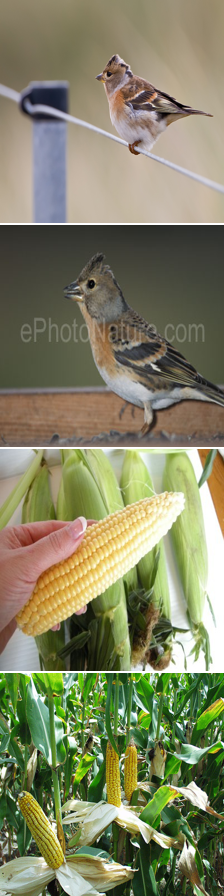}
    & \includegraphics[width=0.2\textwidth]{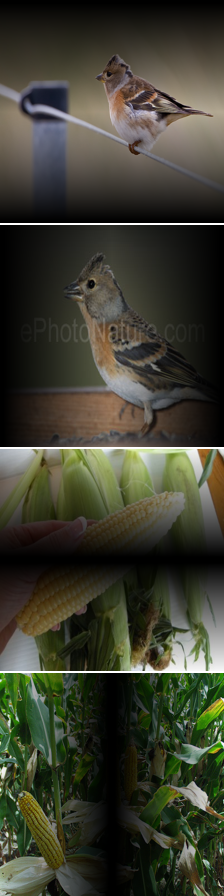}
    & \includegraphics[width=0.2\textwidth]{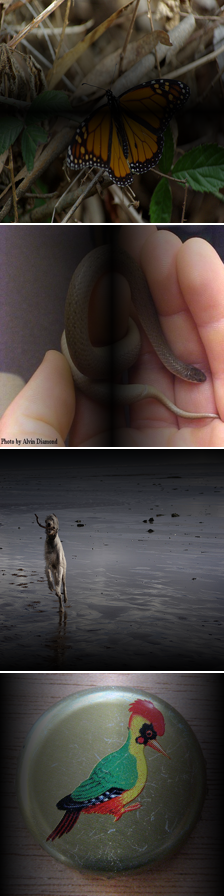}
    & \includegraphics[width=0.2\textwidth]{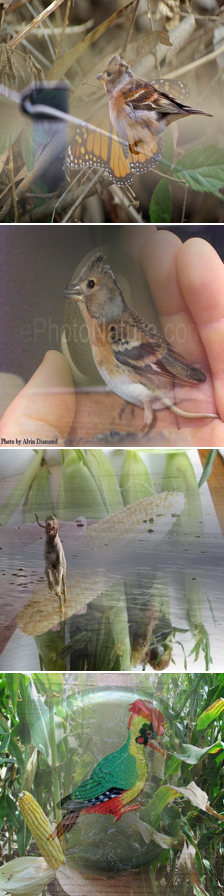}
    & \includegraphics[width=0.2\textwidth]{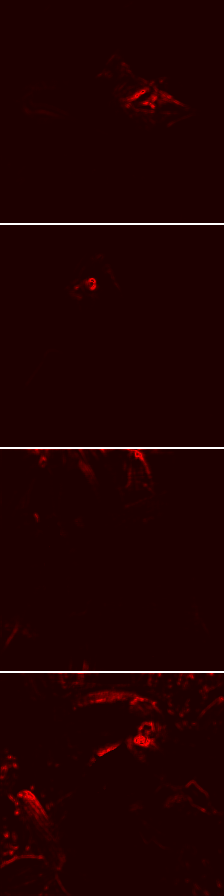}
    & \includegraphics[width=0.2\textwidth]{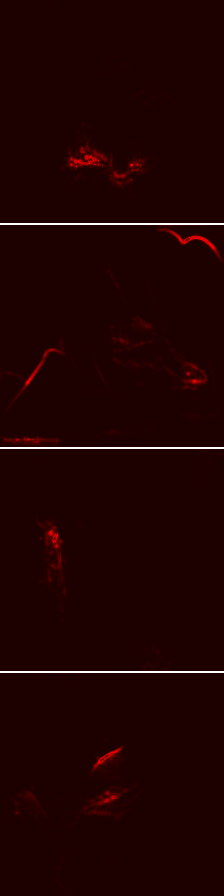}
    \\ 
    \bottomrule
    
\end{tabular}
            }  
        \caption{}
        \label{subfig:attetion}
    \end{subfigure}
    \hfill
    \begin{subfigure}[b]{0.25\textwidth}
        \centering
        \includegraphics[width=\textwidth]{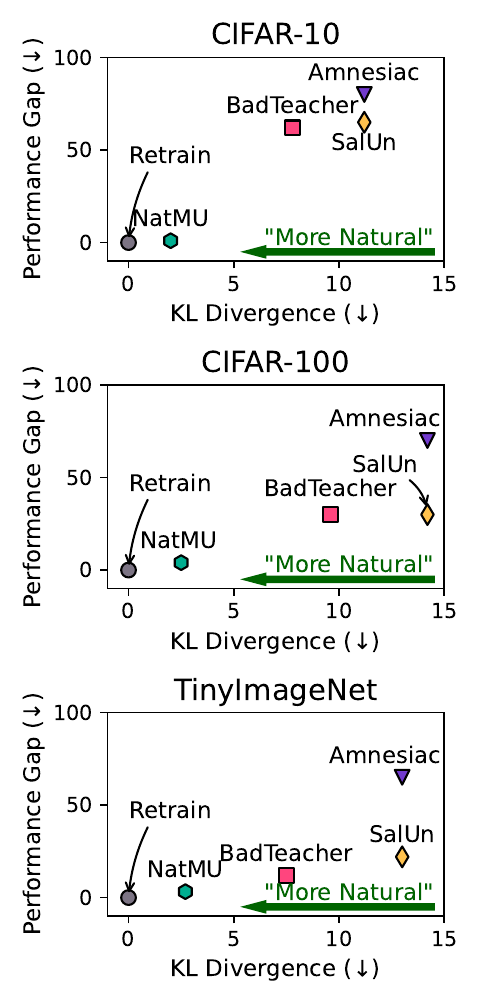}
        \caption{}
        \label{subfig:natural-rate}
    \end{subfigure}

    \caption{
    \textbf{(a)} Visualization of our unlearning instances and their attention maps before and after unlearning. The attention maps are calculated with LRP~\cite{lrp}. After unlearning, the attention is shifted to the remaining information.
    \textbf{(b)} Relationship between the gap of forgetting accuracy and KL divergence on different datasets. 
    A smaller KL divergence indicates a more natural MU process. NatMU's KL divergence is much smaller than other methods, i.e., more natural, thus resulting in a smaller accuracy gap.
    }
    \label{fig:placeholder}
\end{figure*}

\subsection{Overview}
Our NatMU method can be decomposed into three steps. Firstly, we randomly select $n$ different instances from the remaining set for each forgetting sample $\boldsymbol{x}^f$. Secondly, an injecting function $\mathcal{T}$ is employed to inject the information of  remaining samples $\boldsymbol{x}^r$ into $\boldsymbol{x}^f$ by blending them at the pixel level, generating an unlearning instance $(\mathcal{T}(\boldsymbol{x}^f, \boldsymbol{x}^r), y^r)$. 
Thirdly, by merging remaining set with all generated unlearning instances, we finetune original model on the resulting finetuning dataset and obtain the final unlearned model.

\subsection{Details}
\textbf{Selection of remaining instances.} 
For a forgetting instance $(\boldsymbol{x}^f, y^f) \in \mathcal{D}^f$, we would select  $n$ remaining instances $\{ (\boldsymbol{x}_j^r, y_j^r)\}_{j=1}^{n}$ from the remaining set. These remaining instances inherently reflect the correct information of the remaining data's distribution.
Firstly, to ensure the effectiveness of unlearning,
the $n$ selected instances should have different categories from $y^f$.
Secondly, to  prevent that the model from having a preference for mapping the forgetting sample to any reassigned category $y_j^r$, these selected instances should also have different categories from each other.
Thirdly, to reduce the conflicts in unlearning instances $(\mathcal{T}(\boldsymbol{x}^f, \boldsymbol{x}^r_j), y^r_j)$, the categories of remaining instances should be relevant to the forgetting sample. 
To meet these requirements, for a forgetting sample $\boldsymbol{x}^f$, we calculate the top $n$ predicted categories of $\boldsymbol{x}^f$ on the original model, excluding the original category $y^f$. Then, we randomly select one remaining instance for each category and get the resulting $n$ remaining instance $\{(\mathcal{T}(\boldsymbol{x}_i^f,\boldsymbol{x}_j^r),  y_j^r) \}_{j=1}^n$.

\textbf{The injecting function $\mathcal{T}$.} 
Motivated by data augmentation like \mixup~\cite{mixup} and CutMix~\cite{cutmix}, we opt for a straightforward approach  where  two samples are added pixel by pixel: $\mathcal{T}(\boldsymbol{x}^f, \boldsymbol{x}^r) = \boldsymbol{x}^f + \boldsymbol{x}^r$. 
To ensure that the hybrid sample  numerically matches the original distribution, a weighting mask vector $\boldsymbol{m} \in [0,1]^d$ is introduced.
This vector controls the contribution of two samples at each pixel as defined by $\mathcal{T}_{\boldsymbol{m}}(\boldsymbol{x}^f, \boldsymbol{x}^r) =  \boldsymbol{x}^f \circ \boldsymbol{m} + \boldsymbol{x}^r \circ (\boldsymbol{1}_d - \boldsymbol{m})$, where $\circ$ represents element-wise multiplication, and $\boldsymbol{1}_d$ represents a $d$-dimensional vectors of ones.  
The weighting operation offers an additional advantage that $\boldsymbol{x}^f \circ \boldsymbol{m}$ can be regarded as a segment of the whole forgetting sample $\boldsymbol{x}^f$. 
This mechanism reduces the proportion of incorrect information in each unlearning instance and allows each hybrid sample to suppress only a portion of $\boldsymbol{x}^f$, leading to a more natural unlearning process and improved unlearning performance.
We have developed four ``gradual \mixup'' weighting vectors $\{ \boldsymbol{m}_1, \boldsymbol{m}_2, \boldsymbol{m}_3, \boldsymbol{m}_4\}$, with symmetrical values that vary gradually in different directions. The corresponding generated hybrid samples are  illustrated in Fig.~\ref{subfig:attetion}.

To facilitate understanding, the weighting vector $\boldsymbol{m}$ can be reshaped into a $H\times W$ matrix $\boldsymbol{M}$ to align with the two-dimensional structure of images.  $\boldsymbol{M}$ can be calculated with:

\begin{equation}
\boldsymbol{M}[\cdot, i] = 
\left\{
\begin{aligned}
& \frac{2}{W-2} (i-1), \hfill 1 \leq i \leq  W/2, \\
& \frac{2}{W-2} (W-i), \hfill W/2 \leq i \leq  W,
\end{aligned}
\right.
\end{equation}
where $\boldsymbol{M}[\cdot, i]$ denotes the value of the $i$-th column and $i \in \{1,2,\cdots,W \}$. The elements in each column share the same value and the value of two edge columns is~0 while that of center edges is~1. The values in other columns transition uniformly from 0 on the edges to~1 in the center. In order to unlearn the forgetting sample from multiple perspectives, we can perform a rotation or take the complement of 
$\boldsymbol{m}$ to obtain different weight vectors. Let $\boldsymbol{m}_1 = \boldsymbol{m}$, we can obtain $\boldsymbol{m}_2$ with complement operation $\boldsymbol{m}_2 = \boldsymbol{1}_d - \boldsymbol{m}_1$, and rotate them by 90 degrees to obtain $\boldsymbol{m}_3$ and $\boldsymbol{m}_4$. Discussions about mask type could be found in the Appendix.

To adjust the proportion of forgetting samples in the hybrid samples, a scaling factor $\delta$ is introduced, which scales the values of weighting vectors as follows:
\begin{equation}\label{eq:scale}
    \boldsymbol{m}_j^\mathrm{scaled} = \mathrm{clip}_{[0,1]}(\boldsymbol{\delta}_d + \boldsymbol{m}_j),
\end{equation}
where $\boldsymbol{\delta}_d$ denotes a $d$-dimensional vector with all elements equal to $\delta$ and $\mathrm{clip}_{[0,1]}(\cdot)$ denotes a function which truncates  vector elements to  $[0,1]$. 
%A detailed description is provided in  Appendix~\ref{appendix:details of our method}.

\textbf{Constructing the fine-tuning dataset.} 
For a forgetting instance $(\boldsymbol{x}_i^f, y_i^f) \in \mathcal{D}^f$, we select four remaining instances $\{ (\boldsymbol{x}_j^r, y_j^r) \in \mathcal{D}^r\}_{j=1}^{n=4} $ according to the above rules.
Then, we generate unlearning instances using the following formula:
\begin{equation}\label{eq:generating}
    \mathcal{D}_i^f =  \{ (\mathcal{T}_{\boldsymbol{m}_j^\mathrm{scaled}}(\boldsymbol{x}_i^f,\boldsymbol{x}_j^r),  y_j^r) \}_{j=1}^n.
\end{equation}
The algorithm for generating the finetuning dataset could be found in the Appendix.
%Algorithm~\ref{algo:gererate dataset}. 
After executing   the aforementioned operation for all forgetting samples, following \cite{amnesiac, salun}, we merge all the sets from these samples with the remaining set to compile the final finetuning dataset:
\begin{equation}
    \mathcal{D}_\mathrm{Nat} = \mathcal{D}^r \cup \mathcal{D}^f_{\mathrm{Nat}},~~ 
    \mathcal{D}^f_{\mathrm{Nat}} =   \mathcal{D}_1^f \cup  \mathcal{D}_2^f \cup \cdots  \mathcal{D}_{| \mathcal{D}^f|}^f.
\end{equation}
Finally, we finetune the original model with the resulting dataset and get the unlearned model, akin to other optimization-based methods.

\begin{table}[ht]
  \centering
  \captionsetup{type=figure}
  \begin{tabular}{C{0.15\linewidth} C{0.3\linewidth} C{0.3\linewidth}}
    % First row: header names
    \toprule
    & Forget Sample 1: \textbf{Retrain's prediction $=$ Pretrain's}  & Forget Sample 2:  \textbf{Retrain's prediction $\neq$ Pretrain's} \\
    \midrule
    & Label: \textcolor{olive}{\textbf{Baboon}} & Label: \textcolor{olive}{\textbf{Oboe}} \\
    Original Image &
    \includegraphics[width=0.8\linewidth]{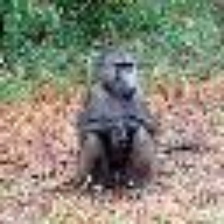} &
    \includegraphics[width=0.8\linewidth]{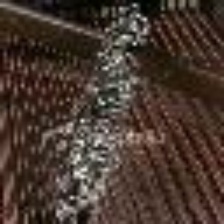} \\
    \midrule
    
    & Prediction: \textcolor{olive}{\textbf{Baboon}} & Prediction: \textcolor{olive}{\textbf{Oboe}} \\
    Pretrain's Attention &
    \includegraphics[width=0.8\linewidth]{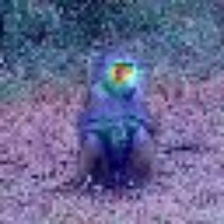} &
    \includegraphics[width=0.8\linewidth]{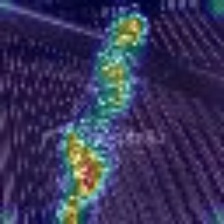} \\
    \midrule
    
    & Prediction: \textcolor{olive}{\textbf{Baboon}} & Prediction: \textcolor{orange}{\textbf{Thatch Roof}} \\
    Retrain's Attention &
    \includegraphics[width=0.8\linewidth]{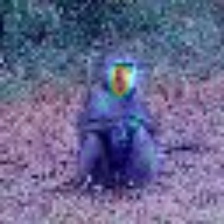} &
    \includegraphics[width=0.8\linewidth]{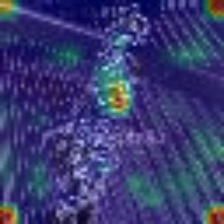} \\
    \midrule

    & Prediction: \textcolor{olive}{\textbf{Baboon}} & Prediction: \textcolor{orange}{\textbf{Thatch Roof}} \\
    NatMU's Attention &
    \includegraphics[width=0.8\linewidth]{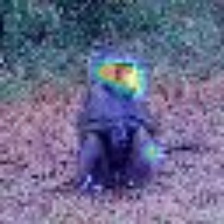} &
    \includegraphics[width=0.8\linewidth]{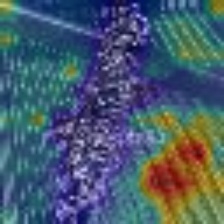} \\
    \midrule

    & Prediction: \textcolor{purple}{\textbf{Mollymawk}} & Prediction: \textcolor{purple}{\textbf{Fountain}}  \\
    Baseline's Attention &
    \includegraphics[width=0.8\linewidth]{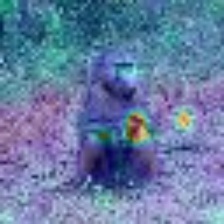} &
    \includegraphics[width=0.8\linewidth]{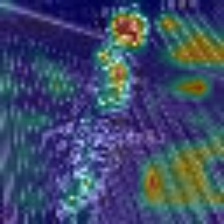} \\

    \bottomrule
  \end{tabular}
  \caption{Visualization of attention maps using TinyViT~\cite{tinyvit} for different unlearning methods on TinyImageNet. The attention maps are calculated with Eigen-Grad-CAM~\cite{DBLP:conf/ijcnn/MuhammadY20, DBLP:conf/iccv/SelvarajuCDVPB17, jacobgilpytorchcam}. NatMU can approach the behavior of the retrained model, while the baseline method (Amnesiac) faces the over-forgetting problem and shows significantly different attention maps from the retrained model.}
  \label{fig:attention-vit}
\end{table}

\subsection{Unlearning Mechanism}
Given an unlearning instance $(\boldsymbol{x}^f \circ \boldsymbol{m} + \boldsymbol{x}^r \circ (\boldsymbol{1}_d - \boldsymbol{m}), y^r) \in \mathcal{D}^f_\mathrm{Nat}$, its input sample consists of two distinct parts of information: 
the information to be forgotten, denoted as $\boldsymbol{x}^f \circ \boldsymbol{m}$, and the injected information, denoted as $\boldsymbol{x}^r \circ (\boldsymbol{1}_d -\boldsymbol{m})$.
In the original pre-trained model, the former information is connected to the forgetting label $y^f$, while the latter information is connected to the reassigned label $y^r$.
During the fine-tuning of unlearning process, the unlearned model learns to pair the unlearning sample with label $y^r$, resulting the reinforcement of the latter connection. It naturally suppresses model's response to the forgetting information, since such response is harmful to make accurate predictions for unlearning instances. As a result, it  effectively removes the learned knowledge in $\boldsymbol{x}^f \circ \boldsymbol{m}$.

Compared to previous relabeling-based work, by training with unlearning set and remaining set, the reinforced information in NatMU is the connection between $\boldsymbol{x}^r\circ(\boldsymbol{1}_d-\boldsymbol{m})$ and $y^r$, which inherently exists in the remaining set. 
This prevents the model from establishing undesirable new associations between $\boldsymbol{x}^f \circ \boldsymbol{m}$ and $y^r$,
thereby promoting a more natural unlearning process.

Fig.~\ref{subfig:attetion} visualizes the unlearning instances of NatMU and their attention maps on hybrid samples before and after unlearning. 
%In the second column, different weighting masks capture different parts of forgetting samples.  
By comparing the attention maps before and after unlearning, we can see that model's attention is effectively shifted to the prominent positions of the remaining samples through unlearning. For example, in the first row, before unlearning the model focuses on the bird's wings in $\boldsymbol{x}^f \circ \boldsymbol{m}$. After unlearning, the model shifts it attention to the butterfly below the bird, which comes from $\boldsymbol{x}^r \circ ( \boldsymbol{1}_d -\boldsymbol{m})$. {Fig.~\ref{fig:attention-vit} furthermore  presents the attention maps of different models before and after unlearning.
The baseline method misclassifies the forgetting samples into classes that differ from those predicted by the retrained model. Moreover, its attention maps are quite different from that of the retrained model. In contrast, NatMU can not only successfully make predictions consistent with Retrain but also generate similar attention maps. This highlights NatMU’s ability to achieve better consistency with Retrain in both prediction and attention.}

\subsection{Natural Property}
\label{sec:natural-property}

The injected information is extracted from remaining data, therefore, the unlearning instances of NatMU  align with the  distribution of remaining data  more closely. It significantly reduces the conflict between the unlearning instances and the remaining data.
To quantify the degree of data conflict and how ``natural'' these unlearning instances are, we adopt the KL divergence~\cite{kullback1951information} between the reassigned unlearning labels and  retrained model's predictions on the unlearning samples. 
{Since the retrained model's knowledge is all from the remaining dataset, it could be used as a proxy to measure how much the unlearning data align with the remaining dataset. A smaller KL divergence means the reassigned labels are closer to the retrained model's prediction, indicating a more natural unlearning process.}
As no unlearning samples are involved in retraining, we define its KL divergence as zero.
In Fig.~\ref{subfig:natural-rate}, we demonstrate the average KL divergence of different methods over their unlearning dataset ${D}_\mathrm{ul}$ :
\begin{equation}
   \mathrm{KL_{avg}}=\frac{1}{|\mathcal{D}_\mathrm{ul}|}\Sigma_{(\boldsymbol{x}_i, y_i) \in \mathcal{D}_\mathrm{ul} } \mathrm{KL}(f_{\theta_r}(\boldsymbol{x}_i), y_i),
\end{equation}
where $\mathrm{KL}(\cdot, \cdot)$ calculates the KL divergence of two labels.
For Amnesiac, the unlearning dataset ${D}_\mathrm{ul}$ is $\mathcal{D}^f_\mathrm{RL}$; for BadTeacher, ${D}_\mathrm{ul}=\mathcal{D}^f_\mathrm{bad}$; and for NatMU, ${D}_\mathrm{ul}=\mathcal{D}^f_\mathrm{Nat}$.
One can see that NatMU has a much smaller KL divergence compared to other methods, indicating that its unlearning is more natural. Therefore, NatMU's performance is closer to that of the retrained model.

\subsection{Preventing Undesirable Reinforcement}

\begin{figure}[ht]
\centering
\includegraphics[width=0.4\textwidth]{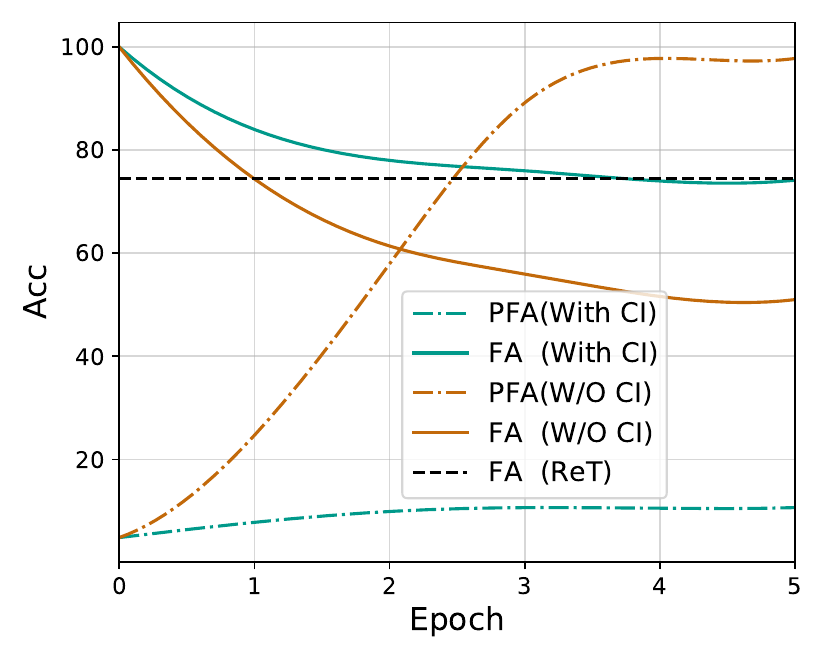}
\caption{Accuracy curves  on partial unlearning instances $\{ (\boldsymbol{x}^f \circ \boldsymbol{m}, y^r)\}$ and  forgetting accuracy of different MU models trained with/without correct information. The experiment is conducted on CIFAR-100 with a forgetting ratio of 1\% using ResNet-18.  
    PFA: accuracy of classifying partial forgetting samples $\boldsymbol{x}^f\circ \boldsymbol{m}$ as random label $y^r$. 
    FA: forgetting accuracy. 
    CI: correct information.
    ReT: the retrained model. }
\label{fig:avoiding incorrect associations}
\end{figure}

In previous work, incorrect information is reinforced during unlearning, leading to the over-forgetting problem. 
Specifically, the model establishes incorrect associations between the forgetting samples and their reassigned labels.
Although NatMU also modifies the forgetting labels, the existence of correct information from $x^r$ can effectively prevent these incorrect associations. 
To verify this, we demonstrate the accuracy curves of sample-label pairs $\{ (\boldsymbol{x}^f \circ \boldsymbol{m}, y^r)\}$ and forgetting accuracy curves of different MU models trained with or without the correct information $\boldsymbol{x}^r\circ(\boldsymbol{1}_d -\boldsymbol{m})$ in Fig.~\ref{fig:avoiding incorrect associations}. 
It can be seen that, without the injected correct information, the model quickly recognizes the partial forgetting sample $\boldsymbol{x}^f \circ \boldsymbol{m}$ as the reassigned class $y^r$. As a result, the forgetting accuracy faces a significant decrease. In contrast, the model trained with correct information consistently remains a low accuracy on $\{ (\boldsymbol{x}^f \circ \boldsymbol{m}, y^r)\}$, showing that injecting correct information can effectively prevent undesirable reinforcement.

The injected correct information form remaining data helps NatMU achieve a more natural machine unlearning, not forced unlearning as other relabeling-based methods do. The natural generalization on forgetting samples is well preserved in NatMU. 
As a result, NatMU can achieve  a smaller performance gap with the retrained model on multiple metrics. Moreover, the natural property gives NatMU greater robustness, allowing us to transfer hyperparameters to unknown settings, making NatMU a promising candidate for practical MU.

\section{Experiments}
In this section, we evaluate NatMU against other baselines across different datasets, models, and unlearning scenarios.
We firstly evaluate unlearning methods under class-wise unlearning, including full-class unlearning and sub-class unlearning, with ideal hyperparameters and practical hyperparameters. Then we evaluate their performance under sample-wise unlearning, where the forgetting samples are randomly selected from the training set or are selected from a specific unknown distribution.  We also provide a hyperparameter tuning method without requiring any retrained model to improve the unlearning methods' practicality.
Finally, ablation studies are conducted to investigate the role of different part of NatMU.

% ---------------------------------------------------------------
% ---------------------------------------------------------------

% Setup

% ---------------------------------------------------------------
% ---------------------------------------------------------------

\subsection{Setup}\label{setup}

\textbf{Datasets and models.}
NatMU is evaluated against other machine unlearning methods in the context of supervised image classification tasks using the CIFAR-10, CIFAR-20, CIFAR-100~\cite{cifar}, and TinyImageNet-200~\cite{tiny-imagenet} datasets. It is noteworthy that CIFAR-20 and CIFAR-100 are closely related.  CIFAR-100 dataset consists of 20 superclasses, each containing 5 subclasses, resulting in a total of 100 classes. 
When only considering  the superclasses, CIFAR-100 reduces to CIFAR-20.
Models of various architectures are trained in different unlearning scenarios, including CNN  models (VGG16-BN~\cite{vgg}, ResNet-18, ResNet-34~\cite{resnet}) and Vision Transformer (TinyViT~\cite{tinyvit}).

\textbf{Unlearning scenarios.} 
Following \cite{badteacher} and \cite{ssd}, we evaluate MU methods across three unlearning scenarios: (1) \emph{full-class unlearning}, (2) \emph{sub-class unlearning}, and (3) \emph{sample-wise unlearning}. For each unlearning scenario, we also evaluate different methods under different tasks. For class-wise unlearning, we select three unlearning classes: ``Rocket'', ``Cattle'' and ``Sea'', where the retrained model's forgetting accuracy varies differently. For sample-wise unlearning, methods are evaluated with 1\% forgetting data and 10\% forgetting data.

\textbf{Baselines.}
We compare NatMU with multiple state-of-the-art MU methods:
one optimization-free method \textit{SSD}~\cite{ssd}, two relabeling-free methods: \textit{NegGrad+} and \textit{SCRUB}~\cite{scrub}, five relabeling-based methods:
\textit{$\delta\text{-targeted}^\lambda$~\cite{targeted}}, \textit{AMUN~\cite{amun}},
\textit{Amnesiac}~\cite{amnesiac}, 
\textit{BadTeacher}~\cite{badteacher} and 
\textit{SalUn}~\cite{salun}.
NegGrad+ serves as a baseline method which combines two basic unlearning methods: gradient descent on remaining data (finetuning the model with remaining data) and gradient ascent on forgetting data.
SSD's results are not shown in sample-wise unlearning experiments since it fails to realize effective unlearning while maintaining model utility.
For all optimization-based methods, the training epoch is set to 5, which is 5\% of pretraining epoch following~\cite{salun}. The experimental details could be found in the Appendix.
%Appendix~\ref{appendix:experiment-details}.

\textbf{Evaluation metrics.} 
Following \cite{badteacher,salun,ssd,sparsity,scrub}, we evaluate MU methods across three metrics: test accuracy~(\textbf{TA}), remaining accuracy~(\textbf{RA}), forgetting accuracy~(\textbf{FA}) and membership inference attack~(\textbf{MIA}) ratio \cite{mia}. Test accuracy  calculates the accuracy on test data, which measures the generalization ability of the unlearned model. Remaining accuracy  calculates the accuracy on the remaining data, measuring the fidelity of an unlearned model on the remaining dataset $\mathcal{D}^r$. Forgetting accuracy evaluates the model's accuracy on the forgetting samples, which measures the effectiveness of MU. In full/sub-class unlearning, we calculate accuracy on samples of the forgetting class in training set and test set respectively, denoted as \textbf{FATrain} and \textbf{FATest}. 
\begin{table*}[t!]
\centering
\caption{
Class-wise unlearning results on CIFAR-100/20 using ResNet-18 with different forgetting classes.
The results are given by $\bm{a_{\pm b}(\textcolor{blue}{c})}$, where $\bm{a}$ denotes the mean value, $\bm{b}$ denotes the standard deviation, and $\bm{\textcolor{blue}{c}}$ denotes the performance gap with the retrained model over 5 independent trails. 
A smaller $\bm{\textcolor{blue}{c}}$ means a better performance. }
\label{table:class-wise}
\resizebox{\textwidth}{!}{
\begin{tabular}{cccccccccccccc}
\toprule[1.5pt]
\multirow{2}{*}{\begin{tabular}[c]{@{}c@{}}Unlearning\\ task\end{tabular}} 
&
\multirow{2}{*}{\begin{tabular}[c]{@{}c@{}}Hyperparameter\\ Type\end{tabular}} 
& \multirow{2}{*}{Metric} 
& \multicolumn{10}{c}{Methods} \\ 
\cmidrule[0.75pt]{4-13} 
 
 & &  & Retrain & SSD & NegGrad+ &SCRUB & AMUN & $\delta\text{-targeted}^\lambda$ &  Amnesiac & BadTeacher & SalUn  &NatMU \\ \midrule[1pt]
%----------------------------------------------------------------------------------
{\multirow{6}{*}{\begin{tabular}[c]{@{}c@{}}Full-class\\(Rocket) \end{tabular}}} 
& {\multirow{6}{*}{\begin{tabular}[c]{@{}c@{}} Ideal \end{tabular}}} 
&
TA
& 76.60\textsubscript{\textpm0.38}
& 76.68\textsubscript{\textpm0.03}(\textcolor{blue}{0.08})
& 75.90\textsubscript{\textpm0.18}(\textcolor{blue}{0.69})
& 75.71\textsubscript{\textpm0.31}(\textcolor{blue}{0.88})
& 76.71\textsubscript{\textpm0.19}(\textcolor{blue}{0.12})
& 76.93\textsubscript{\textpm0.10}(\textcolor{blue}{0.34})
& 76.56\textsubscript{\textpm0.23}(\textcolor{blue}{0.04})
& 76.30\textsubscript{\textpm0.29}(\textcolor{blue}{0.30})
& 76.53\textsubscript{\textpm0.22}(\textcolor{blue}{0.07})
& 76.27\textsubscript{\textpm0.08}(\textcolor{blue}{0.32})

 \\ 
&
&
RA
& 99.94\textsubscript{\textpm0.01}
& 99.95\textsubscript{\textpm0.00}(\textcolor{blue}{0.01})
& 99.97\textsubscript{\textpm0.00}(\textcolor{blue}{0.03})
& 99.96\textsubscript{\textpm0.00}(\textcolor{blue}{0.02})
& 99.98\textsubscript{\textpm0.00}(\textcolor{blue}{0.04})
& 99.98\textsubscript{\textpm0.00}(\textcolor{blue}{0.04})
& 99.98\textsubscript{\textpm0.00}(\textcolor{blue}{0.04})
& 99.85\textsubscript{\textpm0.01}(\textcolor{blue}{0.10})
& 99.98\textsubscript{\textpm0.00}(\textcolor{blue}{0.04})
& 99.55\textsubscript{\textpm0.01}(\textcolor{blue}{0.39})

 \\ 
&
&
FATrain
& 0.00\textsubscript{\textpm0.00}
& 0.00\textsubscript{\textpm0.00}(\textcolor{blue}{0.00})
& 0.00\textsubscript{\textpm0.00}(\textcolor{blue}{0.00})
& 0.04\textsubscript{\textpm0.08}(\textcolor{blue}{0.04})
& 0.00\textsubscript{\textpm0.00}(\textcolor{blue}{0.00})
& 0.56\textsubscript{\textpm0.23}(\textcolor{blue}{0.56})
& 0.00\textsubscript{\textpm0.00}(\textcolor{blue}{0.00})
& 0.00\textsubscript{\textpm0.00}(\textcolor{blue}{0.00})
& 0.00\textsubscript{\textpm0.00}(\textcolor{blue}{0.00})
& 0.00\textsubscript{\textpm0.00}(\textcolor{blue}{0.00})

 \\ 
&
&
FATest
& 0.00\textsubscript{\textpm0.00}
& 0.00\textsubscript{\textpm0.00}(\textcolor{blue}{0.00})
& 0.00\textsubscript{\textpm0.00}(\textcolor{blue}{0.00})
& 0.00\textsubscript{\textpm0.00}(\textcolor{blue}{0.00})
& 0.00\textsubscript{\textpm0.00}(\textcolor{blue}{0.00})
& 0.00\textsubscript{\textpm0.00}(\textcolor{blue}{0.00})
& 0.00\textsubscript{\textpm0.00}(\textcolor{blue}{0.00})
& 0.00\textsubscript{\textpm0.00}(\textcolor{blue}{0.00})
& 0.00\textsubscript{\textpm0.00}(\textcolor{blue}{0.00})
& 0.00\textsubscript{\textpm0.00}(\textcolor{blue}{0.00})

 \\ 
&
&
MIA
& 12.68\textsubscript{\textpm0.92}
& 1.16\textsubscript{\textpm0.29}(\textcolor{blue}{11.52})
& 11.04\textsubscript{\textpm3.15}(\textcolor{blue}{1.64})
& 8.92\textsubscript{\textpm1.41}(\textcolor{blue}{3.76})
& 7.36\textsubscript{\textpm0.73}(\textcolor{blue}{5.32})
& 9.80\textsubscript{\textpm0.52}(\textcolor{blue}{2.88})
& 7.57\textsubscript{\textpm1.11}(\textcolor{blue}{5.11})
& 0.00\textsubscript{\textpm0.00}(\textcolor{blue}{12.68})
& 7.96\textsubscript{\textpm0.54}(\textcolor{blue}{4.72})
& 11.40\textsubscript{\textpm0.95}(\textcolor{blue}{1.28})

 \\ 
\arrayrulecolor{lightgray} 
\cmidrule[0.75pt]{3-13} \arrayrulecolor{black}

&
&  Avg.Gap$\downarrow$  
 & - 
& \textcolor{blue}{2.32}
& \textcolor{blue}{0.47}
& \textcolor{blue}{0.94}
& \textcolor{blue}{1.09}
& \textcolor{blue}{0.76}
& \textcolor{blue}{1.04}
& \textcolor{blue}{2.62}
& \textcolor{blue}{0.97}
& \textbf{\textcolor{blue}{0.40}}

\\ \midrule
%----------------------------------------------------------------------------------
%----------------------------------------------------------------------------------
{\multirow{6}{*}{\begin{tabular}[c]{@{}c@{}}Full-class\\(Cattle) \end{tabular}}} 
& {\multirow{6}{*}{\begin{tabular}[c]{@{}c@{}} Ideal \end{tabular}}} 
&
TA
& 76.53\textsubscript{\textpm0.21}
& 76.80\textsubscript{\textpm0.07}(\textcolor{blue}{0.27})
& 76.12\textsubscript{\textpm0.14}(\textcolor{blue}{0.41})
& 75.46\textsubscript{\textpm0.12}(\textcolor{blue}{1.07})
& 77.12\textsubscript{\textpm0.15}(\textcolor{blue}{0.59})
& 77.08\textsubscript{\textpm0.05}(\textcolor{blue}{0.56})
& 76.69\textsubscript{\textpm0.14}(\textcolor{blue}{0.16})
& 76.39\textsubscript{\textpm0.25}(\textcolor{blue}{0.14})
& 76.74\textsubscript{\textpm0.11}(\textcolor{blue}{0.21})
& 76.52\textsubscript{\textpm0.09}(\textcolor{blue}{0.00})

 \\ 
&
&
RA
& 99.94\textsubscript{\textpm0.00}
& 99.95\textsubscript{\textpm0.00}(\textcolor{blue}{0.01})
& 99.98\textsubscript{\textpm0.00}(\textcolor{blue}{0.03})
& 99.96\textsubscript{\textpm0.01}(\textcolor{blue}{0.02})
& 99.98\textsubscript{\textpm0.00}(\textcolor{blue}{0.04})
& 99.98\textsubscript{\textpm0.00}(\textcolor{blue}{0.04})
& 99.98\textsubscript{\textpm0.00}(\textcolor{blue}{0.04})
& 99.85\textsubscript{\textpm0.01}(\textcolor{blue}{0.09})
& 99.98\textsubscript{\textpm0.00}(\textcolor{blue}{0.04})
& 99.55\textsubscript{\textpm0.03}(\textcolor{blue}{0.39})

 \\ 
&
&
FATrain
& 0.00\textsubscript{\textpm0.00}
& 0.00\textsubscript{\textpm0.00}(\textcolor{blue}{0.00})
& 0.00\textsubscript{\textpm0.00}(\textcolor{blue}{0.00})
& 0.00\textsubscript{\textpm0.00}(\textcolor{blue}{0.00})
& 0.00\textsubscript{\textpm0.00}(\textcolor{blue}{0.00})
& 0.08\textsubscript{\textpm0.10}(\textcolor{blue}{0.08})
& 0.00\textsubscript{\textpm0.00}(\textcolor{blue}{0.00})
& 0.00\textsubscript{\textpm0.00}(\textcolor{blue}{0.00})
& 0.00\textsubscript{\textpm0.00}(\textcolor{blue}{0.00})
& 0.00\textsubscript{\textpm0.00}(\textcolor{blue}{0.00})

 \\ 
&
&
FATest
& 0.00\textsubscript{\textpm0.00}
& 0.00\textsubscript{\textpm0.00}(\textcolor{blue}{0.00})
& 0.00\textsubscript{\textpm0.00}(\textcolor{blue}{0.00})
& 0.00\textsubscript{\textpm0.00}(\textcolor{blue}{0.00})
& 0.00\textsubscript{\textpm0.00}(\textcolor{blue}{0.00})
& 3.20\textsubscript{\textpm1.17}(\textcolor{blue}{3.20})
& 0.00\textsubscript{\textpm0.00}(\textcolor{blue}{0.00})
& 0.00\textsubscript{\textpm0.00}(\textcolor{blue}{0.00})
& 0.00\textsubscript{\textpm0.00}(\textcolor{blue}{0.00})
& 0.00\textsubscript{\textpm0.00}(\textcolor{blue}{0.00})

 \\ 
&
&
MIA
& 8.72\textsubscript{\textpm0.99}
& 1.96\textsubscript{\textpm0.08}(\textcolor{blue}{6.76})
& 5.16\textsubscript{\textpm0.43}(\textcolor{blue}{3.56})
& 9.76\textsubscript{\textpm1.08}(\textcolor{blue}{1.04})
& 15.24\textsubscript{\textpm0.80}(\textcolor{blue}{6.52})
& 7.72\textsubscript{\textpm1.48}(\textcolor{blue}{1.00})
& 13.53\textsubscript{\textpm1.19}(\textcolor{blue}{4.81})
& 0.00\textsubscript{\textpm0.00}(\textcolor{blue}{8.72})
& 11.24\textsubscript{\textpm1.60}(\textcolor{blue}{2.52})
& 8.88\textsubscript{\textpm1.49}(\textcolor{blue}{0.16})

 \\ 
\arrayrulecolor{lightgray} 
\cmidrule[0.75pt]{3-13} \arrayrulecolor{black}

&
&  Avg.Gap$\downarrow$  
 & - 
& \textcolor{blue}{1.41}
& \textcolor{blue}{0.80}
& \textcolor{blue}{0.42}
& \textcolor{blue}{1.43}
& \textcolor{blue}{0.97}
& \textcolor{blue}{1.00}
& \textcolor{blue}{1.79}
& \textcolor{blue}{0.55}
& \textbf{\textcolor{blue}{0.11}}
\\ \midrule
%----------------------------------------------------------------------------------
%----------------------------------------------------------------------------------
{\multirow{6}{*}{\begin{tabular}[c]{@{}c@{}}Full-class\\(Sea) \end{tabular}}} 
& {\multirow{6}{*}{\begin{tabular}[c]{@{}c@{}} Practical \end{tabular}}} 
&
TA
& 76.69\textsubscript{\textpm0.14}
& 76.83\textsubscript{\textpm0.03}(\textcolor{blue}{0.14})
& 76.10\textsubscript{\textpm0.16}(\textcolor{blue}{0.59})
& 75.45\textsubscript{\textpm0.15}(\textcolor{blue}{1.23})
& 77.08\textsubscript{\textpm0.15}(\textcolor{blue}{0.39})
& 77.13\textsubscript{\textpm0.06}(\textcolor{blue}{0.44})
& 76.43\textsubscript{\textpm0.10}(\textcolor{blue}{0.26})
& 76.28\textsubscript{\textpm0.13}(\textcolor{blue}{0.41})
& 76.54\textsubscript{\textpm0.16}(\textcolor{blue}{0.15})
& 76.46\textsubscript{\textpm0.12}(\textcolor{blue}{0.23})

 \\ 
&
&
RA
& 99.94\textsubscript{\textpm0.01}
& 99.96\textsubscript{\textpm0.00}(\textcolor{blue}{0.01})
& 99.97\textsubscript{\textpm0.00}(\textcolor{blue}{0.03})
& 99.96\textsubscript{\textpm0.01}(\textcolor{blue}{0.02})
& 99.98\textsubscript{\textpm0.00}(\textcolor{blue}{0.04})
& 99.98\textsubscript{\textpm0.00}(\textcolor{blue}{0.04})
& 99.98\textsubscript{\textpm0.00}(\textcolor{blue}{0.04})
& 99.86\textsubscript{\textpm0.01}(\textcolor{blue}{0.09})
& 99.98\textsubscript{\textpm0.00}(\textcolor{blue}{0.04})
& 99.57\textsubscript{\textpm0.01}(\textcolor{blue}{0.37})

 \\ 
&
&
FATrain
& 0.00\textsubscript{\textpm0.00}
& 0.28\textsubscript{\textpm0.56}(\textcolor{blue}{0.28})
& 0.00\textsubscript{\textpm0.00}(\textcolor{blue}{0.00})
& 5.60\textsubscript{\textpm1.97}(\textcolor{blue}{5.60})
& 0.00\textsubscript{\textpm0.00}(\textcolor{blue}{0.00})
& 0.44\textsubscript{\textpm0.08}(\textcolor{blue}{0.44})
& 0.00\textsubscript{\textpm0.00}(\textcolor{blue}{0.00})
& 0.00\textsubscript{\textpm0.00}(\textcolor{blue}{0.00})
& 0.00\textsubscript{\textpm0.00}(\textcolor{blue}{0.00})
& 0.00\textsubscript{\textpm0.00}(\textcolor{blue}{0.00})

 \\ 
&
&
FATest
& 0.00\textsubscript{\textpm0.00}
& 0.60\textsubscript{\textpm1.20}(\textcolor{blue}{0.60})
& 0.00\textsubscript{\textpm0.00}(\textcolor{blue}{0.00})
& 4.00\textsubscript{\textpm1.41}(\textcolor{blue}{4.00})
& 0.00\textsubscript{\textpm0.00}(\textcolor{blue}{0.00})
& 1.00\textsubscript{\textpm0.00}(\textcolor{blue}{1.00})
& 0.00\textsubscript{\textpm0.00}(\textcolor{blue}{0.00})
& 0.00\textsubscript{\textpm0.00}(\textcolor{blue}{0.00})
& 0.00\textsubscript{\textpm0.00}(\textcolor{blue}{0.00})
& 0.00\textsubscript{\textpm0.00}(\textcolor{blue}{0.00})

 \\ 
&
&
MIA
& 24.56\textsubscript{\textpm1.46}
& 2.40\textsubscript{\textpm0.47}(\textcolor{blue}{22.16})
& 17.36\textsubscript{\textpm1.05}(\textcolor{blue}{7.20})
& 8.88\textsubscript{\textpm0.97}(\textcolor{blue}{15.68})
& 19.48\textsubscript{\textpm1.37}(\textcolor{blue}{5.08})
& 21.36\textsubscript{\textpm1.13}(\textcolor{blue}{3.20})
& 7.91\textsubscript{\textpm1.49}(\textcolor{blue}{16.65})
& 0.00\textsubscript{\textpm0.00}(\textcolor{blue}{24.56})
& 7.84\textsubscript{\textpm0.99}(\textcolor{blue}{16.72})
& 25.72\textsubscript{\textpm2.27}(\textcolor{blue}{1.16})

 \\ 
\arrayrulecolor{lightgray} 
\cmidrule[0.75pt]{3-13} \arrayrulecolor{black}

&
&  Avg.Gap$\downarrow$  
 & - 
& \textcolor{blue}{4.64}
& \textcolor{blue}{1.56}
& \textcolor{blue}{5.31}
& \textcolor{blue}{1.10}
& \textcolor{blue}{1.02}
& \textcolor{blue}{3.39}
& \textcolor{blue}{5.01}
& \textcolor{blue}{3.38}
& \textbf{\textcolor{blue}{0.35}}
\\ \midrule
%----------------------------------------------------------------------------------
%----------------------------------------------------------------------------------
{\multirow{6}{*}{\begin{tabular}[c]{@{}c@{}}Sub-class\\(Rocket) \end{tabular}}} 
& {\multirow{6}{*}{\begin{tabular}[c]{@{}c@{}} Ideal \end{tabular}}} 
&
TA
& 84.94\textsubscript{\textpm0.15}
& 84.82\textsubscript{\textpm0.06}(\textcolor{blue}{0.12})
& 84.76\textsubscript{\textpm0.25}(\textcolor{blue}{0.18})
& 83.71\textsubscript{\textpm0.28}(\textcolor{blue}{1.23})
& 84.97\textsubscript{\textpm0.24}(\textcolor{blue}{0.03})
& 84.80\textsubscript{\textpm0.08}(\textcolor{blue}{0.14})
& 84.84\textsubscript{\textpm0.01}(\textcolor{blue}{0.10})
& 84.97\textsubscript{\textpm0.06}(\textcolor{blue}{0.03})
& 84.71\textsubscript{\textpm0.11}(\textcolor{blue}{0.23})
& 84.51\textsubscript{\textpm0.22}(\textcolor{blue}{0.43})

 \\ 
&
&
RA
& 99.96\textsubscript{\textpm0.00}
& 99.84\textsubscript{\textpm0.04}(\textcolor{blue}{0.12})
& 99.99\textsubscript{\textpm0.00}(\textcolor{blue}{0.03})
& 99.98\textsubscript{\textpm0.00}(\textcolor{blue}{0.02})
& 99.99\textsubscript{\textpm0.00}(\textcolor{blue}{0.04})
& 99.97\textsubscript{\textpm0.01}(\textcolor{blue}{0.01})
& 99.98\textsubscript{\textpm0.00}(\textcolor{blue}{0.03})
& 99.96\textsubscript{\textpm0.00}(\textcolor{blue}{0.00})
& 99.98\textsubscript{\textpm0.00}(\textcolor{blue}{0.02})
& 98.22\textsubscript{\textpm0.09}(\textcolor{blue}{1.74})

 \\ 
&
&
FATrain
& 2.80\textsubscript{\textpm0.28}
& 12.36\textsubscript{\textpm3.63}(\textcolor{blue}{9.56})
& 2.00\textsubscript{\textpm0.46}(\textcolor{blue}{0.80})
& 3.44\textsubscript{\textpm0.23}(\textcolor{blue}{0.64})
& 6.84\textsubscript{\textpm1.50}(\textcolor{blue}{4.04})
& 7.80\textsubscript{\textpm0.46}(\textcolor{blue}{5.00})
& 5.64\textsubscript{\textpm0.27}(\textcolor{blue}{2.84})
& 10.04\textsubscript{\textpm2.36}(\textcolor{blue}{7.24})
& 7.68\textsubscript{\textpm0.93}(\textcolor{blue}{4.88})
& 5.56\textsubscript{\textpm0.72}(\textcolor{blue}{2.76})

 \\ 
&
&
FATest
& 1.60\textsubscript{\textpm0.49}
& 3.20\textsubscript{\textpm1.94}(\textcolor{blue}{1.60})
& 1.00\textsubscript{\textpm0.00}(\textcolor{blue}{0.60})
& 1.80\textsubscript{\textpm0.75}(\textcolor{blue}{0.20})
& 9.00\textsubscript{\textpm2.45}(\textcolor{blue}{7.40})
& 5.00\textsubscript{\textpm0.00}(\textcolor{blue}{3.40})
& 2.80\textsubscript{\textpm0.75}(\textcolor{blue}{1.20})
& 3.80\textsubscript{\textpm1.17}(\textcolor{blue}{2.20})
& 2.80\textsubscript{\textpm0.75}(\textcolor{blue}{1.20})
& 2.40\textsubscript{\textpm0.80}(\textcolor{blue}{0.80})

 \\ 
&
&
MIA
& 19.32\textsubscript{\textpm1.61}
& 6.08\textsubscript{\textpm1.00}(\textcolor{blue}{13.24})
& 31.96\textsubscript{\textpm19.77}(\textcolor{blue}{12.64})
& 23.68\textsubscript{\textpm2.56}(\textcolor{blue}{4.36})
& 14.04\textsubscript{\textpm2.10}(\textcolor{blue}{5.28})
& 14.48\textsubscript{\textpm1.08}(\textcolor{blue}{4.84})
& 0.00\textsubscript{\textpm0.00}(\textcolor{blue}{19.32})
& 0.00\textsubscript{\textpm0.00}(\textcolor{blue}{19.32})
& 0.00\textsubscript{\textpm0.00}(\textcolor{blue}{19.32})
& 16.36\textsubscript{\textpm0.83}(\textcolor{blue}{2.96})

 \\ 
\arrayrulecolor{lightgray} 
\cmidrule[0.75pt]{3-13} \arrayrulecolor{black}

&
&  Avg.Gap$\downarrow$  
 & - 
& \textcolor{blue}{4.93}
& \textcolor{blue}{2.85}
& \textbf{\textcolor{blue}{1.29}}
& \textcolor{blue}{3.36}
& \textcolor{blue}{2.68}
& \textcolor{blue}{4.70}
& \textcolor{blue}{5.76}
& \textcolor{blue}{5.13}
& \textcolor{blue}{1.74}

\\ \midrule

{\multirow{6}{*}{\begin{tabular}[c]{@{}c@{}}Sub-class\\(Cattle) \end{tabular}}} 
& {\multirow{6}{*}{\begin{tabular}[c]{@{}c@{}} Ideal \end{tabular}}} 
&
TA
& 85.01\textsubscript{\textpm0.17}
& 83.84\textsubscript{\textpm0.32}(\textcolor{blue}{1.16})
& 84.76\textsubscript{\textpm0.13}(\textcolor{blue}{0.24})
& 83.14\textsubscript{\textpm0.49}(\textcolor{blue}{1.86})
& 84.83\textsubscript{\textpm0.20}(\textcolor{blue}{0.18})
& 84.58\textsubscript{\textpm0.09}(\textcolor{blue}{0.43})
& 84.64\textsubscript{\textpm0.15}(\textcolor{blue}{0.37})
& 84.85\textsubscript{\textpm0.12}(\textcolor{blue}{0.15})
& 84.32\textsubscript{\textpm0.11}(\textcolor{blue}{0.69})
& 84.55\textsubscript{\textpm0.11}(\textcolor{blue}{0.46})

 \\ 
&
&
RA
& 99.96\textsubscript{\textpm0.01}
& 98.23\textsubscript{\textpm0.54}(\textcolor{blue}{1.73})
& 99.99\textsubscript{\textpm0.00}(\textcolor{blue}{0.02})
& 99.97\textsubscript{\textpm0.01}(\textcolor{blue}{0.00})
& 99.99\textsubscript{\textpm0.00}(\textcolor{blue}{0.03})
& 99.90\textsubscript{\textpm0.01}(\textcolor{blue}{0.07})
& 99.99\textsubscript{\textpm0.00}(\textcolor{blue}{0.02})
& 99.95\textsubscript{\textpm0.01}(\textcolor{blue}{0.01})
& 99.96\textsubscript{\textpm0.00}(\textcolor{blue}{0.01})
& 98.21\textsubscript{\textpm0.04}(\textcolor{blue}{1.76})

 \\ 
&
&
FATrain
& 45.20\textsubscript{\textpm1.48}
& 35.34\textsubscript{\textpm14.18}(\textcolor{blue}{9.86})
& 64.44\textsubscript{\textpm2.96}(\textcolor{blue}{19.24})
& 40.12\textsubscript{\textpm7.81}(\textcolor{blue}{5.08})
& 26.48\textsubscript{\textpm2.40}(\textcolor{blue}{18.72})
& 40.40\textsubscript{\textpm1.92}(\textcolor{blue}{4.80})
& 46.88\textsubscript{\textpm1.91}(\textcolor{blue}{1.68})
& 45.40\textsubscript{\textpm7.25}(\textcolor{blue}{0.20})
& 50.16\textsubscript{\textpm1.93}(\textcolor{blue}{4.96})
& 47.60\textsubscript{\textpm2.72}(\textcolor{blue}{2.40})

 \\ 
&
&
FATest
& 49.60\textsubscript{\textpm2.42}
& 25.50\textsubscript{\textpm11.88}(\textcolor{blue}{24.10})
& 56.60\textsubscript{\textpm3.44}(\textcolor{blue}{7.00})
& 55.50\textsubscript{\textpm8.67}(\textcolor{blue}{5.90})
& 56.40\textsubscript{\textpm1.36}(\textcolor{blue}{6.80})
& 40.00\textsubscript{\textpm0.63}(\textcolor{blue}{9.60})
& 39.60\textsubscript{\textpm2.65}(\textcolor{blue}{10.00})
& 38.20\textsubscript{\textpm3.92}(\textcolor{blue}{11.40})
& 36.40\textsubscript{\textpm1.62}(\textcolor{blue}{13.20})
& 54.00\textsubscript{\textpm2.90}(\textcolor{blue}{4.40})

 \\ 
&
&
MIA
& 22.52\textsubscript{\textpm1.31}
& 2.14\textsubscript{\textpm0.66}(\textcolor{blue}{20.38})
& 19.28\textsubscript{\textpm1.31}(\textcolor{blue}{3.24})
& 26.06\textsubscript{\textpm4.95}(\textcolor{blue}{3.54})
& 22.76\textsubscript{\textpm1.55}(\textcolor{blue}{0.24})
& 5.04\textsubscript{\textpm0.59}(\textcolor{blue}{17.48})
& 0.04\textsubscript{\textpm0.08}(\textcolor{blue}{22.48})
& 0.00\textsubscript{\textpm0.00}(\textcolor{blue}{22.52})
& 0.36\textsubscript{\textpm0.08}(\textcolor{blue}{22.16})
& 20.44\textsubscript{\textpm2.45}(\textcolor{blue}{2.08})

 \\ 
\arrayrulecolor{lightgray} 
\cmidrule[0.75pt]{3-13} \arrayrulecolor{black}

&
&  Avg.Gap$\downarrow$  
 & - 
& \textcolor{blue}{11.45}
& \textcolor{blue}{5.95}
& \textcolor{blue}{3.28}
& \textcolor{blue}{5.19}
& \textcolor{blue}{6.47}
& \textcolor{blue}{6.91}
& \textcolor{blue}{6.86}
& \textcolor{blue}{8.20}
& \textbf{\textcolor{blue}{2.22}}

\\ \midrule

{\multirow{6}{*}{\begin{tabular}[c]{@{}c@{}}Sub-class\\(Sea) \end{tabular}}} 
& {\multirow{6}{*}{\begin{tabular}[c]{@{}c@{}} Practical \end{tabular}}} 
&
TA
& 84.95\textsubscript{\textpm0.24}
& 84.85\textsubscript{\textpm0.04}(\textcolor{blue}{0.10})
& 84.52\textsubscript{\textpm0.08}(\textcolor{blue}{0.43})
& 83.52\textsubscript{\textpm0.08}(\textcolor{blue}{1.44})
& 84.84\textsubscript{\textpm0.20}(\textcolor{blue}{0.11})
& 84.34\textsubscript{\textpm0.02}(\textcolor{blue}{0.62})
& 84.56\textsubscript{\textpm0.06}(\textcolor{blue}{0.39})
& 84.76\textsubscript{\textpm0.12}(\textcolor{blue}{0.19})
& 84.45\textsubscript{\textpm0.06}(\textcolor{blue}{0.50})
& 84.33\textsubscript{\textpm0.26}(\textcolor{blue}{0.62})

 \\ 
&
&
RA
& 99.96\textsubscript{\textpm0.01}
& 99.96\textsubscript{\textpm0.01}(\textcolor{blue}{0.00})
& 99.99\textsubscript{\textpm0.00}(\textcolor{blue}{0.03})
& 99.98\textsubscript{\textpm0.00}(\textcolor{blue}{0.02})
& 99.99\textsubscript{\textpm0.00}(\textcolor{blue}{0.03})
& 99.84\textsubscript{\textpm0.01}(\textcolor{blue}{0.11})
& 99.97\textsubscript{\textpm0.00}(\textcolor{blue}{0.02})
& 99.96\textsubscript{\textpm0.00}(\textcolor{blue}{0.00})
& 99.97\textsubscript{\textpm0.00}(\textcolor{blue}{0.01})
& 98.21\textsubscript{\textpm0.04}(\textcolor{blue}{1.75})

 \\ 
&
&
FATrain
& 79.72\textsubscript{\textpm2.09}
& 99.44\textsubscript{\textpm0.54}(\textcolor{blue}{19.72})
& 92.88\textsubscript{\textpm0.48}(\textcolor{blue}{13.16})
& 88.24\textsubscript{\textpm0.79}(\textcolor{blue}{8.52})
& 71.80\textsubscript{\textpm2.14}(\textcolor{blue}{7.92})
& 62.28\textsubscript{\textpm1.02}(\textcolor{blue}{17.44})
& 63.08\textsubscript{\textpm3.42}(\textcolor{blue}{16.64})
& 85.40\textsubscript{\textpm4.12}(\textcolor{blue}{5.68})
& 69.60\textsubscript{\textpm1.63}(\textcolor{blue}{10.12})
& 77.28\textsubscript{\textpm3.73}(\textcolor{blue}{2.44})

 \\ 
&
&
FATest
& 81.60\textsubscript{\textpm2.15}
& 93.20\textsubscript{\textpm1.33}(\textcolor{blue}{11.60})
& 90.40\textsubscript{\textpm1.36}(\textcolor{blue}{8.80})
& 86.00\textsubscript{\textpm2.10}(\textcolor{blue}{4.40})
& 82.20\textsubscript{\textpm0.75}(\textcolor{blue}{0.60})
& 61.60\textsubscript{\textpm1.85}(\textcolor{blue}{20.00})
& 59.00\textsubscript{\textpm4.00}(\textcolor{blue}{22.60})
& 78.00\textsubscript{\textpm2.00}(\textcolor{blue}{3.60})
& 63.60\textsubscript{\textpm2.65}(\textcolor{blue}{18.00})
& 74.20\textsubscript{\textpm4.17}(\textcolor{blue}{7.40})

 \\ 
&
&
MIA
& 56.28\textsubscript{\textpm2.56}
& 67.92\textsubscript{\textpm14.82}(\textcolor{blue}{11.64})
& 56.40\textsubscript{\textpm2.70}(\textcolor{blue}{0.12})
& 63.88\textsubscript{\textpm2.40}(\textcolor{blue}{7.60})
& 35.56\textsubscript{\textpm1.35}(\textcolor{blue}{20.72})
& 6.64\textsubscript{\textpm0.67}(\textcolor{blue}{49.64})
& 0.24\textsubscript{\textpm0.08}(\textcolor{blue}{56.04})
& 0.00\textsubscript{\textpm0.00}(\textcolor{blue}{56.28})
& 0.04\textsubscript{\textpm0.08}(\textcolor{blue}{56.24})
& 50.36\textsubscript{\textpm5.45}(\textcolor{blue}{5.92})

 \\ 
\arrayrulecolor{lightgray} 
\cmidrule[0.75pt]{3-13} \arrayrulecolor{black}

&
&  Avg.Gap$\downarrow$  
 & - 
& \textcolor{blue}{8.61}
& \textcolor{blue}{4.51}
& \textcolor{blue}{4.40}
& \textcolor{blue}{5.88}
& \textcolor{blue}{17.56}
& \textcolor{blue}{19.14}
& \textcolor{blue}{13.15}
& \textcolor{blue}{16.97}
& \textbf{\textcolor{blue}{3.63}}

\\ \bottomrule[1.5pt]
\end{tabular}
}
\end{table*}
MIA  ratio provides another view to evaluate how an unlearned model is similar to the retrained model  from the entropy of model outputs. MIA attempts to utilize model outputs to determine whether a sample is in the training set. The attack ratio means how many forgetting samples are recognized. Although the MIA used in this paper cannot precisely infer membership, it is widely adopted in recent MU research because it provides an additional perspective to evaluate the output distribution shift after unlearning, which has been discussed in Fig.~\ref{fig:entropy}.
We report the individual results and corresponding gaps with Retrain for each metric in our main tables. One can interpret specific trade-offs and gain a more nuanced understanding of each method’s strengths and limitations.
We also report the average performance gap of above metrics, denoted as \textbf{Avg.Gap}, a metric adopted in many recent works on machine unlearning~\cite{salun,DBLP:conf/nips/HuangCZWHLH24,DBLP:conf/eccv/FanLHL24,DBLP:conf/mm/0006ZS0024}.
\textbf{Avg.Gap} tends to ignore minor differences but highlights significant flaws in specific metrics when they occur. This also aligns with practical evaluation needs—any significant deviation in a metric can render an MU algorithm unusable for real-world applications.

% ---------------------------------------------------------------
% ---------------------------------------------------------------

% Class-wise unlearning

% ---------------------------------------------------------------
% ---------------------------------------------------------------

\subsection{Class-wise Unlearning}
\label{sec:class-wise-unlearning}
To evaluate the unlearning performance to forget an entire class or sub-class, we compare different methods with two different hyperparameter selection ways: 

1) \textbf{ideal} hyperparameters. Following \cite{salun}, ideal hyperparameters mean that we have the model trained without the forgetting class, i.e., the retrained model, to optimize hyperparameters. Results with the ideal hyperparameters can demonstrate the best performance of different methods.

2) \textbf{practical} hyperparameters.  With practical hyperparameters, we transfer the ideal hyperparameters of specific classes to unknown forgetting classes. Since we cannot obtain the retrained model in real-world unlearning applications, the ideal hyperparameters cannot be searched. Thus, we can only transfer hyperparameters of known scenarios to unknown  scenarios in a practical way.

Table~\ref{table:class-wise} shows the performance of different methods under full-class and sub-class unlearning scenarios with three classes. As indicated by the Avg.Gap metric, NatMU achieves the smallest average performance gap in all unlearning tasks with different types of hyperparameters, except that   SCRUB has the best performance when unlearning the sub-class ``Rocket'' with ideal hyperparameters.

In \textbf{full-class} unlearning, all methods can achieve the retrained forgetting accuracy, which is consistently zero. These methods also perform well on the test accuracy metric. 
NatMU's superior performance mainly comes from the smaller gap in MIA ratio, as it does not directly change the output distribution during optimization. 
Additionally, while the baseline methods experience a decline in performance with practical hyperparameters, NatMU maintains a stable low Avg.Gap to the retrained model.

\begin{table*}[t!]
\centering
\caption{Random-sample-wise unlearning results on CIFAR-10 using VGG16-BN, CIFAR-100 using ResNet-18 and TinyImageNet-200 using ResNet-34 and Vision Transformer with different forgetting ratios. The results are given by $\bm{a_{\pm b}(\textcolor{blue}{c})}$, sharing the same format with Table \ref{table:class-wise}.}
\label{table:sample-wise}
\resizebox{\textwidth}{!}{
\begin{tabular}{ccccccccccc}
\toprule[1.5pt]
\multirow{2}{*}{\begin{tabular}[c]{@{}c@{}}Unlearning\\ task\end{tabular}} 
& \multirow{2}{*}{Metric} 
& \multicolumn{9}{c}{Methods} \\ 
\cmidrule[0.75pt]{3-11} 
 
 &  & Retrain  & NegGrad+ &SCRUB & AMUN & $\delta\text{-targeted}^\lambda $ & Amnesiac & BadTeacher & SalUn  &NatMU \\ \midrule[1pt]
%----------------------------------------------------------------------------------
{\multirow{5}{*}{\begin{tabular}[c]{@{}c@{}}CIFAR-10\\ (VGG16-BN, 1\%)\end{tabular}}} & 
TA
& 93.32\textsubscript{\textpm0.22}
& 92.80\textsubscript{\textpm0.19}(\textcolor{blue}{0.52})
& 92.86\textsubscript{\textpm0.16}(\textcolor{blue}{0.46})
& 92.89\textsubscript{\textpm0.11}(\textcolor{blue}{0.43})
& 92.90\textsubscript{\textpm0.04}(\textcolor{blue}{0.42})
& 92.95\textsubscript{\textpm0.09}(\textcolor{blue}{0.37})
& 93.11\textsubscript{\textpm0.06}(\textcolor{blue}{0.21})
& 92.90\textsubscript{\textpm0.08}(\textcolor{blue}{0.41})
& 92.77\textsubscript{\textpm0.19}(\textcolor{blue}{0.55})

\\
&
RA
& 99.95\textsubscript{\textpm0.01}
& 100.00\textsubscript{\textpm0.00}(\textcolor{blue}{0.05})
& 99.99\textsubscript{\textpm0.00}(\textcolor{blue}{0.04})
& 100.00\textsubscript{\textpm0.00}(\textcolor{blue}{0.05})
& 100.00\textsubscript{\textpm0.00}(\textcolor{blue}{0.05})
& 100.00\textsubscript{\textpm0.00}(\textcolor{blue}{0.05})
& 99.93\textsubscript{\textpm0.01}(\textcolor{blue}{0.01})
& 100.00\textsubscript{\textpm0.00}(\textcolor{blue}{0.05})
& 98.95\textsubscript{\textpm0.03}(\textcolor{blue}{0.99})

\\
&
FA
& 94.00\textsubscript{\textpm0.49}
& 93.16\textsubscript{\textpm0.59}(\textcolor{blue}{0.84})
& 93.20\textsubscript{\textpm0.25}(\textcolor{blue}{0.80})
& 92.32\textsubscript{\textpm0.93}(\textcolor{blue}{1.68})
& 93.08\textsubscript{\textpm0.37}(\textcolor{blue}{0.92})
& 89.76\textsubscript{\textpm0.67}(\textcolor{blue}{4.24})
& 91.32\textsubscript{\textpm0.53}(\textcolor{blue}{2.68})
& 91.92\textsubscript{\textpm0.64}(\textcolor{blue}{2.08})
& 96.04\textsubscript{\textpm0.45}(\textcolor{blue}{2.04})

\\
&
MIA
& 83.54\textsubscript{\textpm1.32}
& 90.84\textsubscript{\textpm1.17}(\textcolor{blue}{7.30})
& 75.88\textsubscript{\textpm0.91}(\textcolor{blue}{7.66})
& 70.36\textsubscript{\textpm1.44}(\textcolor{blue}{13.18})
& 60.08\textsubscript{\textpm1.58}(\textcolor{blue}{23.46})
& 33.72\textsubscript{\textpm0.97}(\textcolor{blue}{49.82})
& 51.48\textsubscript{\textpm0.57}(\textcolor{blue}{32.06})
& 62.32\textsubscript{\textpm0.95}(\textcolor{blue}{21.22})
& 80.92\textsubscript{\textpm0.48}(\textcolor{blue}{2.62})

\\
\arrayrulecolor{lightgray} 
\cmidrule[0.75pt]{2-11} \arrayrulecolor{black}

& Avg.Gap$\downarrow$ 
 & - 
& \textcolor{blue}{2.18}
& \textcolor{blue}{2.24}
& \textcolor{blue}{3.84}
& \textcolor{blue}{6.21}
& \textcolor{blue}{13.62}
& \textcolor{blue}{8.74}
& \textcolor{blue}{5.94}
& \textbf{\textcolor{blue}{1.55}}

\\ \midrule
%----------------------------------------------------------------------------------
{\multirow{5}{*}{\begin{tabular}[c]{@{}c@{}}CIFAR-10\\ (VGG16-BN, 10\%)\end{tabular}}} & 
TA
& 93.12\textsubscript{\textpm0.17}
& 92.35\textsubscript{\textpm0.08}(\textcolor{blue}{0.77})
& 92.79\textsubscript{\textpm0.13}(\textcolor{blue}{0.32})
& 92.71\textsubscript{\textpm0.07}(\textcolor{blue}{0.41})
& 92.78\textsubscript{\textpm0.05}(\textcolor{blue}{0.34})
& 92.92\textsubscript{\textpm0.08}(\textcolor{blue}{0.20})
& 92.87\textsubscript{\textpm0.06}(\textcolor{blue}{0.25})
& 92.11\textsubscript{\textpm0.08}(\textcolor{blue}{1.00})
& 92.40\textsubscript{\textpm0.13}(\textcolor{blue}{0.72})

\\
&
RA
& 99.94\textsubscript{\textpm0.01}
& 100.00\textsubscript{\textpm0.00}(\textcolor{blue}{0.05})
& 99.99\textsubscript{\textpm0.00}(\textcolor{blue}{0.04})
& 100.00\textsubscript{\textpm0.00}(\textcolor{blue}{0.06})
& 99.94\textsubscript{\textpm0.00}(\textcolor{blue}{0.01})
& 100.00\textsubscript{\textpm0.00}(\textcolor{blue}{0.06})
& 99.94\textsubscript{\textpm0.01}(\textcolor{blue}{0.00})
& 99.97\textsubscript{\textpm0.01}(\textcolor{blue}{0.03})
& 98.93\textsubscript{\textpm0.09}(\textcolor{blue}{1.01})

\\
&
FA
& 93.20\textsubscript{\textpm0.18}
& 92.12\textsubscript{\textpm0.23}(\textcolor{blue}{1.09})
& 98.53\textsubscript{\textpm0.09}(\textcolor{blue}{5.32})
& 97.14\textsubscript{\textpm0.13}(\textcolor{blue}{3.94})
& 98.62\textsubscript{\textpm0.03}(\textcolor{blue}{5.42})
& 97.30\textsubscript{\textpm0.26}(\textcolor{blue}{4.10})
& 93.88\textsubscript{\textpm0.25}(\textcolor{blue}{0.67})
& 97.02\textsubscript{\textpm0.17}(\textcolor{blue}{3.82})
& 94.76\textsubscript{\textpm0.17}(\textcolor{blue}{1.56})

\\
&
MIA
& 80.80\textsubscript{\textpm0.12}
& 91.61\textsubscript{\textpm1.28}(\textcolor{blue}{10.81})
& 83.56\textsubscript{\textpm0.28}(\textcolor{blue}{2.76})
& 76.51\textsubscript{\textpm0.67}(\textcolor{blue}{4.29})
& 53.93\textsubscript{\textpm0.59}(\textcolor{blue}{26.87})
& 42.78\textsubscript{\textpm0.74}(\textcolor{blue}{38.02})
& 52.26\textsubscript{\textpm0.42}(\textcolor{blue}{28.54})
& 51.82\textsubscript{\textpm0.53}(\textcolor{blue}{28.98})
& 78.32\textsubscript{\textpm0.21}(\textcolor{blue}{2.48})

\\
\arrayrulecolor{lightgray} 
\cmidrule[0.75pt]{2-11} \arrayrulecolor{black}

& Avg.Gap$\downarrow$ 
 & - 
& \textcolor{blue}{3.18}
& \textcolor{blue}{2.11}
& \textcolor{blue}{2.17}
& \textcolor{blue}{8.16}
& \textcolor{blue}{10.59}
& \textcolor{blue}{7.37}
& \textcolor{blue}{8.46}
& \textbf{\textcolor{blue}{1.44}}
 
\\ \midrule[1.5pt]
%----------------------------------------------------------------------------------
{\multirow{5}{*}{\begin{tabular}[c]{@{}c@{}}CIFAR-100\\ (ResNet-18, 1\%)\end{tabular}}} & 
TA
& 76.68\textsubscript{\textpm0.16}
& 76.45\textsubscript{\textpm0.25}(\textcolor{blue}{0.24})
& 76.42\textsubscript{\textpm0.08}(\textcolor{blue}{0.27})
& 75.10\textsubscript{\textpm0.05}(\textcolor{blue}{1.58})
& 75.56\textsubscript{\textpm0.07}(\textcolor{blue}{1.12})
& 75.24\textsubscript{\textpm0.19}(\textcolor{blue}{1.44})
& 77.80\textsubscript{\textpm0.13}(\textcolor{blue}{1.11})
& 75.39\textsubscript{\textpm0.14}(\textcolor{blue}{1.30})
& 75.53\textsubscript{\textpm0.13}(\textcolor{blue}{1.15})

\\
&
RA
& 99.94\textsubscript{\textpm0.01}
& 99.98\textsubscript{\textpm0.00}(\textcolor{blue}{0.04})
& 99.97\textsubscript{\textpm0.00}(\textcolor{blue}{0.03})
& 99.97\textsubscript{\textpm0.00}(\textcolor{blue}{0.03})
& 99.96\textsubscript{\textpm0.00}(\textcolor{blue}{0.02})
& 99.98\textsubscript{\textpm0.00}(\textcolor{blue}{0.04})
& 99.88\textsubscript{\textpm0.00}(\textcolor{blue}{0.06})
& 99.98\textsubscript{\textpm0.00}(\textcolor{blue}{0.04})
& 97.47\textsubscript{\textpm0.05}(\textcolor{blue}{2.47})

\\
&
FA
& 77.60\textsubscript{\textpm1.01}
& 78.08\textsubscript{\textpm0.68}(\textcolor{blue}{0.48})
& 75.12\textsubscript{\textpm0.57}(\textcolor{blue}{2.48})
& 77.00\textsubscript{\textpm0.59}(\textcolor{blue}{0.60})
& 78.56\textsubscript{\textpm0.59}(\textcolor{blue}{0.96})
& 75.08\textsubscript{\textpm1.22}(\textcolor{blue}{2.52})
& 70.28\textsubscript{\textpm1.78}(\textcolor{blue}{7.32})
& 76.76\textsubscript{\textpm1.65}(\textcolor{blue}{0.84})
& 78.88\textsubscript{\textpm1.43}(\textcolor{blue}{1.28})

\\
&
MIA
& 55.40\textsubscript{\textpm1.23}
& 74.56\textsubscript{\textpm1.55}(\textcolor{blue}{19.16})
& 78.40\textsubscript{\textpm1.23}(\textcolor{blue}{23.00})
& 44.00\textsubscript{\textpm1.74}(\textcolor{blue}{11.40})
& 26.96\textsubscript{\textpm0.81}(\textcolor{blue}{28.44})
& 5.52\textsubscript{\textpm0.61}(\textcolor{blue}{49.88})
& 0.72\textsubscript{\textpm0.32}(\textcolor{blue}{54.68})
& 5.12\textsubscript{\textpm0.55}(\textcolor{blue}{50.28})
& 53.52\textsubscript{\textpm1.01}(\textcolor{blue}{1.88})

\\
\arrayrulecolor{lightgray} 
\cmidrule[0.75pt]{2-11} \arrayrulecolor{black}

& Avg.Gap$\downarrow$ 
 & - 
& \textcolor{blue}{4.98}
& \textcolor{blue}{6.44}
& \textcolor{blue}{3.40}
& \textcolor{blue}{7.64}
& \textcolor{blue}{13.47}
& \textcolor{blue}{15.79}
& \textcolor{blue}{13.11}
& \textbf{\textcolor{blue}{1.70}}

\\ \midrule

%----------------------------------------------------------------------------------
{\multirow{5}{*}{\begin{tabular}[c]{@{}c@{}}CIFAR-100\\ (ResNet-18, 10\%)\end{tabular}}} & 
TA
& 75.43\textsubscript{\textpm0.47}
& 73.86\textsubscript{\textpm0.06}(\textcolor{blue}{1.57})
& 75.02\textsubscript{\textpm0.14}(\textcolor{blue}{0.41})
& 74.29\textsubscript{\textpm0.18}(\textcolor{blue}{1.15})
& 72.28\textsubscript{\textpm0.11}(\textcolor{blue}{3.16})
& 71.23\textsubscript{\textpm0.13}(\textcolor{blue}{4.21})
& 76.97\textsubscript{\textpm0.20}(\textcolor{blue}{1.54})
& 71.57\textsubscript{\textpm0.15}(\textcolor{blue}{3.86})
& 74.27\textsubscript{\textpm0.12}(\textcolor{blue}{1.16})

\\
&
RA
& 99.95\textsubscript{\textpm0.01}
& 99.98\textsubscript{\textpm0.00}(\textcolor{blue}{0.03})
& 99.97\textsubscript{\textpm0.00}(\textcolor{blue}{0.02})
& 99.97\textsubscript{\textpm0.00}(\textcolor{blue}{0.03})
& 99.61\textsubscript{\textpm0.02}(\textcolor{blue}{0.34})
& 99.92\textsubscript{\textpm0.00}(\textcolor{blue}{0.03})
& 99.88\textsubscript{\textpm0.01}(\textcolor{blue}{0.06})
& 99.88\textsubscript{\textpm0.01}(\textcolor{blue}{0.07})
& 96.88\textsubscript{\textpm0.08}(\textcolor{blue}{3.07})

\\
&
FA
& 75.92\textsubscript{\textpm0.17}
& 73.99\textsubscript{\textpm0.29}(\textcolor{blue}{1.93})
& 75.20\textsubscript{\textpm0.33}(\textcolor{blue}{0.72})
& 92.19\textsubscript{\textpm0.20}(\textcolor{blue}{16.28})
& 92.61\textsubscript{\textpm0.32}(\textcolor{blue}{16.70})
& 94.90\textsubscript{\textpm0.31}(\textcolor{blue}{18.98})
& 78.08\textsubscript{\textpm0.43}(\textcolor{blue}{2.16})
& 94.30\textsubscript{\textpm0.22}(\textcolor{blue}{18.38})
& 76.73\textsubscript{\textpm0.46}(\textcolor{blue}{0.82})

\\
&
MIA
& 53.49\textsubscript{\textpm0.51}
& 72.76\textsubscript{\textpm0.52}(\textcolor{blue}{19.27})
& 61.45\textsubscript{\textpm0.57}(\textcolor{blue}{7.96})
& 53.42\textsubscript{\textpm0.53}(\textcolor{blue}{0.06})
& 50.42\textsubscript{\textpm0.41}(\textcolor{blue}{3.07})
& 20.66\textsubscript{\textpm0.22}(\textcolor{blue}{32.82})
& 0.60\textsubscript{\textpm0.07}(\textcolor{blue}{52.89})
& 20.69\textsubscript{\textpm0.20}(\textcolor{blue}{32.80})
& 51.45\textsubscript{\textpm0.65}(\textcolor{blue}{2.04})

\\
\arrayrulecolor{lightgray} 
\cmidrule[0.75pt]{2-11} \arrayrulecolor{black}

& Avg.Gap$\downarrow$ 
 & - 
& \textcolor{blue}{5.70}
& \textcolor{blue}{2.28}
& \textcolor{blue}{4.38}
& \textcolor{blue}{5.81}
& \textcolor{blue}{14.01}
& \textcolor{blue}{14.16}
& \textcolor{blue}{13.78}
& \textbf{\textcolor{blue}{1.77}}

\\ \midrule[1.5pt]
%----------------------------------------------------------------------------------
{\multirow{5}{*}{\begin{tabular}[c]{@{}c@{}}TinyImageNet\\ (ResNet-34, 1\%)\end{tabular}}} & 
TA
& 66.55\textsubscript{\textpm0.22}
& 66.07\textsubscript{\textpm0.13}(\textcolor{blue}{0.47})
& 66.27\textsubscript{\textpm0.18}(\textcolor{blue}{0.28})
& 65.21\textsubscript{\textpm0.11}(\textcolor{blue}{1.34})
& 65.35\textsubscript{\textpm0.11}(\textcolor{blue}{1.19})
& 64.87\textsubscript{\textpm0.15}(\textcolor{blue}{1.68})
& 67.66\textsubscript{\textpm0.16}(\textcolor{blue}{1.11})
& 64.72\textsubscript{\textpm0.07}(\textcolor{blue}{1.83})
& 66.06\textsubscript{\textpm0.13}(\textcolor{blue}{0.49})

\\
&
RA
& 99.97\textsubscript{\textpm0.00}
& 99.98\textsubscript{\textpm0.00}(\textcolor{blue}{0.01})
& 99.97\textsubscript{\textpm0.00}(\textcolor{blue}{0.01})
& 99.94\textsubscript{\textpm0.00}(\textcolor{blue}{0.04})
& 99.98\textsubscript{\textpm0.00}(\textcolor{blue}{0.00})
& 99.98\textsubscript{\textpm0.00}(\textcolor{blue}{0.00})
& 99.93\textsubscript{\textpm0.01}(\textcolor{blue}{0.04})
& 99.98\textsubscript{\textpm0.00}(\textcolor{blue}{0.00})
& 98.40\textsubscript{\textpm0.02}(\textcolor{blue}{1.57})

\\
&
FA
& 67.06\textsubscript{\textpm0.79}
& 67.56\textsubscript{\textpm0.79}(\textcolor{blue}{0.50})
& 54.54\textsubscript{\textpm0.60}(\textcolor{blue}{12.52})
& 66.14\textsubscript{\textpm0.97}(\textcolor{blue}{0.92})
& 50.32\textsubscript{\textpm0.74}(\textcolor{blue}{16.74})
& 64.06\textsubscript{\textpm0.83}(\textcolor{blue}{3.00})
& 63.82\textsubscript{\textpm0.55}(\textcolor{blue}{3.24})
& 64.82\textsubscript{\textpm1.04}(\textcolor{blue}{2.24})
& 67.60\textsubscript{\textpm0.54}(\textcolor{blue}{0.54})

\\
&
MIA
& 44.54\textsubscript{\textpm1.75}
& 62.22\textsubscript{\textpm0.70}(\textcolor{blue}{17.68})
& 62.68\textsubscript{\textpm1.12}(\textcolor{blue}{18.14})
& 30.12\textsubscript{\textpm0.90}(\textcolor{blue}{14.42})
& 8.36\textsubscript{\textpm0.22}(\textcolor{blue}{36.18})
& 0.36\textsubscript{\textpm0.10}(\textcolor{blue}{44.18})
& 0.00\textsubscript{\textpm0.00}(\textcolor{blue}{44.54})
& 0.37\textsubscript{\textpm0.05}(\textcolor{blue}{44.17})
& 38.42\textsubscript{\textpm0.60}(\textcolor{blue}{6.12})

\\
\arrayrulecolor{lightgray} 
\cmidrule[0.75pt]{2-11} \arrayrulecolor{black}

& Avg.Gap$\downarrow$ 
 & - 
& \textcolor{blue}{4.66}
& \textcolor{blue}{7.74}
& \textcolor{blue}{4.18}
& \textcolor{blue}{13.53}
& \textcolor{blue}{12.22}
& \textcolor{blue}{12.23}
& \textcolor{blue}{12.06}
& \textbf{\textcolor{blue}{2.18}}

\\ \midrule
%----------------------------------------------------------------------------------
{\multirow{5}{*}{\begin{tabular}[c]{@{}c@{}}TinyImageNet\\ (ResNet-34, 10\%)\end{tabular}}} & 
TA
& 65.45\textsubscript{\textpm0.22}
& 62.91\textsubscript{\textpm0.15}(\textcolor{blue}{2.54})
& 65.27\textsubscript{\textpm0.65}(\textcolor{blue}{0.18})
& 64.59\textsubscript{\textpm0.28}(\textcolor{blue}{0.86})
& 62.33\textsubscript{\textpm0.12}(\textcolor{blue}{3.12})
& 61.21\textsubscript{\textpm0.11}(\textcolor{blue}{4.24})
& 66.06\textsubscript{\textpm0.08}(\textcolor{blue}{0.61})
& 60.65\textsubscript{\textpm0.11}(\textcolor{blue}{4.80})
& 64.69\textsubscript{\textpm0.17}(\textcolor{blue}{0.76})

\\
&
RA
& 99.97\textsubscript{\textpm0.00}
& 99.98\textsubscript{\textpm0.00}(\textcolor{blue}{0.01})
& 98.55\textsubscript{\textpm2.30}(\textcolor{blue}{1.43})
& 99.88\textsubscript{\textpm0.01}(\textcolor{blue}{0.10})
& 99.95\textsubscript{\textpm0.00}(\textcolor{blue}{0.02})
& 99.98\textsubscript{\textpm0.00}(\textcolor{blue}{0.00})
& 99.93\textsubscript{\textpm0.00}(\textcolor{blue}{0.05})
& 99.97\textsubscript{\textpm0.00}(\textcolor{blue}{0.00})
& 97.23\textsubscript{\textpm0.03}(\textcolor{blue}{2.74})

\\
&
FA
& 65.29\textsubscript{\textpm0.31}
& 62.06\textsubscript{\textpm0.20}(\textcolor{blue}{3.23})
& 77.47\textsubscript{\textpm0.65}(\textcolor{blue}{12.18})
& 80.95\textsubscript{\textpm0.24}(\textcolor{blue}{15.66})
& 92.86\textsubscript{\textpm0.08}(\textcolor{blue}{27.57})
& 95.81\textsubscript{\textpm0.16}(\textcolor{blue}{30.52})
& 67.32\textsubscript{\textpm0.25}(\textcolor{blue}{2.03})
& 93.66\textsubscript{\textpm0.22}(\textcolor{blue}{28.37})
& 66.16\textsubscript{\textpm0.42}(\textcolor{blue}{0.87})

\\
&
MIA
& 41.94\textsubscript{\textpm0.84}
& 61.07\textsubscript{\textpm0.63}(\textcolor{blue}{19.13})
& 48.69\textsubscript{\textpm3.21}(\textcolor{blue}{6.74})
& 37.71\textsubscript{\textpm0.38}(\textcolor{blue}{4.23})
& 45.80\textsubscript{\textpm0.18}(\textcolor{blue}{3.86})
& 7.83\textsubscript{\textpm0.19}(\textcolor{blue}{34.11})
& 0.00\textsubscript{\textpm0.00}(\textcolor{blue}{41.94})
& 5.95\textsubscript{\textpm0.13}(\textcolor{blue}{35.99})
& 38.38\textsubscript{\textpm0.23}(\textcolor{blue}{3.56})

\\
\arrayrulecolor{lightgray} 
\cmidrule[0.75pt]{2-11} \arrayrulecolor{black}

& Avg.Gap$\downarrow$ 
 & - 
& \textcolor{blue}{6.23}
& \textcolor{blue}{5.13}
& \textcolor{blue}{5.21}
& \textcolor{blue}{8.64}
& \textcolor{blue}{17.22}
& \textcolor{blue}{11.16}
& \textcolor{blue}{17.29}
& \textbf{\textcolor{blue}{1.98}}

\\ \midrule

%----------------------------------------------------------------------------------

{\multirow{5}{*}{\begin{tabular}[c]{@{}c@{}}TinyImageNet\\ (ViT, 1\%)\end{tabular}}} & 
 TA
&  70.42\textsubscript{\textpm0.15}
&  68.06\textsubscript{\textpm0.12}(\textcolor{blue}{2.36})
&  68.73\textsubscript{\textpm0.22}(\textcolor{blue}{1.69})
&  69.66\textsubscript{\textpm0.24}(\textcolor{blue}{0.75})
&  67.06\textsubscript{\textpm0.06}(\textcolor{blue}{3.36})
&  65.54\textsubscript{\textpm0.17}(\textcolor{blue}{4.88})
&  70.94\textsubscript{\textpm0.06}(\textcolor{blue}{0.52})
&  65.30\textsubscript{\textpm0.18}(\textcolor{blue}{5.12})
&  68.11\textsubscript{\textpm0.20}(\textcolor{blue}{2.30})

\\
&
 RA
&  99.98\textsubscript{\textpm0.00}
&  99.98\textsubscript{\textpm0.01}(\textcolor{blue}{0.00})
&  99.94\textsubscript{\textpm0.01}(\textcolor{blue}{0.04})
&  99.97\textsubscript{\textpm0.00}(\textcolor{blue}{0.01})
&  99.94\textsubscript{\textpm0.08}(\textcolor{blue}{0.04})
&  99.98\textsubscript{\textpm0.00}(\textcolor{blue}{0.00})
&  99.98\textsubscript{\textpm0.00}(\textcolor{blue}{0.00})
&  99.98\textsubscript{\textpm0.00}(\textcolor{blue}{0.00})
&  99.95\textsubscript{\textpm0.01}(\textcolor{blue}{0.03})

\\
&
 FA
&  71.46\textsubscript{\textpm0.44}
&  68.24\textsubscript{\textpm1.11}(\textcolor{blue}{3.22})
&  86.32\textsubscript{\textpm0.85}(\textcolor{blue}{14.86})
&  67.00\textsubscript{\textpm0.59}(\textcolor{blue}{4.46})
&  72.26\textsubscript{\textpm0.31}(\textcolor{blue}{0.80})
&  8.22\textsubscript{\textpm1.05}(\textcolor{blue}{63.24})
&  74.24\textsubscript{\textpm0.66}(\textcolor{blue}{2.78})
&  10.20\textsubscript{\textpm0.74}(\textcolor{blue}{61.26})
&  74.16\textsubscript{\textpm1.08}(\textcolor{blue}{2.70})

\\
&
 MIA
&  55.16\textsubscript{\textpm0.39}
&  73.20\textsubscript{\textpm1.04}(\textcolor{blue}{18.04})
&  90.74\textsubscript{\textpm0.75}(\textcolor{blue}{35.58})
&  30.52\textsubscript{\textpm0.69}(\textcolor{blue}{24.64})
&  33.90\textsubscript{\textpm0.27}(\textcolor{blue}{21.26})
&  2.68\textsubscript{\textpm0.28}(\textcolor{blue}{52.48})
&  0.00\textsubscript{\textpm0.00}(\textcolor{blue}{55.16})
&  2.66\textsubscript{\textpm0.42}(\textcolor{blue}{52.50})
&  45.50\textsubscript{\textpm0.44}(\textcolor{blue}{9.66})

\\
\arrayrulecolor{lightgray} 
\cmidrule[0.75pt]{2-11} \arrayrulecolor{black}

&  Avg.Gap$\downarrow$ 
 &  - 
& \textcolor{blue}{5.91}
& \textcolor{blue}{13.04}
& \textcolor{blue}{7.47}
& \textcolor{blue}{6.37}
& \textcolor{blue}{30.15}
& \textcolor{blue}{14.62}
& \textcolor{blue}{29.72}
& \textcolor{blue}{\textbf{3.67}}

\\ \midrule

%----------------------------------------------------------------------------------

{\multirow{5}{*}{\begin{tabular}[c]{@{}c@{}} TinyImageNet\\  (ViT, 10\%)\end{tabular}}} & 
 TA
&  70.28\textsubscript{\textpm0.31}
&  64.71\textsubscript{\textpm0.26}(\textcolor{blue}{5.57})
&  66.27\textsubscript{\textpm0.95}(\textcolor{blue}{4.01})
&  69.81\textsubscript{\textpm0.08}(\textcolor{blue}{0.47})
&  65.36\textsubscript{\textpm0.12}(\textcolor{blue}{4.92})
&  66.58\textsubscript{\textpm0.21}(\textcolor{blue}{3.70})
&  69.52\textsubscript{\textpm0.08}(\textcolor{blue}{0.76})
&  66.38\textsubscript{\textpm0.20}(\textcolor{blue}{3.90})
&  66.31\textsubscript{\textpm0.13}(\textcolor{blue}{3.97})

\\
&
 RA
&  99.98\textsubscript{\textpm0.00}
&  99.98\textsubscript{\textpm0.00}(\textcolor{blue}{0.00})
&  99.73\textsubscript{\textpm0.48}(\textcolor{blue}{0.25})
&  99.97\textsubscript{\textpm0.00}(\textcolor{blue}{0.01})
&  99.18\textsubscript{\textpm0.01}(\textcolor{blue}{0.80})
&  99.97\textsubscript{\textpm0.00}(\textcolor{blue}{0.01})
&  99.98\textsubscript{\textpm0.00}(\textcolor{blue}{0.00})
&  99.97\textsubscript{\textpm0.00}(\textcolor{blue}{0.01})
&  99.89\textsubscript{\textpm0.02}(\textcolor{blue}{0.09})

\\
&
 FA
&  69.98\textsubscript{\textpm0.21}
&  63.78\textsubscript{\textpm0.30}(\textcolor{blue}{6.20})
&  88.71\textsubscript{\textpm1.24}(\textcolor{blue}{18.73})
&  88.80\textsubscript{\textpm0.32}(\textcolor{blue}{18.81})
&  97.68\textsubscript{\textpm0.04}(\textcolor{blue}{27.69})
&  97.86\textsubscript{\textpm0.11}(\textcolor{blue}{27.88})
&  74.25\textsubscript{\textpm0.43}(\textcolor{blue}{4.27})
&  97.87\textsubscript{\textpm0.09}(\textcolor{blue}{27.89})
&  71.87\textsubscript{\textpm0.29}(\textcolor{blue}{1.88})

\\
&
 MIA
&  51.87\textsubscript{\textpm0.36}
&  75.15\textsubscript{\textpm0.73}(\textcolor{blue}{23.28})
&  73.93\textsubscript{\textpm1.54}(\textcolor{blue}{22.06})
&  47.68\textsubscript{\textpm1.04}(\textcolor{blue}{4.18})
&  72.70\textsubscript{\textpm0.16}(\textcolor{blue}{20.83})
&  26.60\textsubscript{\textpm0.49}(\textcolor{blue}{25.27})
&  0.28\textsubscript{\textpm0.03}(\textcolor{blue}{51.59})
&  26.87\textsubscript{\textpm0.33}(\textcolor{blue}{24.99})
&  48.38\textsubscript{\textpm0.49}(\textcolor{blue}{3.48})

\\
\arrayrulecolor{lightgray} 
\cmidrule[0.75pt]{2-11} \arrayrulecolor{black}

&  Avg.Gap$\downarrow$ 
 &  - 
& \textcolor{blue}{8.76}
& \textcolor{blue}{11.26}
& \textcolor{blue}{5.87}
& \textcolor{blue}{13.56}
& \textcolor{blue}{14.21}
& \textcolor{blue}{14.15}
& \textcolor{blue}{14.20}
& \textcolor{blue}{\textbf{2.36}}

\\ \bottomrule[1.5pt]
\end{tabular}
}
\end{table*}

\textbf{Sub-class} unlearning presents greater challenges than full-class unlearning, as the forgetting accuracy can significantly vary depending on the specific forgetting sub-class. 

With ideal hyperparameters, the three relabeling-based methods demonstrate reasonable performance. But with practical hyperparameters, the hard-label methods (Amnesiac and SalUn) suffer the over-forgetting problem. All the three methods also have an extremely low MIA ratio, nearing zero, while NatMU  exhibits a smaller MIA ratio gap. Such low MIA ratio would lead to a ``Streisand Effect'', where the forgetting samples are actually more noticeable~\cite{fisherforgetting}.
We notice that NatMU has a bigger remaining accuracy gap than other method, about 2\%, since when introducing correct information of remaining samples into unlearning instances, we also apply masking operation to the remaining samples, which may affect their learning. 
However, compared to other metrics, a slight change in remaining accuracy is acceptable.

% ---------------------------------------------------------------
% ---------------------------------------------------------------

% Sample-wise unlearning

% ---------------------------------------------------------------
% ---------------------------------------------------------------

\subsection{Sample-wise Unlearning}
In sample-wise unlearning, we again evaluate NatMU and other methods under two experimental settings along with different hyperparameter selection ways:

1) \textbf{$\mathcal{D}^f$ from a known distribution.} In this setting, the forgetting samples are randomly selected from the training set. As a result, the distribution of the forgetting set aligns with the test distribution, which can help to determine hyperparameters without the retrained model. This is a widely used but relatively idealized setting, as we have the prior knowledge about the forgetting distribution. We denote this setting as \textbf{random-sample-wise unlearning}.

2) \textbf{$\mathcal{D}^f$ from an unknown distribution.} 
This setting is more practical but also more challenging, since the hyperparameters cannot be directly tuned due to the lack of prior knowledge. To address this, we propose transferring hyperparameters optimized in setting (1) to this scenario. To evaluate the unlearning performance under such setting, we construct the forgetting set using the most difficult-to-learn samples to conduct \textbf{difficult-sample-wise unlearning}, to simulate the potential worst-case scenario.

\subsubsection{Random-Sample-Wise Unlearning}\label{random-sample-wise-unlearning}

Following the experimental setup in \cite{badteacher, salun, ssd, sparsity, scrub}, we uniformly select a subset of training samples as forgetting samples to evaluate the sample-wise unlearning performance. In this context, the forgetting samples and test samples are assumed to share the same distribution. Consequently, a subset of test samples can be randomly selected as proxy forgetting samples to approximate the model's performance on the actual forgetting samples. This proxy unlearning performance provides a reference  and can be further utilized for hyperparameter tuning, without requiring access to the retrained model.

To ensure we can get robust hyperparameters, the hyperparameter search process is conducted across various forgetting ratios. 
Specifically, we search a set of hyperparameters, with which the unlearning method can approximate the proxy performance with both 1\% and 10\%  forgetting ratio.
This also accounts for potential shifts in the distribution of forgetting samples, which will be discussed in Section~\ref{sec:difficult-sample-unlearning}.

\textbf{Results.} As shown in Table~\ref{table:sample-wise}, NatMU maintains the best performance across multiple datasets and forgetting ratios. 
Since the forgetting samples with wrong hard labels introduce massive incorrect information, Amnesiac and SalUn represent a significant test accuracy drop with complex datasets and large forgetting ratios. Moreover, they cannot approach the forgetting accuracy with different forgetting ratios. With 1\% forgetting ratio, their FA is lower; while with 10\% forgetting ratio, their FA is much higher than the retrained model. The phenomenon is also observed in SCRUB. 
The soft label with knowledge distillation framework helps BadTeacher maintain a good model utility (TA) and unlearning effectiveness (FA),  it would lead to an extremely low MIA ratio. In contrast, NegGrad+ has a much higher MIA ratio. NatMU is the only method capable of achieving a balance across various metrics and achieves the smallest performance gap.
One also observe that he unlearning of ViT models is harder than that of CNNs, as observed in~\cite{learningtounlearn, DBLP:conf/nips/HuangCZWHLH24, DBLP:conf/mm/0006ZS0024}. This may be attributed to the strong fitting capacity of ViT models~\cite{DBLP:conf/nips/LiuSBSLN21, app13095521, DBLP:journals/corr/abs-2206-00389}, which makes them more resistant to forgetting.

\subsubsection{Difficult-Sample-Wise Unlearning}\label{sec:difficult-sample-unlearning}

However, in practical applications, assuming that the distribution of forgetting samples is similar to test distribution is overly idealistic.
A more likely scenario is that forgetting requests originate from a specific group of users, resulting in forgetting samples belonging to a certain unknown distribution.
As a result, the proxy-based approach above cannot be directly applied to determine hyperparameters. To address this challenge, we propose transferring the hyperparameters optimized under idealized conditions to real-world application scenarios. This hyperparameter transfer relies on the unlearning robustness across different unlearning tasks, providing a way to determine hyperparameters without requiring the retrained model and any prior knowledge about the forgetting set.

\begin{table}[h]
\centering
\caption{Difficult-sample-wise unlearning results using ResNet-18 on CIFAR-100 with a forgetting ratio of 10\%. The results are given by $\bm{a_{\pm b}(\textcolor{blue}{c})}$, sharing the same format with Table \ref{table:class-wise}.}
\label{table:hardsample-cifar100-10}
\resizebox{0.48\textwidth}{!}{
\begin{tabular}{ccccccc}
\toprule[1.5pt]
Method & TA& RA& FA& MIA & Avg.Gap$\downarrow$   \\
\midrule
Retrain
& 75.28\textsubscript{\textpm0.16}
& 99.99\textsubscript{\textpm0.00}
& 6.40\textsubscript{\textpm0.18}
& 14.74\textsubscript{\textpm0.32}
& - \\  \midrule

\multirow{1}{*}{NegGrad+} 
& \begin{tabular}[c]{@{}c@{}}70.97\textsubscript{\textpm0.24}\\ (\textcolor{blue}{4.31})\end{tabular}
& \begin{tabular}[c]{@{}c@{}}99.96\textsubscript{\textpm0.01}\\ (\textcolor{blue}{0.03})\end{tabular}
& \begin{tabular}[c]{@{}c@{}}7.75\textsubscript{\textpm0.31}\\ (\textcolor{blue}{1.35})\end{tabular}
& \begin{tabular}[c]{@{}c@{}}70.74\textsubscript{\textpm0.73}\\ (\textcolor{blue}{56.00})\end{tabular}
& \multirow{1}{*}{\textcolor{blue}{15.42}}
\\ \midrule

\multirow{1}{*}{SCRUB} 
& \begin{tabular}[c]{@{}c@{}}73.90\textsubscript{\textpm0.10}\\ (\textcolor{blue}{1.38})\end{tabular}
& \begin{tabular}[c]{@{}c@{}}99.98\textsubscript{\textpm0.01}\\ (\textcolor{blue}{0.01})\end{tabular}
& \begin{tabular}[c]{@{}c@{}}30.80\textsubscript{\textpm0.22}\\ (\textcolor{blue}{24.40})\end{tabular}
& \begin{tabular}[c]{@{}c@{}}31.54\textsubscript{\textpm0.35}\\ (\textcolor{blue}{16.80})\end{tabular}
& \multirow{1}{*}{\textcolor{blue}{10.65}}
 \\ \midrule

\multirow{1}{*}{AMUN} 
& \begin{tabular}[c]{@{}c@{}}75.28\textsubscript{\textpm0.11}\\ (\textcolor{blue}{0.00})\end{tabular}
& \begin{tabular}[c]{@{}c@{}}100.00\textsubscript{\textpm0.00}\\ (\textcolor{blue}{0.01})\end{tabular}
& \begin{tabular}[c]{@{}c@{}}75.43\textsubscript{\textpm0.54}\\ (\textcolor{blue}{69.04})\end{tabular}
& \begin{tabular}[c]{@{}c@{}}5.92\textsubscript{\textpm0.17}\\ (\textcolor{blue}{8.83})\end{tabular}
& \multirow{1}{*}{\textcolor{blue}{19.47}}
\\ \midrule

 \multirow{1}{*}{$\delta\text{-targeted}^\lambda$} 
& \begin{tabular}[c]{@{}c@{}}75.00\textsubscript{\textpm0.05}\\ (\textcolor{blue}{0.28})\end{tabular}
& \begin{tabular}[c]{@{}c@{}}99.82\textsubscript{\textpm0.01}\\ (\textcolor{blue}{0.17})\end{tabular}
& \begin{tabular}[c]{@{}c@{}}78.33\textsubscript{\textpm0.20}\\ (\textcolor{blue}{71.94})\end{tabular}
& \begin{tabular}[c]{@{}c@{}}7.04\textsubscript{\textpm0.16}\\ (\textcolor{blue}{7.71})\end{tabular}
& \multirow{1}{*}{\textcolor{blue}{20.03}}
 \\ \midrule

\multirow{1}{*}{Amnesiac} 
& \begin{tabular}[c]{@{}c@{}}73.56\textsubscript{\textpm0.10}\\ (\textcolor{blue}{1.72})\end{tabular}
& \begin{tabular}[c]{@{}c@{}}99.97\textsubscript{\textpm0.00}\\ (\textcolor{blue}{0.02})\end{tabular}
& \begin{tabular}[c]{@{}c@{}}83.85\textsubscript{\textpm0.16}\\ (\textcolor{blue}{77.46})\end{tabular}
& \begin{tabular}[c]{@{}c@{}}0.94\textsubscript{\textpm0.09}\\ (\textcolor{blue}{13.80})\end{tabular}
& \multirow{1}{*}{\textcolor{blue}{23.25}}
 \\ \midrule

\multirow{1}{*}{BadTeacher} 
& \begin{tabular}[c]{@{}c@{}}77.56\textsubscript{\textpm0.13}\\ (\textcolor{blue}{2.28})\end{tabular}
& \begin{tabular}[c]{@{}c@{}}99.96\textsubscript{\textpm0.00}\\ (\textcolor{blue}{0.03})\end{tabular}
& \begin{tabular}[c]{@{}c@{}}61.18\textsubscript{\textpm0.53}\\ (\textcolor{blue}{54.78})\end{tabular}
& \begin{tabular}[c]{@{}c@{}}0.00\textsubscript{\textpm0.00}\\ (\textcolor{blue}{14.74})\end{tabular}
& \multirow{1}{*}{\textcolor{blue}{17.96}}
 \\ \midrule

\multirow{1}{*}{SalUn} 
& \begin{tabular}[c]{@{}c@{}}73.79\textsubscript{\textpm0.15}\\ (\textcolor{blue}{1.50})\end{tabular}
& \begin{tabular}[c]{@{}c@{}}99.96\textsubscript{\textpm0.01}\\ (\textcolor{blue}{0.03})\end{tabular}
& \begin{tabular}[c]{@{}c@{}}84.07\textsubscript{\textpm0.47}\\ (\textcolor{blue}{77.67})\end{tabular}
& \begin{tabular}[c]{@{}c@{}}1.45\textsubscript{\textpm0.05}\\ (\textcolor{blue}{13.30})\end{tabular}
& \multirow{1}{*}{\textcolor{blue}{23.12}}
 \\ \midrule

\multirow{1}{*}{NatMU} 
& \begin{tabular}[c]{@{}c@{}}74.99\textsubscript{\textpm0.18}\\ (\textcolor{blue}{0.30})\end{tabular}
& \begin{tabular}[c]{@{}c@{}}99.06\textsubscript{\textpm0.05}\\ (\textcolor{blue}{0.93})\end{tabular}
& \begin{tabular}[c]{@{}c@{}}14.98\textsubscript{\textpm0.16}\\ (\textcolor{blue}{8.58})\end{tabular}
& \begin{tabular}[c]{@{}c@{}}13.56\textsubscript{\textpm0.45}\\ (\textcolor{blue}{1.19})\end{tabular}
& \multirow{1}{*}{\textcolor{blue}{\textbf{2.75}}}
 \\ 
\bottomrule[1.5pt]

\end{tabular}
}
\end{table}

To evaluate the performance of various methods under this practical setting, we select the most difficult-to-learn samples as the forgetting samples. The difficulty of a sample is determined by its learning speed during pretraining—the slower the sample is learned, the higher the difficulty of it~\cite{DBLP:conf/cvpr/YuGZFWC22}. Since these samples exhibit low forgetting accuracy on the retrained model, they can, to some extent, represent the worst-case unlearning performance across different methods. Table~\ref{table:hardsample-cifar100-10} shows the performance of different methods on CIFAR-100 with a forgetting ratio of 10\%, and the results with different datasets and forgetting ratios are shown in the Appendix. 
%Table~\ref{table:hard-sample}.

\textbf{Results.} As shown in Table~\ref{table:hardsample-cifar100-10}, when the distribution of forgetting samples shifts, the overall performance of other methods deteriorates significantly. In contrast, NatMU still maintains strong performance, where various metrics  are close to those of the retrained model. The robust performance of NatMU in practical settings demonstrates its great potential for real-world unlearning applications, making it a promising candidate for practical machine unlearning.

% ---------------------------------------------------------------
% ---------------------------------------------------------------

% Ablation

% ---------------------------------------------------------------
% ---------------------------------------------------------------

\subsection{Unlearning efficiency}
\begin{figure}[ht]
\centering
\includegraphics[width=0.45\textwidth]{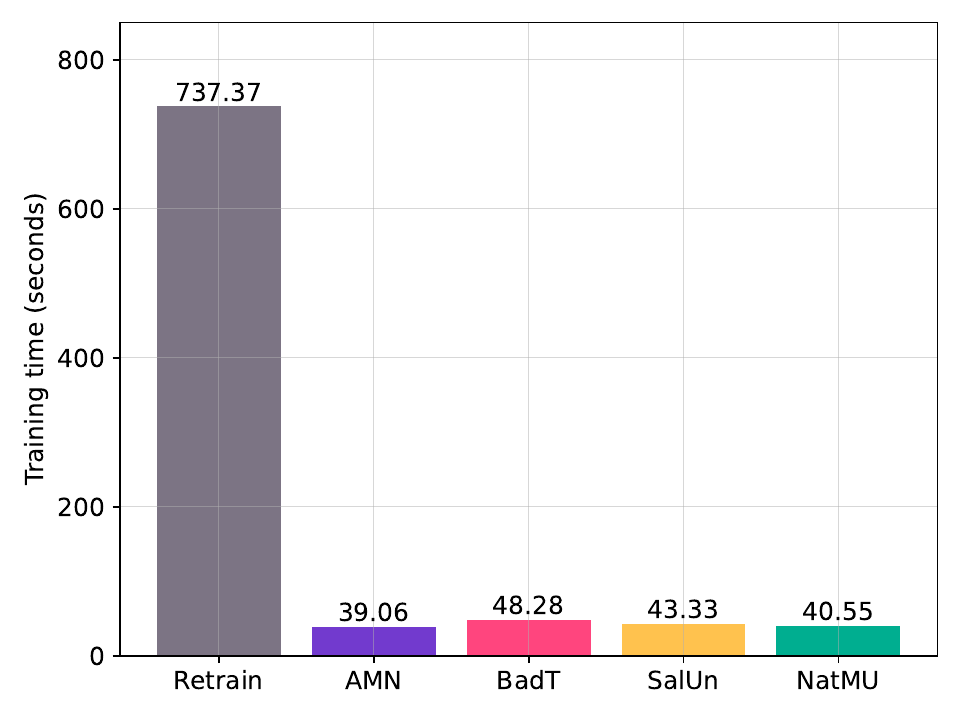}
\caption{Training time of different unlearning methods. NatMU achieves comparable unlearning efficiency to other baselines, and all unlearning methods require significantly less training time than retraining from scratch.}
\label{fig:training-efficiency}
\end{figure}

To compare the unlearning efficiency of different methods, we report the running time under random-sample-wise unlearning with 1\% forgetting ratio on CIFAR-100 using ResNet-18 in Fig.~\ref{fig:training-efficiency}. Our method, NatMU, introduces four hybrid unlearning samples for each forgetting data point, which increases the size of the fine-tuning dataset. However, this increase is marginal compared to the remaining dataset, and as a result, NatMU maintains an unlearning efficiency comparable to baselines. Notably, NatMU achieves a better unlearning performance, as shown in our main results.
Among the other methods, Amnesiac shows the highest efficiency, as it does not expand the dataset. BadTeacher incurs longer training time due to the need to compute soft labels with a teacher model at each epoch. SalUn also adds computational overhead, as it must compute a  weight saliency mask before unlearning. Despite these additional costs, all unlearning methods—including NatMU—require significantly less training time than full retraining, which must start from scratch.

\subsection{Ablation Study}

\subsubsection{The role of correct information}

\begin{table}[h]
\centering
\caption{Ablation study on the role of correct information.}
\label{table:ablation1}
\resizebox{0.48\textwidth}{!}{
\begin{tabular}{ccccccc}
\toprule[1.5pt]
Method & TA& RA& FA& MIA & Avg.Gap$\downarrow$   \\
\midrule
Retrain
& 76.68
& 99.94
& 77.60
& 55.40
& - \\  \midrule

% \multirow{1}{*}{$\{(\boldsymbol{x}^f,  y_j^r) \}_{j=1}^{n=4}$} 
\begin{tabular}[c]{@{}c@{}}$\{(\boldsymbol{x}^f,  y_j^r) \}_{j=1}^{n=4}$\\ 
(Multi-labeling)
\end{tabular}
& 
\begin{tabular}[c]{@{}c@{}} 75.42\\ 
(\textcolor{blue}{1.26})
\end{tabular}
& 
\begin{tabular}[c]{@{}c@{}} 97.12\\ 
(\textcolor{blue}{2.82})
\end{tabular}
& 
\begin{tabular}[c]{@{}c@{}} 6.04\\ 
(\textcolor{blue}{71.56})
\end{tabular}
& 
\begin{tabular}[c]{@{}c@{}} 0.00\\ 
(\textcolor{blue}{55.40})
\end{tabular}
& \multirow{1}{*}{\textcolor{blue}{32.76}}
\\ \midrule

% \multirow{1}{*}{} 
\begin{tabular}[c]{@{}c@{}} $\{ (\mathcal{T}_j(\boldsymbol{x}^f,\boldsymbol{0}_d),  y_j^r) \}_{j=1}^{n=4}$\\ 
(Segmentation-only)
\end{tabular}
& 
\begin{tabular}[c]{@{}c@{}} 74.62\\ 
(\textcolor{blue}{2.06})
\end{tabular}
& 
\begin{tabular}[c]{@{}c@{}} 97.29\\ 
(\textcolor{blue}{2.65})
\end{tabular}
& 
\begin{tabular}[c]{@{}c@{}} 64.72\\ 
(\textcolor{blue}{12.88})
\end{tabular}
& 
\begin{tabular}[c]{@{}c@{}} 39.76\\ 
(\textcolor{blue}{15.64})
\end{tabular}
& \multirow{1}{*}{\textcolor{blue}{8.31}}
\\ \midrule
% \multirow{1}{*}{$\{ (\mathcal{T}_j(\boldsymbol{x}^f,\boldsymbol{x}_j^r),  y_j^r) \}_{j=1}^{n=4}$} 
\begin{tabular}[c]{@{}c@{}} $\{ (\mathcal{T}_j(\boldsymbol{x}^f,\boldsymbol{x}_j^r),  y_j^r) \}_{j=1}^{n=4}$\\ 
(NatMU)
\end{tabular}
& 
\begin{tabular}[c]{@{}c@{}} 75.53\\ 
(\textcolor{blue}{1.15})
\end{tabular}
& 
\begin{tabular}[c]{@{}c@{}} 97.47\\ 
(\textcolor{blue}{2.47})
\end{tabular}
& 
\begin{tabular}[c]{@{}c@{}} 78.88\\ 
(\textcolor{blue}{1.28})
\end{tabular}
& 
\begin{tabular}[c]{@{}c@{}} 53.52\\ 
(\textcolor{blue}{1.88})
\end{tabular}
& \multirow{1}{*}{\textcolor{blue}{1.70}}
\\ 
\bottomrule[1.5pt]

\end{tabular}
}
\end{table}

NatMU improves previous relabeling-based unlearning methods by injecting correct information into unlearning instances. 
For a forgetting instance $(\boldsymbol{x}^f, y^f) \in \mathcal{D}^f$, we generate four unlearning instances 
$\{ (\mathcal{T}_j(\boldsymbol{x}^f,\boldsymbol{x}_j^r),  y_j^r) \}_{j=1}^{n=4}$ (we use $\mathcal{T}_j$ instead of $\mathcal{T}_{\boldsymbol{m}_j^\mathrm{scaled}}$ for short) .
To investigate the effectiveness of correct information, we replace the generation operation with two variants and conduct experiments on CIFAR-100 with a forgetting ratio of 1\%:

1) \textbf{Multi-labeling}: $\{(\boldsymbol{x}^f,  y_j^r) \}_{j=1}^{n=4}$. This relabels a forgetting instance with four most-related different labels.

2) \textbf{Segmentation-only}: $\{ (\mathcal{T}_j(\boldsymbol{x}^f,\boldsymbol{0}_d),  y_j^r) \}_{j=1}^{n=4}$. This replaces the remaining samples with a zero vector, thus removing the injected correct information but preserving the segmentation operation, which is introduced by the weighting mask.

As shown in Table~\ref{table:ablation1}, both multi-labeling NatMU and segmentation-only NatMU face a severe over-forgetting problem, just as other relabeling-based methods do. 
Moreover, the TA of segmentation-only NatMU decrease significantly, since the segmentation operation changes the input distribution. By injecting correct information, NatMU can not only eliminates the over-forgetting, but also maintain good model utility.

\begin{figure}[ht]
\centering
\includegraphics[width=0.45\textwidth]{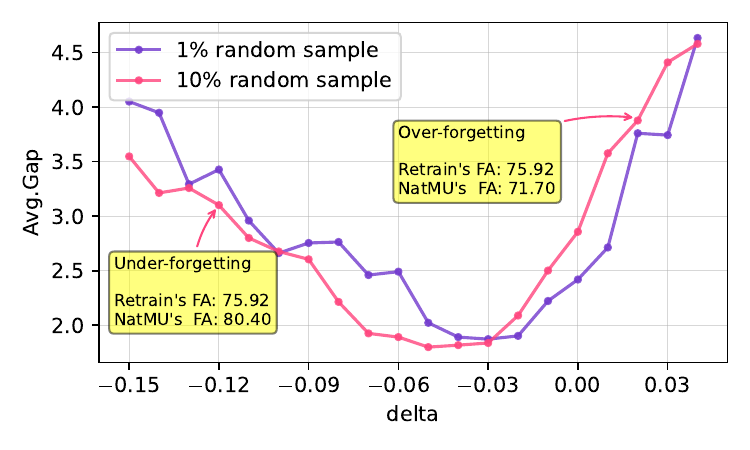}
\caption{Ablation study on $\delta$. The optimal values of $\delta$ are similar across different unlearning settings, suggesting that$\delta$ is transferable across unlearning settings.}
\label{fig:ablate_delta}
\end{figure}

\subsubsection{Results of different \texorpdfstring{$\delta$}{δ}}
The hyperparameter $\delta$ controls the proportion of forgetting samples in each hybrid unlearning instance. To investigate the impact of $\delta$ on unlearning performance, we vary its values and report the results under random-sample-wise unlearning with different forgetting ratios on CIFAR-100 using ResNet-18, as shown in Fig.~\ref{fig:ablate_delta}. 

When $\delta$ is small, the forgetting information in each hybrid sample is insufficient, making it difficult to effectively erase target information within a limited number of training epochs. Conversely, when $\delta$ is large, NatMU behaves similarly to Amnesiac, which adopts the random-relabeling technique and faces the over-forgetting problem. Notably, the optimal value of $\delta$ remains similar across different unlearning settings (different forgetting ratio),  indicating its transferability across settings and enabling practical unlearning without extensive hyperparameter tuning.

\subsubsection{Results of different \texorpdfstring{$n$}{n} }
To investigate how unlearning performance varies with different values of $n$, we modify $n$ and report the results in Table~\ref{table:ablation2}. For $n = 1$ or $2$, we randomly select $n$ masks from the four predefined hybridization masks for each forgetting sample. For $n = 8$, we repeat the full set of four masks to reach the desired count. All experiments are conducted on CIFAR-100 using ResNet-18, with hyperparameters searched under the 10\% random-sample-wise unlearning and transferred to the 10\% difficult-sample-wise unlearning setting.

As shown in Table~\ref{table:ablation2}, smaller values of $n$ result in worse performance. While increasing $n$ from 4 to 8 doubles the additional training cost, it yields only marginal performance gains. Based on this trade-off between unlearning effectiveness and efficiency, we adopt $n = 4$ in NatMU as a balanced and practical default.

\begin{table}[h]
\centering
\caption{ Ablation study on the unlearning instances number $n$. $n=4$ achieves a balance between unlearning efficiency and effectiveness. }
\label{table:ablation2}
\resizebox{0.48\textwidth}{!}{
\begin{tabular}{ccccccc}
\toprule[1.5pt]
Method & TA& RA& FA& MIA & Avg.Gap$\downarrow$   \\
\midrule
Retrain
& 75.28
& 99.99
& 6.40
& 14.74
& - \\  \midrule

% \multirow{1}{*}{$\{(\boldsymbol{x}^f,  y_j^r) \}_{j=1}^{n=4}$} 
\begin{tabular}[c]{@{}c@{}}$n=1$\\ 

\end{tabular}
& 
\begin{tabular}[c]{@{}c@{}} 74.76\\ 
(\textcolor{blue}{0.52})
\end{tabular}
& 
\begin{tabular}[c]{@{}c@{}} 98.95\\ 
(\textcolor{blue}{1.04})
\end{tabular}
& 
\begin{tabular}[c]{@{}c@{}} 21.58\\ 
(\textcolor{blue}{15.18})
\end{tabular}
& 
\begin{tabular}[c]{@{}c@{}} 21.90\\ 
(\textcolor{blue}{7.16})
\end{tabular}
& \multirow{1}{*}{\textcolor{blue}{5.98}}
\\ \midrule

% \multirow{1}{*}{} 
\begin{tabular}[c]{@{}c@{}} $n=2$\\ 

\end{tabular}
& 
\begin{tabular}[c]{@{}c@{}} 74.35\\ 
(\textcolor{blue}{0.93})
\end{tabular}
& 
\begin{tabular}[c]{@{}c@{}} 98.69\\ 
(\textcolor{blue}{1.30})
\end{tabular}
& 
\begin{tabular}[c]{@{}c@{}} 17.14\\ 
(\textcolor{blue}{10.74})
\end{tabular}
& 
\begin{tabular}[c]{@{}c@{}} 16.28\\ 
(\textcolor{blue}{1.54})
\end{tabular}
& \multirow{1}{*}{\textcolor{blue}{3.63}}
\\ \midrule
% \multirow{1}{*}{$\{ (\mathcal{T}_j(\boldsymbol{x}^f,\boldsymbol{x}_j^r),  y_j^r) \}_{j=1}^{n=4}$} 
\begin{tabular}[c]{@{}c@{}} $n=4$\\ 
(NatMU)
\end{tabular}
& 
\begin{tabular}[c]{@{}c@{}} 74.99\\ 
(\textcolor{blue}{0.29})
\end{tabular}
& 
\begin{tabular}[c]{@{}c@{}} 99.06\\ 
(\textcolor{blue}{0.93})
\end{tabular}
& 
\begin{tabular}[c]{@{}c@{}} 14.98\\ 
(\textcolor{blue}{8.58})
\end{tabular}
& 
\begin{tabular}[c]{@{}c@{}} 13.56\\ 
(\textcolor{blue}{1.18})
\end{tabular}
& \multirow{1}{*}{\textcolor{blue}{2.75}}
\\ \midrule
\begin{tabular}[c]{@{}c@{}} $n=8$\\ 

\end{tabular}
& 
\begin{tabular}[c]{@{}c@{}} 74.58\\ 
(\textcolor{blue}{0.70})
\end{tabular}
& 
\begin{tabular}[c]{@{}c@{}} 99.27\\ 
(\textcolor{blue}{0.72})
\end{tabular}
& 
\begin{tabular}[c]{@{}c@{}} 14.26\\ 
(\textcolor{blue}{7.86})
\end{tabular}
& 
\begin{tabular}[c]{@{}c@{}} 13.20\\ 
(\textcolor{blue}{1.54})
\end{tabular}
& \multirow{1}{*}{\textcolor{blue}{2.71}}
\\ 
\bottomrule[1.5pt]

\end{tabular}
}
\end{table}

\section{Conclusion}
Revisiting current popular machine unlearning methods, we identify their unnatural properties and resulting issues, such as unnatural generalization and impracticality. To address these issues, we propose a straightforward but effective machine unlearning method, NatMU, which injects correct information from remaining data into forgetting samples to achieve unlearning. NatMU demonstrates superior performance and robustness across different unlearning settings. Our initial step towards natural machine unlearning opens up new perspectives for achieving more efficient and effective machine unlearning.

\newpage

% \ifCLASSOPTIONcompsoc
%   % The Computer Society usually uses the plural form
%   \section*{Acknowledgments}
% \else
%   % regular IEEE prefers the singular form
% \fi

% This work was supported by the National Natural Science
% Foundation of China (No. 62376155, 61977046),
% Shanghai Municipal Science and Technology Major Project
% (No. 2021SHZDZX0102) and the Shanghai Science and Technology
% Program under Grant (No. 21JC1400600).

\ifCLASSOPTIONcaptionsoff
  \newpage
\fi



\bibliographystyle{IEEEtran}
\bibliography{reference}

\begin{thebibliography}{10}
\providecommand{\url}[1]{#1}
\csname url@samestyle\endcsname
\providecommand{\newblock}{\relax}
\providecommand{\bibinfo}[2]{#2}
\providecommand{\BIBentrySTDinterwordspacing}{\spaceskip=0pt\relax}
\providecommand{\BIBentryALTinterwordstretchfactor}{4}
\providecommand{\BIBentryALTinterwordspacing}{\spaceskip=\fontdimen2\font plus
\BIBentryALTinterwordstretchfactor\fontdimen3\font minus \fontdimen4\font\relax}
\providecommand{\BIBforeignlanguage}[2]{{%
\expandafter\ifx\csname l@#1\endcsname\relax
\typeout{** WARNING: IEEEtran.bst: No hyphenation pattern has been}%
\typeout{** loaded for the language `#1'. Using the pattern for}%
\typeout{** the default language instead.}%
\else
\language=\csname l@#1\endcsname
\fi
#2}}
\providecommand{\BIBdecl}{\relax}
\BIBdecl

\bibitem{DBLP:journals/pami/BadrinarayananK17}
V.~Badrinarayanan, A.~Kendall, and R.~Cipolla, ``Segnet: {A} deep convolutional encoder-decoder architecture for image segmentation,'' \emph{{IEEE} Trans. Pattern Anal. Mach. Intell.}, vol.~39, no.~12, pp. 2481--2495, 2017.

\bibitem{DBLP:journals/pami/CroitoruHIS23}
F.~Croitoru, V.~Hondru, R.~T. Ionescu, and M.~Shah, ``Diffusion models in vision: {A} survey,'' \emph{{IEEE} Trans. Pattern Anal. Mach. Intell.}, vol.~45, no.~9, pp. 10\,850--10\,869, 2023.

\bibitem{DBLP:journals/pami/HongZLLLYYLGJPGBC24}
D.~Hong, B.~Zhang, X.~Li, Y.~Li, C.~Li, J.~Yao, N.~Yokoya, H.~Li, P.~Ghamisi, X.~Jia, A.~Plaza, P.~Gamba, J.~A. Benediktsson, and J.~Chanussot, ``Spectralgpt: Spectral remote sensing foundation model,'' \emph{{IEEE} Trans. Pattern Anal. Mach. Intell.}, vol.~46, no.~8, pp. 5227--5244, 2024.

\bibitem{DBLP:journals/pami/BaltrusaitisAM19}
T.~Baltrusaitis, C.~Ahuja, and L.~Morency, ``Multimodal machine learning: {A} survey and taxonomy,'' \emph{{IEEE} Trans. Pattern Anal. Mach. Intell.}, vol.~41, no.~2, pp. 423--443, 2019.

\bibitem{hoofnagle2019european}
C.~J. Hoofnagle, B.~Van Der~Sloot, and F.~Z. Borgesius, ``The european union general data protection regulation: what it is and what it means,'' \emph{Information \& Communications Technology Law}, 2019.

\bibitem{GinartGVZ19}
A.~Ginart, M.~Y. Guan, G.~Valiant, and J.~Zou, ``Making {AI} forget you: Data deletion in machine learning,'' in \emph{NeurIPS}, 2019, pp. 3513--3526.

\bibitem{GuoGHM20}
C.~Guo, T.~Goldstein, A.~Y. Hannun, and L.~van~der Maaten, ``Certified data removal from machine learning models,'' in \emph{{ICML}}, ser. Proceedings of Machine Learning Research, vol. 119.\hskip 1em plus 0.5em minus 0.4em\relax {PMLR}, 2020, pp. 3832--3842.

\bibitem{UllahM0RA21}
E.~Ullah, T.~Mai, A.~Rao, R.~A. Rossi, and R.~Arora, ``Machine unlearning via algorithmic stability,'' in \emph{{COLT}}, ser. Proceedings of Machine Learning Research, vol. 134.\hskip 1em plus 0.5em minus 0.4em\relax {PMLR}, 2021, pp. 4126--4142.

\bibitem{goel2022towards}
G.~Shashwat, P.~Ameya, S.~Amartya, L.~Ser-Nam, T.~Philip, and K.~Ponnurangam, ``Towards adversarial evaluations for inexact machine unlearning,'' \emph{CoRR}, vol. abs/2201.06640, 2022.

\bibitem{ChundawatTMK23}
V.~S. Chundawat, A.~K. Tarun, M.~Mandal, and M.~S. Kankanhalli, ``Zero-shot machine unlearning,'' \emph{{IEEE} Trans. Inf. Forensics Secur.}, vol.~18, pp. 2345--2354, 2023.

\bibitem{assd}
S.~Schoepf, J.~Foster, and A.~Brintrup, ``Parameter-tuning-free data entry error unlearning with adaptive selective synaptic dampening,'' \emph{CoRR}, vol. abs/2402.10098, 2024.

\bibitem{puma}
G.~Wu, M.~Hashemi, and C.~Srinivasa, ``{PUMA:} performance unchanged model augmentation for training data removal,'' in \emph{{AAAI}}.\hskip 1em plus 0.5em minus 0.4em\relax {AAAI} Press, 2022, pp. 8675--8682.

\bibitem{amnesiac}
L.~Graves, V.~Nagisetty, and V.~Ganesh, ``Amnesiac machine learning,'' in \emph{{AAAI}}.\hskip 1em plus 0.5em minus 0.4em\relax {AAAI} Press, 2021, pp. 11\,516--11\,524.

\bibitem{badteacher}
V.~S. Chundawat, A.~K. Tarun, M.~Mandal, and M.~S. Kankanhalli, ``Can bad teaching induce forgetting? unlearning in deep networks using an incompetent teacher,'' in \emph{{AAAI}}.\hskip 1em plus 0.5em minus 0.4em\relax {AAAI} Press, 2023, pp. 7210--7217.

\bibitem{memorylocated}
P.~Maini, M.~C. Mozer, H.~Sedghi, Z.~C. Lipton, J.~Z. Kolter, and C.~Zhang, ``Can neural network memorization be localized?'' in \emph{{ICML}}, ser. Proceedings of Machine Learning Research, vol. 202.\hskip 1em plus 0.5em minus 0.4em\relax {PMLR}, 2023, pp. 23\,536--23\,557.

\bibitem{salun}
C.~Fan, J.~Liu, Y.~Zhang, D.~Wei, E.~Wong, and S.~Liu, ``Salun: Empowering machine unlearning via gradient-based weight saliency in both image classification and generation,'' \emph{CoRR}, vol. abs/2310.12508, 2023.

\bibitem{DBLP:conf/uss/ThudiJSP22}
A.~Thudi, H.~Jia, I.~Shumailov, and N.~Papernot, ``On the necessity of auditable algorithmic definitions for machine unlearning,'' in \emph{{USENIX} Security Symposium}.\hskip 1em plus 0.5em minus 0.4em\relax {USENIX} Association, 2022, pp. 4007--4022.

\bibitem{unrolling-sgd}
A.~Thudi, G.~Deza, V.~Chandrasekaran, and N.~Papernot, ``Unrolling {SGD:} understanding factors influencing machine unlearning,'' in \emph{EuroS{\&}P}.\hskip 1em plus 0.5em minus 0.4em\relax {IEEE}, 2022, pp. 303--319.

\bibitem{DBLP:journals/corr/abs-2308-07061}
J.~Xu, Z.~Wu, C.~Wang, and X.~Jia, ``Machine unlearning: Solutions and challenges,'' \emph{CoRR}, vol. abs/2308.07061, 2023.

\bibitem{CLIP}
A.~Radford, J.~W. Kim, C.~Hallacy, A.~Ramesh, G.~Goh, S.~Agarwal, G.~Sastry, A.~Askell, P.~Mishkin, J.~Clark, G.~Krueger, and I.~Sutskever, ``Learning transferable visual models from natural language supervision,'' in \emph{{ICML}}, ser. Proceedings of Machine Learning Research, vol. 139.\hskip 1em plus 0.5em minus 0.4em\relax {PMLR}, 2021, pp. 8748--8763.

\bibitem{gpt}
T.~B. Brown, B.~Mann, N.~Ryder, M.~Subbiah, J.~Kaplan, P.~Dhariwal, A.~Neelakantan, P.~Shyam, G.~Sastry, A.~Askell, S.~Agarwal, A.~Herbert{-}Voss, G.~Krueger, T.~Henighan, R.~Child, A.~Ramesh, D.~M. Ziegler, J.~Wu, C.~Winter, C.~Hesse, M.~Chen, E.~Sigler, M.~Litwin, S.~Gray, B.~Chess, J.~Clark, C.~Berner, S.~McCandlish, A.~Radford, I.~Sutskever, and D.~Amodei, ``Language models are few-shot learners,'' in \emph{NeurIPS}, 2020.

\bibitem{mixup}
H.~Zhang, M.~Ciss{\'{e}}, Y.~N. Dauphin, and D.~Lopez{-}Paz, ``mixup: Beyond empirical risk minimization,'' in \emph{{ICLR} (Poster)}.\hskip 1em plus 0.5em minus 0.4em\relax OpenReview.net, 2018.

\bibitem{first}
Y.~Cao and J.~Yang, ``Towards making systems forget with machine unlearning,'' in \emph{{IEEE} Symposium on Security and Privacy}.\hskip 1em plus 0.5em minus 0.4em\relax {IEEE} Computer Society, 2015, pp. 463--480.

\bibitem{KohL17}
P.~W. Koh and P.~Liang, ``Understanding black-box predictions via influence functions,'' in \emph{{ICML}}, ser. Proceedings of Machine Learning Research, vol.~70.\hskip 1em plus 0.5em minus 0.4em\relax {PMLR}, 2017, pp. 1885--1894.

\bibitem{DBLP:conf/nips/TannoPNL22}
R.~Tanno, M.~F. Pradier, A.~V. Nori, and Y.~Li, ``Repairing neural networks by leaving the right past behind,'' in \emph{NeurIPS}, 2022.

\bibitem{IzzoSCZ21}
Z.~Izzo, M.~A. Smart, K.~Chaudhuri, and J.~Zou, ``Approximate data deletion from machine learning models,'' in \emph{{AISTATS}}, ser. Proceedings of Machine Learning Research, vol. 130.\hskip 1em plus 0.5em minus 0.4em\relax {PMLR}, 2021, pp. 2008--2016.

\bibitem{MehtaPSR22}
R.~Mehta, S.~Pal, V.~Singh, and S.~N. Ravi, ``Deep unlearning via randomized conditionally independent hessians,'' in \emph{{CVPR}}.\hskip 1em plus 0.5em minus 0.4em\relax {IEEE}, 2022, pp. 10\,412--10\,421.

\bibitem{DBLP:conf/ndss/WarneckePWR23}
A.~Warnecke, L.~Pirch, C.~Wressnegger, and K.~Rieck, ``Machine unlearning of features and labels,'' in \emph{{NDSS}}.\hskip 1em plus 0.5em minus 0.4em\relax The Internet Society, 2023.

\bibitem{fisherforgetting}
A.~Golatkar, A.~Achille, and S.~Soatto, ``Eternal sunshine of the spotless net: Selective forgetting in deep networks,'' in \emph{{CVPR}}.\hskip 1em plus 0.5em minus 0.4em\relax Computer Vision Foundation / {IEEE}, 2020, pp. 9301--9309.

\bibitem{fim}
J.~Martens, ``New insights and perspectives on the natural gradient method,'' \emph{J. Mach. Learn. Res.}, vol.~21, pp. 146:1--146:76, 2020.

\bibitem{ssd}
J.~Foster, S.~Schoepf, and A.~Brintrup, ``Fast machine unlearning without retraining through selective synaptic dampening,'' in \emph{{AAAI}}.\hskip 1em plus 0.5em minus 0.4em\relax {AAAI} Press, 2024, pp. 12\,043--12\,051.

\bibitem{targeted}
S.~Ye, J.~Lu, and G.~Zhang, ``Towards safe machine unlearning: {A} paradigm that mitigates performance degradation,'' in \emph{{WWW}}.\hskip 1em plus 0.5em minus 0.4em\relax {ACM}, 2025, pp. 4635--4652.

\bibitem{ltu}
M.~H. Huang, L.~G. Foo, and J.~Liu, ``Learning to unlearn for robust machine unlearning,'' in \emph{{ECCV} {(52)}}, ser. Lecture Notes in Computer Science, vol. 15110.\hskip 1em plus 0.5em minus 0.4em\relax Springer, 2024, pp. 202--219.

\bibitem{scrub}
M.~Kurmanji, P.~Triantafillou, J.~Hayes, and E.~Triantafillou, ``Towards unbounded machine unlearning,'' in \emph{NeurIPS}, 2023.

\bibitem{cf-k}
S.~Goel, A.~Prabhu, A.~Sanyal, S.-N. Lim, P.~Torr, and P.~Kumaraguru, ``Towards adversarial evaluations for inexact machine unlearning,'' \emph{arXiv preprint arXiv:2201.06640}, 2022.

\bibitem{sparse}
J.~Jia, J.~Liu, P.~Ram, Y.~Yao, G.~Liu, Y.~Liu, P.~Sharma, and S.~Liu, ``Model sparsity can simplify machine unlearning,'' in \emph{NeurIPS}, 2023.

\bibitem{mu-mis}
X.~Cheng, Z.~Huang, and X.~Huang, ``Machine unlearning by suppressing sample contribution,'' \emph{CoRR}, vol. abs/2402.15109, 2024.

\bibitem{zero-shot-lips}
J.~Foster, K.~Fogarty, S.~Schoepf, C.~{\"{O}}ztireli, and A.~Brintrup, ``Zero-shot machine unlearning at scale via lipschitz regularization,'' \emph{CoRR}, vol. abs/2402.01401, 2024.

\bibitem{learningtounlearn}
S.~Cha, S.~Cho, D.~Hwang, H.~Lee, T.~Moon, and M.~Lee, ``Learning to unlearn: Instance-wise unlearning for pre-trained classifiers,'' in \emph{{AAAI}}.\hskip 1em plus 0.5em minus 0.4em\relax {AAAI} Press, 2024, pp. 11\,186--11\,194.

\bibitem{cifar}
A.~Krizhevsky, G.~Hinton \emph{et~al.}, ``Learning multiple layers of features from tiny images,'' 2009.

\bibitem{rethinking}
C.~Zhang, S.~Bengio, M.~Hardt, B.~Recht, and O.~Vinyals, ``Understanding deep learning requires rethinking generalization,'' in \emph{{ICLR}}.\hskip 1em plus 0.5em minus 0.4em\relax OpenReview.net, 2017.

\bibitem{lrp}
A.~Binder, G.~Montavon, S.~Lapuschkin, K.~M{\"{u}}ller, and W.~Samek, ``Layer-wise relevance propagation for neural networks with local renormalization layers,'' in \emph{{ICANN} {(2)}}, ser. Lecture Notes in Computer Science, vol. 9887.\hskip 1em plus 0.5em minus 0.4em\relax Springer, 2016, pp. 63--71.

\bibitem{cutmix}
S.~Yun, D.~Han, S.~Chun, S.~J. Oh, Y.~Yoo, and J.~Choe, ``Cutmix: Regularization strategy to train strong classifiers with localizable features,'' in \emph{{ICCV}}.\hskip 1em plus 0.5em minus 0.4em\relax {IEEE}, 2019, pp. 6022--6031.

\bibitem{tinyvit}
K.~Wu, J.~Zhang, H.~Peng, M.~Liu, B.~Xiao, J.~Fu, and L.~Yuan, ``Tinyvit: Fast pretraining distillation for small vision transformers,'' in \emph{{ECCV} {(21)}}, ser. Lecture Notes in Computer Science, vol. 13681.\hskip 1em plus 0.5em minus 0.4em\relax Springer, 2022, pp. 68--85.

\bibitem{DBLP:conf/ijcnn/MuhammadY20}
M.~B. Muhammad and M.~Yeasin, ``Eigen-cam: Class activation map using principal components,'' in \emph{{IJCNN}}.\hskip 1em plus 0.5em minus 0.4em\relax {IEEE}, 2020, pp. 1--7.

\bibitem{DBLP:conf/iccv/SelvarajuCDVPB17}
R.~R. Selvaraju, M.~Cogswell, A.~Das, R.~Vedantam, D.~Parikh, and D.~Batra, ``Grad-cam: Visual explanations from deep networks via gradient-based localization,'' in \emph{{ICCV}}.\hskip 1em plus 0.5em minus 0.4em\relax {IEEE} Computer Society, 2017, pp. 618--626.

\bibitem{jacobgilpytorchcam}
J.~Gildenblat and contributors, ``Pytorch library for cam methods,'' \url{https://github.com/jacobgil/pytorch-grad-cam}, 2021.

\bibitem{kullback1951information}
S.~Kullback and R.~A. Leibler, ``On information and sufficiency,'' \emph{The annals of mathematical statistics}, vol.~22, no.~1, pp. 79--86, 1951.

\bibitem{tiny-imagenet}
\BIBentryALTinterwordspacing
M.~A. mnmoustafa, ``Tiny imagenet,'' 2017. [Online]. Available: \url{https://kaggle.com/competitions/tiny-imagenet}
\BIBentrySTDinterwordspacing

\bibitem{vgg}
K.~Simonyan and A.~Zisserman, ``Very deep convolutional networks for large-scale image recognition,'' in \emph{{ICLR}}, 2015.

\bibitem{resnet}
K.~He, X.~Zhang, S.~Ren, and J.~Sun, ``Deep residual learning for image recognition,'' in \emph{{CVPR}}.\hskip 1em plus 0.5em minus 0.4em\relax {IEEE} Computer Society, 2016, pp. 770--778.

\bibitem{amun}
\BIBentryALTinterwordspacing
A.~Ebrahimpour-Boroojeny, H.~Sundaram, and V.~Chandrasekaran, ``Not all wrong is bad: Using adversarial examples for unlearning,'' in \emph{ICML}, 2025. [Online]. Available: \url{https://openreview.net/forum?id=BkrIQPREkn}
\BIBentrySTDinterwordspacing

\bibitem{sparsity}
J.~Jia, J.~Liu, P.~Ram, Y.~Yao, G.~Liu, Y.~Liu, P.~Sharma, and S.~Liu, ``Model sparsity can simplify machine unlearning,'' in \emph{NeurIPS}, 2023.

\bibitem{mia}
L.~Song and P.~Mittal, ``Systematic evaluation of privacy risks of machine learning models,'' in \emph{{USENIX} Security Symposium}.\hskip 1em plus 0.5em minus 0.4em\relax {USENIX} Association, 2021, pp. 2615--2632.

\bibitem{DBLP:conf/nips/HuangCZWHLH24}
Z.~Huang, X.~Cheng, J.~Zheng, H.~Wang, Z.~He, T.~Li, and X.~Huang, ``Unified gradient-based machine unlearning with remain geometry enhancement,'' in \emph{NeurIPS}, 2024.

\bibitem{DBLP:conf/eccv/FanLHL24}
C.~Fan, J.~Liu, A.~O. Hero, and S.~Liu, ``Challenging forgets: Unveiling the worst-case forget sets in machine unlearning,'' in \emph{{ECCV} {(21)}}, ser. Lecture Notes in Computer Science, vol. 15079.\hskip 1em plus 0.5em minus 0.4em\relax Springer, 2024, pp. 278--297.

\bibitem{DBLP:conf/mm/0006ZS0024}
S.~Lin, X.~Zhang, W.~Susilo, X.~Chen, and J.~Liu, ``{GDR-GMA:} machine unlearning via direction-rectified and magnitude-adjusted gradients,'' in \emph{{ACM} Multimedia}.\hskip 1em plus 0.5em minus 0.4em\relax {ACM}, 2024, pp. 9087--9095.

\bibitem{DBLP:conf/nips/LiuSBSLN21}
Y.~Liu, E.~Sangineto, W.~Bi, N.~Sebe, B.~Lepri, and M.~D. Nadai, ``Efficient training of visual transformers with small datasets,'' in \emph{NeurIPS}, 2021, pp. 23\,818--23\,830.

\bibitem{app13095521}
J.~Maurício, I.~Domingues, and J.~Bernardino, ``Comparing vision transformers and convolutional neural networks for image classification: A literature review,'' \emph{Applied Sciences}, vol.~13, no.~9, 2023.

\bibitem{DBLP:journals/corr/abs-2206-00389}
L.~Deininger, B.~Stimpel, A.~Y{\"{u}}ce, S.~Abbasi{-}Sureshjani, S.~Sch{\"{o}}nenberger, P.~Ocampo, K.~Korski, and F.~Gaire, ``A comparative study between vision transformers and cnns in digital pathology,'' \emph{CoRR}, vol. abs/2206.00389, 2022.

\bibitem{DBLP:conf/cvpr/YuGZFWC22}
S.~Yu, J.~Guo, R.~Zhang, Y.~Fan, Z.~Wang, and X.~Cheng, ``A re-balancing strategy for class-imbalanced classification based on instance difficulty,'' in \emph{{CVPR}}.\hskip 1em plus 0.5em minus 0.4em\relax {IEEE}, 2022, pp. 70--79.

\bibitem{lira}
N.~Carlini, S.~Chien, M.~Nasr, S.~Song, A.~Terzis, and F.~Tram{\`{e}}r, ``Membership inference attacks from first principles,'' in \emph{{SP}}.\hskip 1em plus 0.5em minus 0.4em\relax {IEEE}, 2022, pp. 1897--1914.

\bibitem{timm}
R.~Wightman, ``Pytorch image models,'' \url{https://github.com/rwightman/pytorch-image-models}, 2019.

\bibitem{DBLP:conf/kdd/AkibaSYOK19}
T.~Akiba, S.~Sano, T.~Yanase, T.~Ohta, and M.~Koyama, ``Optuna: {A} next-generation hyperparameter optimization framework,'' in \emph{{KDD}}.\hskip 1em plus 0.5em minus 0.4em\relax {ACM}, 2019, pp. 2623--2631.

\bibitem{cosine-lr}
I.~Loshchilov and F.~Hutter, ``{SGDR:} stochastic gradient descent with warm restarts,'' in \emph{{ICLR} (Poster)}.\hskip 1em plus 0.5em minus 0.4em\relax OpenReview.net, 2017.

\bibitem{impair}
A.~K. Tarun, V.~S. Chundawat, M.~Mandal, and M.~Kankanhalli, ``Fast yet effective machine unlearning,'' \emph{IEEE Transactions on Neural Networks and Learning Systems}, 2023.

\end{thebibliography}

\newpage
\appendices

\setcounter{section}{0}      % 重置section编号
\setcounter{subsection}{0}   % 重置subsection编号
\setcounter{figure}{0}       % 重置图片编号
\setcounter{table}{0}        % 重置表格编号
\setcounter{equation}{0}     % 重置公式编号
\setcounter{algorithm}{0}     % 重置公式编号
% 修改section的编号格式为A,B,C...
\renewcommand{\thesection}{\Alph{section}}
% 修改subsection的编号格式为A1,A2,B1,B2...
\renewcommand{\thesubsection}{\Alph{section}\arabic{subsection}}
% 修改图片编号格式为A.1,A.2,B.1...
\renewcommand{\thefigure}{\Alph{section}.\arabic{figure}}
% 修改表格编号格式为A.1,A.2,B.1...
\renewcommand{\thetable}{\Alph{section}.\arabic{table}}
% 修改公式编号格式为(A.1),(A.2),(B.1)...
\renewcommand{\theequation}{\Alph{section}.\arabic{equation}}
\renewcommand{\thealgorithm}{\Alph{section}.\arabic{algorithm}}

\section{Additional Experiment Results}

\subsection{Class-wise Results with More Classes}
In Table~\ref{table:class-wise-more-classes}, we report the results of more forgetting classes. The hyperparameters are set to practical type, transferred from the class Rocker and Cattle. 

\subsection{Difficult-Sample Unlearning Results}

In Table~\ref{table:hard-sample}, we report the results of difficult-sample-wise unlearning under various datasets and forgetting ratios. We can see that NatMU achieves the smallest average performance gap across all settings. We note that in TinyImageNet, relabeling-free methods could obtain a smaller forgetting accuracy gap compared to NatMU. However, NatMU could make a smaller MIA ratio gap, resulting a small Avg.Gap.

\begin{table*}[t!]
\centering
\caption{
Class-wise unlearning results on CIFAR-100/20 using ResNet-18 with more forgetting classes.
The results are given by $\bm{a_{\pm b}(\textcolor{blue}{c})}$, sharing the same format with Table \ref{table:class-wise}.}
\label{table:class-wise-more-classes}
\resizebox{\textwidth}{!}{
\begin{tabular}{cccccccccccccc}
\toprule[1.5pt]
\multirow{2}{*}{\begin{tabular}[c]{@{}c@{}}Unlearning\\ task\end{tabular}} 
&
\multirow{2}{*}{\begin{tabular}[c]{@{}c@{}}Hyperparameter\\ Type\end{tabular}} 
& \multirow{2}{*}{Metric} 
& \multicolumn{10}{c}{Methods} \\ 
\cmidrule[0.75pt]{4-13} 
 
 & &  & Retrain & SSD & NegGrad+ &SCRUB & AMUN &$\delta\text{-targeted}$ & Amnesiac & BadTeacher & SalUn  &NatMU \\ \midrule[1pt]
%----------------------------------------------------------------------------------
{\multirow{6}{*}{\begin{tabular}[c]{@{}c@{}}Full-class\\(Mushroom) \end{tabular}}} 
& {\multirow{6}{*}{\begin{tabular}[c]{@{}c@{}} Practical \end{tabular}}} 
&
TA
& 76.56\textsubscript{\textpm0.18}
& 76.78\textsubscript{\textpm0.00}(\textcolor{blue}{0.22})
& 76.10\textsubscript{\textpm0.09}(\textcolor{blue}{0.46})
& 75.56\textsubscript{\textpm0.16}(\textcolor{blue}{1.00})
& 77.04\textsubscript{\textpm0.11}(\textcolor{blue}{0.48})
& 77.02\textsubscript{\textpm0.13}(\textcolor{blue}{0.46})
& 76.81\textsubscript{\textpm0.12}(\textcolor{blue}{0.25})
& 76.36\textsubscript{\textpm0.11}(\textcolor{blue}{0.20})
& 76.58\textsubscript{\textpm0.18}(\textcolor{blue}{0.02})
& 76.43\textsubscript{\textpm0.19}(\textcolor{blue}{0.13})

 \\ 
&
&
RA
& 99.94\textsubscript{\textpm0.01}
& 99.95\textsubscript{\textpm0.00}(\textcolor{blue}{0.02})
& 99.98\textsubscript{\textpm0.00}(\textcolor{blue}{0.04})
& 99.96\textsubscript{\textpm0.00}(\textcolor{blue}{0.02})
& 99.98\textsubscript{\textpm0.00}(\textcolor{blue}{0.04})
& 99.98\textsubscript{\textpm0.00}(\textcolor{blue}{0.04})
& 99.98\textsubscript{\textpm0.00}(\textcolor{blue}{0.04})
& 99.84\textsubscript{\textpm0.02}(\textcolor{blue}{0.09})
& 99.98\textsubscript{\textpm0.00}(\textcolor{blue}{0.04})
& 99.54\textsubscript{\textpm0.01}(\textcolor{blue}{0.40})

 \\ 
&
&
FATrain
& 0.00\textsubscript{\textpm0.00}
& 0.00\textsubscript{\textpm0.00}(\textcolor{blue}{0.00})
& 0.00\textsubscript{\textpm0.00}(\textcolor{blue}{0.00})
& 0.04\textsubscript{\textpm0.08}(\textcolor{blue}{0.04})
& 0.00\textsubscript{\textpm0.00}(\textcolor{blue}{0.00})
& 0.72\textsubscript{\textpm0.10}(\textcolor{blue}{0.72})
& 0.00\textsubscript{\textpm0.00}(\textcolor{blue}{0.00})
& 0.00\textsubscript{\textpm0.00}(\textcolor{blue}{0.00})
& 0.00\textsubscript{\textpm0.00}(\textcolor{blue}{0.00})
& 0.00\textsubscript{\textpm0.00}(\textcolor{blue}{0.00})

 \\ 
&
&
FATest
& 0.00\textsubscript{\textpm0.00}
& 0.00\textsubscript{\textpm0.00}(\textcolor{blue}{0.00})
& 0.00\textsubscript{\textpm0.00}(\textcolor{blue}{0.00})
& 0.00\textsubscript{\textpm0.00}(\textcolor{blue}{0.00})
& 0.00\textsubscript{\textpm0.00}(\textcolor{blue}{0.00})
& 7.80\textsubscript{\textpm1.17}(\textcolor{blue}{7.80})
& 0.00\textsubscript{\textpm0.00}(\textcolor{blue}{0.00})
& 0.00\textsubscript{\textpm0.00}(\textcolor{blue}{0.00})
& 0.00\textsubscript{\textpm0.00}(\textcolor{blue}{0.00})
& 0.00\textsubscript{\textpm0.00}(\textcolor{blue}{0.00})

 \\ 
&
&
MIA
& 7.28\textsubscript{\textpm1.23}
& 0.40\textsubscript{\textpm0.00}(\textcolor{blue}{6.88})
& 4.28\textsubscript{\textpm0.47}(\textcolor{blue}{3.00})
& 16.08\textsubscript{\textpm1.33}(\textcolor{blue}{8.80})
& 8.68\textsubscript{\textpm0.90}(\textcolor{blue}{1.40})
& 5.08\textsubscript{\textpm0.41}(\textcolor{blue}{2.20})
& 12.84\textsubscript{\textpm0.79}(\textcolor{blue}{5.56})
& 0.00\textsubscript{\textpm0.00}(\textcolor{blue}{7.28})
& 9.76\textsubscript{\textpm0.71}(\textcolor{blue}{2.48})
& 6.12\textsubscript{\textpm0.41}(\textcolor{blue}{1.16})

 \\ 
\arrayrulecolor{lightgray} 
\cmidrule[0.75pt]{3-13} \arrayrulecolor{black}

&
&  Avg.Gap$\downarrow$  
 & - 
& \textcolor{blue}{1.42}
& \textcolor{blue}{0.70}
& \textcolor{blue}{1.97}
& \textcolor{blue}{0.39}
& \textcolor{blue}{2.24}
& \textcolor{blue}{1.17}
& \textcolor{blue}{1.51}
& \textcolor{blue}{0.51}
& \textbf{\textcolor{blue}{0.34}}

\\ \midrule
%----------------------------------------------------------------------------------
{\multirow{6}{*}{\begin{tabular}[c]{@{}c@{}}Full-class\\(Lamp) \end{tabular}}} 
& {\multirow{6}{*}{\begin{tabular}[c]{@{}c@{}} Practical \end{tabular}}} 
& 
TA
& 76.61\textsubscript{\textpm0.24}
& 76.89\textsubscript{\textpm0.02}(\textcolor{blue}{0.28})
& 76.16\textsubscript{\textpm0.30}(\textcolor{blue}{0.45})
& 75.64\textsubscript{\textpm0.34}(\textcolor{blue}{0.96})
& 77.15\textsubscript{\textpm0.20}(\textcolor{blue}{0.54})
& 77.24\textsubscript{\textpm0.08}(\textcolor{blue}{0.63})
& 76.66\textsubscript{\textpm0.11}(\textcolor{blue}{0.05})
& 76.46\textsubscript{\textpm0.17}(\textcolor{blue}{0.15})
& 76.76\textsubscript{\textpm0.15}(\textcolor{blue}{0.16})
& 76.60\textsubscript{\textpm0.21}(\textcolor{blue}{0.01})

 \\ 
&
&
RA
& 99.94\textsubscript{\textpm0.00}
& 99.96\textsubscript{\textpm0.00}(\textcolor{blue}{0.01})
& 99.97\textsubscript{\textpm0.00}(\textcolor{blue}{0.03})
& 99.96\textsubscript{\textpm0.01}(\textcolor{blue}{0.02})
& 99.98\textsubscript{\textpm0.00}(\textcolor{blue}{0.04})
& 99.98\textsubscript{\textpm0.00}(\textcolor{blue}{0.04})
& 99.98\textsubscript{\textpm0.00}(\textcolor{blue}{0.03})
& 99.85\textsubscript{\textpm0.01}(\textcolor{blue}{0.09})
& 99.98\textsubscript{\textpm0.00}(\textcolor{blue}{0.04})
& 99.54\textsubscript{\textpm0.03}(\textcolor{blue}{0.40})

 \\ 
&
&
FATrain
& 0.00\textsubscript{\textpm0.00}
& 0.00\textsubscript{\textpm0.00}(\textcolor{blue}{0.00})
& 0.00\textsubscript{\textpm0.00}(\textcolor{blue}{0.00})
& 0.40\textsubscript{\textpm0.70}(\textcolor{blue}{0.40})
& 0.00\textsubscript{\textpm0.00}(\textcolor{blue}{0.00})
& 0.72\textsubscript{\textpm0.16}(\textcolor{blue}{0.72})
& 0.00\textsubscript{\textpm0.00}(\textcolor{blue}{0.00})
& 0.00\textsubscript{\textpm0.00}(\textcolor{blue}{0.00})
& 0.00\textsubscript{\textpm0.00}(\textcolor{blue}{0.00})
& 0.00\textsubscript{\textpm0.00}(\textcolor{blue}{0.00})

 \\ 
&
&
FATest
& 0.00\textsubscript{\textpm0.00}
& 0.00\textsubscript{\textpm0.00}(\textcolor{blue}{0.00})
& 0.00\textsubscript{\textpm0.00}(\textcolor{blue}{0.00})
& 0.20\textsubscript{\textpm0.40}(\textcolor{blue}{0.20})
& 0.00\textsubscript{\textpm0.00}(\textcolor{blue}{0.00})
& 3.80\textsubscript{\textpm0.40}(\textcolor{blue}{3.80})
& 0.00\textsubscript{\textpm0.00}(\textcolor{blue}{0.00})
& 0.00\textsubscript{\textpm0.00}(\textcolor{blue}{0.00})
& 0.00\textsubscript{\textpm0.00}(\textcolor{blue}{0.00})
& 0.00\textsubscript{\textpm0.00}(\textcolor{blue}{0.00})

 \\ 
&
&
MIA
& 5.00\textsubscript{\textpm0.49}
& 0.00\textsubscript{\textpm0.00}(\textcolor{blue}{5.00})
& 2.56\textsubscript{\textpm0.62}(\textcolor{blue}{2.44})
& 6.68\textsubscript{\textpm0.88}(\textcolor{blue}{1.68})
& 5.48\textsubscript{\textpm1.67}(\textcolor{blue}{0.48})
& 4.04\textsubscript{\textpm0.43}(\textcolor{blue}{0.96})
& 10.72\textsubscript{\textpm1.82}(\textcolor{blue}{5.72})
& 0.00\textsubscript{\textpm0.00}(\textcolor{blue}{5.00})
& 8.80\textsubscript{\textpm1.21}(\textcolor{blue}{3.80})
& 5.36\textsubscript{\textpm0.45}(\textcolor{blue}{0.36})

 \\ 
\arrayrulecolor{lightgray} 
\cmidrule[0.75pt]{3-13} \arrayrulecolor{black}

&
&  Avg.Gap$\downarrow$  
 & - 
& \textcolor{blue}{1.06}
& \textcolor{blue}{0.58}
& \textcolor{blue}{0.65}
& \textcolor{blue}{0.21}
& \textcolor{blue}{1.23}
& \textcolor{blue}{1.16}
& \textcolor{blue}{1.05}
& \textcolor{blue}{0.80}
& \textbf{\textcolor{blue}{0.15}}

\\ \midrule[1.5pt]

%----------------------------------------------------------------------------------
{\multirow{6}{*}{\begin{tabular}[c]{@{}c@{}}Sub-class\\(Mushroom) \end{tabular}}} 
& {\multirow{6}{*}{\begin{tabular}[c]{@{}c@{}} Practical \end{tabular}}} 
&
TA
& 85.15\textsubscript{\textpm0.16}
& 84.22\textsubscript{\textpm0.58}(\textcolor{blue}{0.92})
& 84.79\textsubscript{\textpm0.21}(\textcolor{blue}{0.36})
& 83.79\textsubscript{\textpm0.19}(\textcolor{blue}{1.36})
& 85.01\textsubscript{\textpm0.09}(\textcolor{blue}{0.14})
& 85.08\textsubscript{\textpm0.06}(\textcolor{blue}{0.07})
& 85.17\textsubscript{\textpm0.05}(\textcolor{blue}{0.02})
& 85.08\textsubscript{\textpm0.05}(\textcolor{blue}{0.07})
& 84.80\textsubscript{\textpm0.09}(\textcolor{blue}{0.35})
& 84.57\textsubscript{\textpm0.08}(\textcolor{blue}{0.58})

 \\ 
&
&
RA
& 99.96\textsubscript{\textpm0.01}
& 98.76\textsubscript{\textpm0.82}(\textcolor{blue}{1.19})
& 99.99\textsubscript{\textpm0.00}(\textcolor{blue}{0.03})
& 99.97\textsubscript{\textpm0.00}(\textcolor{blue}{0.01})
& 99.99\textsubscript{\textpm0.00}(\textcolor{blue}{0.03})
& 99.97\textsubscript{\textpm0.00}(\textcolor{blue}{0.01})
& 99.98\textsubscript{\textpm0.00}(\textcolor{blue}{0.02})
& 99.96\textsubscript{\textpm0.00}(\textcolor{blue}{0.00})
& 99.97\textsubscript{\textpm0.00}(\textcolor{blue}{0.01})
& 98.16\textsubscript{\textpm0.06}(\textcolor{blue}{1.80})

 \\ 
&
&
FATrain
& 7.32\textsubscript{\textpm1.07}
& 11.88\textsubscript{\textpm10.98}(\textcolor{blue}{4.56})
& 6.56\textsubscript{\textpm1.29}(\textcolor{blue}{0.76})
& 3.64\textsubscript{\textpm0.29}(\textcolor{blue}{3.68})
& 7.08\textsubscript{\textpm0.70}(\textcolor{blue}{0.24})
& 16.88\textsubscript{\textpm0.86}(\textcolor{blue}{9.56})
& 9.24\textsubscript{\textpm1.50}(\textcolor{blue}{1.92})
& 20.28\textsubscript{\textpm2.88}(\textcolor{blue}{12.96})
& 14.56\textsubscript{\textpm1.55}(\textcolor{blue}{7.24})
& 12.08\textsubscript{\textpm1.18}(\textcolor{blue}{4.76})

 \\ 
&
&
FATest
& 4.40\textsubscript{\textpm1.74}
& 7.40\textsubscript{\textpm7.03}(\textcolor{blue}{3.00})
& 2.00\textsubscript{\textpm1.10}(\textcolor{blue}{2.40})
& 2.00\textsubscript{\textpm1.41}(\textcolor{blue}{2.40})
& 9.20\textsubscript{\textpm0.98}(\textcolor{blue}{4.80})
& 10.00\textsubscript{\textpm0.63}(\textcolor{blue}{5.60})
& 2.60\textsubscript{\textpm1.36}(\textcolor{blue}{1.80})
& 5.80\textsubscript{\textpm1.47}(\textcolor{blue}{1.40})
& 2.00\textsubscript{\textpm0.00}(\textcolor{blue}{2.40})
& 6.20\textsubscript{\textpm2.04}(\textcolor{blue}{1.80})

 \\ 
&
&
MIA
& 14.72\textsubscript{\textpm0.84}
& 1.20\textsubscript{\textpm0.28}(\textcolor{blue}{13.52})
& 4.12\textsubscript{\textpm0.80}(\textcolor{blue}{10.60})
& 45.04\textsubscript{\textpm0.48}(\textcolor{blue}{30.32})
& 17.72\textsubscript{\textpm2.13}(\textcolor{blue}{3.00})
& 2.08\textsubscript{\textpm0.20}(\textcolor{blue}{12.64})
& 0.00\textsubscript{\textpm0.00}(\textcolor{blue}{14.72})
& 0.00\textsubscript{\textpm0.00}(\textcolor{blue}{14.72})
& 0.24\textsubscript{\textpm0.08}(\textcolor{blue}{14.48})
& 11.16\textsubscript{\textpm1.13}(\textcolor{blue}{3.56})

 \\ 
\arrayrulecolor{lightgray} 
\cmidrule[0.75pt]{3-13} \arrayrulecolor{black}

&
&  Avg.Gap$\downarrow$  
 & - 
& \textcolor{blue}{4.64}
& \textcolor{blue}{2.83}
& \textcolor{blue}{7.55}
& \textbf{\textcolor{blue}{1.64}}
& \textcolor{blue}{5.57}
& \textcolor{blue}{3.70}
& \textcolor{blue}{5.83}
& \textcolor{blue}{4.90}
& \textcolor{blue}{2.50}

\\ \midrule
%----------------------------------------------------------------------------------
{\multirow{6}{*}{\begin{tabular}[c]{@{}c@{}}Sub-class\\(Lamp) \end{tabular}}} 
& {\multirow{6}{*}{\begin{tabular}[c]{@{}c@{}} Practical \end{tabular}}} 
& 
TA
& 85.03\textsubscript{\textpm0.18}
& 81.94\textsubscript{\textpm0.09}(\textcolor{blue}{3.09})
& 84.93\textsubscript{\textpm0.21}(\textcolor{blue}{0.10})
& 83.86\textsubscript{\textpm0.17}(\textcolor{blue}{1.17})
& 85.01\textsubscript{\textpm0.23}(\textcolor{blue}{0.02})
& 84.70\textsubscript{\textpm0.07}(\textcolor{blue}{0.33})
& 84.84\textsubscript{\textpm0.09}(\textcolor{blue}{0.19})
& 85.13\textsubscript{\textpm0.19}(\textcolor{blue}{0.10})
& 84.46\textsubscript{\textpm0.05}(\textcolor{blue}{0.57})
& 84.53\textsubscript{\textpm0.05}(\textcolor{blue}{0.50})

 \\ 
&
&
RA
& 99.95\textsubscript{\textpm0.01}
& 95.99\textsubscript{\textpm0.14}(\textcolor{blue}{3.96})
& 99.99\textsubscript{\textpm0.00}(\textcolor{blue}{0.04})
& 99.98\textsubscript{\textpm0.00}(\textcolor{blue}{0.02})
& 99.99\textsubscript{\textpm0.00}(\textcolor{blue}{0.04})
& 99.92\textsubscript{\textpm0.01}(\textcolor{blue}{0.03})
& 99.98\textsubscript{\textpm0.00}(\textcolor{blue}{0.03})
& 99.95\textsubscript{\textpm0.00}(\textcolor{blue}{0.00})
& 99.97\textsubscript{\textpm0.00}(\textcolor{blue}{0.02})
& 98.22\textsubscript{\textpm0.04}(\textcolor{blue}{1.73})

 \\ 
&
&
FATrain
& 16.08\textsubscript{\textpm2.26}
& 0.08\textsubscript{\textpm0.16}(\textcolor{blue}{16.00})
& 22.12\textsubscript{\textpm1.61}(\textcolor{blue}{6.04})
& 14.36\textsubscript{\textpm1.75}(\textcolor{blue}{1.72})
& 21.04\textsubscript{\textpm1.46}(\textcolor{blue}{4.96})
& 31.12\textsubscript{\textpm1.56}(\textcolor{blue}{15.04})
& 24.72\textsubscript{\textpm1.18}(\textcolor{blue}{8.64})
& 32.64\textsubscript{\textpm3.59}(\textcolor{blue}{16.56})
& 26.56\textsubscript{\textpm0.98}(\textcolor{blue}{10.48})
& 21.88\textsubscript{\textpm2.49}(\textcolor{blue}{5.80})

 \\ 
&
&
FATest
& 11.00\textsubscript{\textpm1.90}
& 0.20\textsubscript{\textpm0.40}(\textcolor{blue}{10.80})
& 15.80\textsubscript{\textpm1.94}(\textcolor{blue}{4.80})
& 15.40\textsubscript{\textpm3.67}(\textcolor{blue}{4.40})
& 30.20\textsubscript{\textpm2.14}(\textcolor{blue}{19.20})
& 12.40\textsubscript{\textpm0.80}(\textcolor{blue}{1.40})
& 8.40\textsubscript{\textpm1.02}(\textcolor{blue}{2.60})
& 12.00\textsubscript{\textpm1.41}(\textcolor{blue}{1.00})
& 9.80\textsubscript{\textpm0.75}(\textcolor{blue}{1.20})
& 16.60\textsubscript{\textpm1.36}(\textcolor{blue}{5.60})

 \\ 
&
&
MIA
& 14.24\textsubscript{\textpm1.01}
& 1.76\textsubscript{\textpm0.51}(\textcolor{blue}{12.48})
& 7.60\textsubscript{\textpm0.61}(\textcolor{blue}{6.64})
& 24.88\textsubscript{\textpm1.63}(\textcolor{blue}{10.64})
& 12.16\textsubscript{\textpm0.64}(\textcolor{blue}{2.08})
& 3.56\textsubscript{\textpm0.29}(\textcolor{blue}{10.68})
& 0.00\textsubscript{\textpm0.00}(\textcolor{blue}{14.24})
& 0.00\textsubscript{\textpm0.00}(\textcolor{blue}{14.24})
& 0.20\textsubscript{\textpm0.00}(\textcolor{blue}{14.04})
& 13.88\textsubscript{\textpm1.42}(\textcolor{blue}{0.36})

 \\ 
\arrayrulecolor{lightgray} 
\cmidrule[0.75pt]{3-13} \arrayrulecolor{black}

&
&  Avg.Gap$\downarrow$  
 & - 
& \textcolor{blue}{9.27}
& \textcolor{blue}{3.52}
& \textcolor{blue}{3.59}
& \textcolor{blue}{5.26}
& \textcolor{blue}{5.50}
& \textcolor{blue}{5.14}
& \textcolor{blue}{6.38}
& \textcolor{blue}{5.26}
& \textbf{\textcolor{blue}{2.80}}

\\ \bottomrule[1.5pt]
\end{tabular}
}
\end{table*}

\begin{table*}[t!]
\centering
\caption{Difficult-sample-wise unlearning results with different datasets, models, and forgetting ratios. The results are given by $\bm{a_{\pm b}(\textcolor{blue}{c})}$, sharing the same format with Table \ref{table:class-wise}.}
\label{table:hard-sample}
\resizebox{\textwidth}{!}{
\begin{tabular}{ccccccccccc}
\toprule[1.5pt]
\multirow{2}{*}{\begin{tabular}[c]{@{}c@{}}Unlearning\\ task\end{tabular}} 
& \multirow{2}{*}{Metric} 
& \multicolumn{9}{c}{Methods} \\ 
\cmidrule[0.75pt]{3-11} 
 
 &  & Retrain  & NegGrad+ &SCRUB & AMUN & $\delta\text{-targeted}^\lambda$ & Amnesiac & BadTeacher & SalUn  &NatMU \\ \midrule[1pt]
%----------------------------------------------------------------------------------
{\multirow{5}{*}{\begin{tabular}[c]{@{}c@{}}CIFAR-10\\ (VGG16-BN, 1\%)\end{tabular}}} & 
TA
& 93.24\textsubscript{\textpm0.06}
& 92.37\textsubscript{\textpm0.26}(\textcolor{blue}{0.87})
& 92.84\textsubscript{\textpm0.16}(\textcolor{blue}{0.40})
& 93.04\textsubscript{\textpm0.11}(\textcolor{blue}{0.20})
& 92.97\textsubscript{\textpm0.07}(\textcolor{blue}{0.27})
& 92.92\textsubscript{\textpm0.07}(\textcolor{blue}{0.32})
& 93.27\textsubscript{\textpm0.09}(\textcolor{blue}{0.03})
& 92.91\textsubscript{\textpm0.05}(\textcolor{blue}{0.32})
& 92.80\textsubscript{\textpm0.09}(\textcolor{blue}{0.44})

\\
&
RA
& 99.97\textsubscript{\textpm0.01}
& 99.99\textsubscript{\textpm0.01}(\textcolor{blue}{0.03})
& 99.99\textsubscript{\textpm0.00}(\textcolor{blue}{0.02})
& 100.00\textsubscript{\textpm0.00}(\textcolor{blue}{0.03})
& 99.99\textsubscript{\textpm0.00}(\textcolor{blue}{0.03})
& 100.00\textsubscript{\textpm0.00}(\textcolor{blue}{0.03})
& 99.96\textsubscript{\textpm0.01}(\textcolor{blue}{0.01})
& 100.00\textsubscript{\textpm0.00}(\textcolor{blue}{0.03})
& 99.25\textsubscript{\textpm0.02}(\textcolor{blue}{0.71})

\\
&
FA
& 5.96\textsubscript{\textpm0.64}
& 18.56\textsubscript{\textpm1.44}(\textcolor{blue}{12.60})
& 43.36\textsubscript{\textpm2.13}(\textcolor{blue}{37.40})
& 59.76\textsubscript{\textpm1.34}(\textcolor{blue}{53.80})
& 52.64\textsubscript{\textpm0.91}(\textcolor{blue}{46.68})
& 51.28\textsubscript{\textpm3.38}(\textcolor{blue}{45.32})
& 62.60\textsubscript{\textpm1.18}(\textcolor{blue}{56.64})
& 57.28\textsubscript{\textpm1.78}(\textcolor{blue}{51.32})
& 15.40\textsubscript{\textpm0.57}(\textcolor{blue}{9.44})

\\
&
MIA
& 30.68\textsubscript{\textpm1.14}
& 88.28\textsubscript{\textpm1.35}(\textcolor{blue}{57.60})
& 5.88\textsubscript{\textpm0.99}(\textcolor{blue}{24.80})
& 6.20\textsubscript{\textpm0.69}(\textcolor{blue}{24.48})
& 1.96\textsubscript{\textpm0.08}(\textcolor{blue}{28.72})
& 0.36\textsubscript{\textpm0.23}(\textcolor{blue}{30.32})
& 0.16\textsubscript{\textpm0.08}(\textcolor{blue}{30.52})
& 2.60\textsubscript{\textpm0.33}(\textcolor{blue}{28.08})
& 12.08\textsubscript{\textpm1.38}(\textcolor{blue}{18.60})

\\
\arrayrulecolor{lightgray} 
\cmidrule[0.75pt]{2-11} \arrayrulecolor{black}

& Avg.Gap$\downarrow$ 
 & - 
& \textcolor{blue}{17.77}
& \textcolor{blue}{15.66}
& \textcolor{blue}{19.63}
& \textcolor{blue}{18.92}
& \textcolor{blue}{19.00}
& \textcolor{blue}{21.80}
& \textcolor{blue}{19.94}
& \textbf{\textcolor{blue}{7.30}}

\\ \midrule
%----------------------------------------------------------------------------------
{\multirow{5}{*}{\begin{tabular}[c]{@{}c@{}}CIFAR-10\\ (VGG16-BN, 10\%)\end{tabular}}} & 
TA
& 91.62\textsubscript{\textpm0.15}
& 88.62\textsubscript{\textpm0.29}(\textcolor{blue}{3.00})
& 92.40\textsubscript{\textpm0.11}(\textcolor{blue}{0.78})
& 93.02\textsubscript{\textpm0.10}(\textcolor{blue}{1.40})
& 92.03\textsubscript{\textpm0.01}(\textcolor{blue}{0.41})
& 91.23\textsubscript{\textpm0.22}(\textcolor{blue}{0.39})
& 91.74\textsubscript{\textpm0.24}(\textcolor{blue}{0.12})
& 91.37\textsubscript{\textpm0.08}(\textcolor{blue}{0.25})
& 91.71\textsubscript{\textpm0.08}(\textcolor{blue}{0.09})

\\
&
RA
& 99.99\textsubscript{\textpm0.00}
& 99.64\textsubscript{\textpm0.02}(\textcolor{blue}{0.35})
& 100.00\textsubscript{\textpm0.00}(\textcolor{blue}{0.01})
& 100.00\textsubscript{\textpm0.00}(\textcolor{blue}{0.01})
& 99.85\textsubscript{\textpm0.01}(\textcolor{blue}{0.14})
& 99.97\textsubscript{\textpm0.01}(\textcolor{blue}{0.02})
& 99.99\textsubscript{\textpm0.00}(\textcolor{blue}{0.00})
& 99.91\textsubscript{\textpm0.01}(\textcolor{blue}{0.08})
& 99.84\textsubscript{\textpm0.00}(\textcolor{blue}{0.15})

\\
&
FA
& 35.73\textsubscript{\textpm0.27}
& 28.18\textsubscript{\textpm0.72}(\textcolor{blue}{7.55})
& 76.54\textsubscript{\textpm0.90}(\textcolor{blue}{40.81})
& 91.51\textsubscript{\textpm0.28}(\textcolor{blue}{55.78})
& 84.71\textsubscript{\textpm0.40}(\textcolor{blue}{48.98})
& 75.95\textsubscript{\textpm2.07}(\textcolor{blue}{40.22})
& 68.87\textsubscript{\textpm2.83}(\textcolor{blue}{33.14})
& 77.76\textsubscript{\textpm0.51}(\textcolor{blue}{42.03})
& 51.02\textsubscript{\textpm0.53}(\textcolor{blue}{15.29})

\\
&
MIA
& 16.18\textsubscript{\textpm0.53}
& 80.05\textsubscript{\textpm0.82}(\textcolor{blue}{63.87})
& 14.02\textsubscript{\textpm0.66}(\textcolor{blue}{2.16})
& 13.39\textsubscript{\textpm0.40}(\textcolor{blue}{2.80})
& 2.66\textsubscript{\textpm0.06}(\textcolor{blue}{13.52})
& 1.02\textsubscript{\textpm0.07}(\textcolor{blue}{15.16})
& 0.74\textsubscript{\textpm0.03}(\textcolor{blue}{15.44})
& 4.01\textsubscript{\textpm0.27}(\textcolor{blue}{12.18})
& 13.71\textsubscript{\textpm0.64}(\textcolor{blue}{2.48})

\\
\arrayrulecolor{lightgray} 
\cmidrule[0.75pt]{2-11} \arrayrulecolor{black}

& Avg.Gap$\downarrow$ 
 & - 
& \textcolor{blue}{18.69}
& \textcolor{blue}{10.94}
& \textcolor{blue}{15.00}
& \textcolor{blue}{15.76}
& \textcolor{blue}{13.95}
& \textcolor{blue}{12.17}
& \textcolor{blue}{13.63}
& \textbf{\textcolor{blue}{4.50}}
 
\\ \midrule[1.5pt]
%----------------------------------------------------------------------------------
{\multirow{5}{*}{\begin{tabular}[c]{@{}c@{}}CIFAR-100\\ (ResNet-18, 1\%)\end{tabular}}} & 
TA
& 76.54\textsubscript{\textpm0.17}
& 75.80\textsubscript{\textpm0.28}(\textcolor{blue}{0.74})
& 76.50\textsubscript{\textpm0.09}(\textcolor{blue}{0.04})
& 75.14\textsubscript{\textpm0.29}(\textcolor{blue}{1.40})
& 76.02\textsubscript{\textpm0.03}(\textcolor{blue}{0.52})
& 76.02\textsubscript{\textpm0.10}(\textcolor{blue}{0.52})
& 77.81\textsubscript{\textpm0.12}(\textcolor{blue}{1.27})
& 76.22\textsubscript{\textpm0.09}(\textcolor{blue}{0.32})
& 75.77\textsubscript{\textpm0.09}(\textcolor{blue}{0.77})

\\
&
RA
& 99.97\textsubscript{\textpm0.00}
& 99.99\textsubscript{\textpm0.00}(\textcolor{blue}{0.02})
& 99.99\textsubscript{\textpm0.00}(\textcolor{blue}{0.02})
& 99.99\textsubscript{\textpm0.00}(\textcolor{blue}{0.02})
& 99.98\textsubscript{\textpm0.00}(\textcolor{blue}{0.01})
& 99.99\textsubscript{\textpm0.00}(\textcolor{blue}{0.02})
& 99.94\textsubscript{\textpm0.01}(\textcolor{blue}{0.03})
& 99.99\textsubscript{\textpm0.00}(\textcolor{blue}{0.02})
& 97.94\textsubscript{\textpm0.06}(\textcolor{blue}{2.03})

\\
&
FA
& 1.20\textsubscript{\textpm0.42}
& 5.72\textsubscript{\textpm1.11}(\textcolor{blue}{4.52})
& 33.00\textsubscript{\textpm1.51}(\textcolor{blue}{31.80})
& 40.00\textsubscript{\textpm1.69}(\textcolor{blue}{38.80})
& 35.44\textsubscript{\textpm0.73}(\textcolor{blue}{34.24})
& 57.40\textsubscript{\textpm2.38}(\textcolor{blue}{56.20})
& 45.48\textsubscript{\textpm1.22}(\textcolor{blue}{44.28})
& 63.36\textsubscript{\textpm0.80}(\textcolor{blue}{62.16})
& 3.92\textsubscript{\textpm0.92}(\textcolor{blue}{2.72})

\\
&
MIA
& 26.76\textsubscript{\textpm0.76}
& 76.00\textsubscript{\textpm1.75}(\textcolor{blue}{49.24})
& 39.48\textsubscript{\textpm1.06}(\textcolor{blue}{12.72})
& 9.32\textsubscript{\textpm0.32}(\textcolor{blue}{17.44})
& 0.52\textsubscript{\textpm0.10}(\textcolor{blue}{26.24})
& 0.12\textsubscript{\textpm0.10}(\textcolor{blue}{26.64})
& 0.00\textsubscript{\textpm0.00}(\textcolor{blue}{26.76})
& 0.08\textsubscript{\textpm0.10}(\textcolor{blue}{26.68})
& 20.64\textsubscript{\textpm0.78}(\textcolor{blue}{6.12})

\\
\arrayrulecolor{lightgray} 
\cmidrule[0.75pt]{2-11} \arrayrulecolor{black}

& Avg.Gap$\downarrow$ 
 & - 
& \textcolor{blue}{13.63}
& \textcolor{blue}{11.14}
& \textcolor{blue}{14.41}
& \textcolor{blue}{15.25}
& \textcolor{blue}{20.85}
& \textcolor{blue}{18.08}
& \textcolor{blue}{22.30}
& \textbf{\textcolor{blue}{2.91}}

\\ \midrule

%----------------------------------------------------------------------------------
{\multirow{5}{*}{\begin{tabular}[c]{@{}c@{}}CIFAR-100\\ (ResNet-18, 10\%)\end{tabular}}} & 
TA
& 75.28\textsubscript{\textpm0.16}
& 70.97\textsubscript{\textpm0.24}(\textcolor{blue}{4.31})
& 73.90\textsubscript{\textpm0.10}(\textcolor{blue}{1.38})
& 75.28\textsubscript{\textpm0.11}(\textcolor{blue}{0.00})
& 75.00\textsubscript{\textpm0.05}(\textcolor{blue}{0.28})
& 73.56\textsubscript{\textpm0.10}(\textcolor{blue}{1.72})
& 77.56\textsubscript{\textpm0.13}(\textcolor{blue}{2.28})
& 73.79\textsubscript{\textpm0.15}(\textcolor{blue}{1.50})
& 74.99\textsubscript{\textpm0.18}(\textcolor{blue}{0.30})

\\
&
RA
& 99.99\textsubscript{\textpm0.00}
& 99.96\textsubscript{\textpm0.01}(\textcolor{blue}{0.03})
& 99.98\textsubscript{\textpm0.01}(\textcolor{blue}{0.01})
& 100.00\textsubscript{\textpm0.00}(\textcolor{blue}{0.01})
& 99.82\textsubscript{\textpm0.01}(\textcolor{blue}{0.17})
& 99.97\textsubscript{\textpm0.00}(\textcolor{blue}{0.02})
& 99.96\textsubscript{\textpm0.00}(\textcolor{blue}{0.03})
& 99.96\textsubscript{\textpm0.01}(\textcolor{blue}{0.03})
& 99.06\textsubscript{\textpm0.05}(\textcolor{blue}{0.93})

\\
&
FA
& 6.40\textsubscript{\textpm0.18}
& 7.75\textsubscript{\textpm0.31}(\textcolor{blue}{1.35})
& 30.80\textsubscript{\textpm0.22}(\textcolor{blue}{24.40})
& 75.43\textsubscript{\textpm0.54}(\textcolor{blue}{69.04})
& 78.33\textsubscript{\textpm0.20}(\textcolor{blue}{71.94})
& 83.85\textsubscript{\textpm0.16}(\textcolor{blue}{77.46})
& 61.18\textsubscript{\textpm0.53}(\textcolor{blue}{54.78})
& 84.07\textsubscript{\textpm0.47}(\textcolor{blue}{77.67})
& 14.98\textsubscript{\textpm0.16}(\textcolor{blue}{8.58})

\\
&
MIA
& 14.74\textsubscript{\textpm0.32}
& 70.74\textsubscript{\textpm0.73}(\textcolor{blue}{56.00})
& 31.54\textsubscript{\textpm0.35}(\textcolor{blue}{16.80})
& 5.92\textsubscript{\textpm0.17}(\textcolor{blue}{8.83})
& 7.04\textsubscript{\textpm0.16}(\textcolor{blue}{7.71})
& 0.94\textsubscript{\textpm0.09}(\textcolor{blue}{13.80})
& 0.00\textsubscript{\textpm0.00}(\textcolor{blue}{14.74})
& 1.45\textsubscript{\textpm0.05}(\textcolor{blue}{13.30})
& 13.56\textsubscript{\textpm0.45}(\textcolor{blue}{1.19})

\\
\arrayrulecolor{lightgray} 
\cmidrule[0.75pt]{2-11} \arrayrulecolor{black}

& Avg.Gap$\downarrow$ 
 & - 
& \textcolor{blue}{15.42}
& \textcolor{blue}{10.65}
& \textcolor{blue}{19.47}
& \textcolor{blue}{20.03}
& \textcolor{blue}{23.25}
& \textcolor{blue}{17.96}
& \textcolor{blue}{23.12}
& \textbf{\textcolor{blue}{2.75}}

\\ \midrule
%----------------------------------------------------------------------------------
{\multirow{5}{*}{\begin{tabular}[c]{@{}c@{}}TinyImageNet\\ (ResNet-34, 1\%)\end{tabular}}} & 
TA
& 66.67\textsubscript{\textpm0.32}
& 65.70\textsubscript{\textpm0.12}(\textcolor{blue}{0.97})
& 66.22\textsubscript{\textpm0.13}(\textcolor{blue}{0.45})
& 65.48\textsubscript{\textpm0.33}(\textcolor{blue}{1.19})
& 66.24\textsubscript{\textpm0.03}(\textcolor{blue}{0.43})
& 65.78\textsubscript{\textpm0.16}(\textcolor{blue}{0.89})
& 67.95\textsubscript{\textpm0.19}(\textcolor{blue}{1.28})
& 65.61\textsubscript{\textpm0.09}(\textcolor{blue}{1.06})
& 66.49\textsubscript{\textpm0.19}(\textcolor{blue}{0.18})

\\
&
RA
& 99.99\textsubscript{\textpm0.00}
& 100.00\textsubscript{\textpm0.00}(\textcolor{blue}{0.00})
& 99.99\textsubscript{\textpm0.00}(\textcolor{blue}{0.01})
& 99.98\textsubscript{\textpm0.00}(\textcolor{blue}{0.01})
& 100.00\textsubscript{\textpm0.00}(\textcolor{blue}{0.00})
& 100.00\textsubscript{\textpm0.00}(\textcolor{blue}{0.00})
& 99.97\textsubscript{\textpm0.00}(\textcolor{blue}{0.02})
& 100.00\textsubscript{\textpm0.00}(\textcolor{blue}{0.00})
& 98.80\textsubscript{\textpm0.02}(\textcolor{blue}{1.19})

\\
&
FA
& 0.28\textsubscript{\textpm0.17}
& 0.60\textsubscript{\textpm0.24}(\textcolor{blue}{0.32})
& 1.60\textsubscript{\textpm0.21}(\textcolor{blue}{1.32})
& 19.10\textsubscript{\textpm0.66}(\textcolor{blue}{18.82})
& 24.60\textsubscript{\textpm0.42}(\textcolor{blue}{24.32})
& 54.96\textsubscript{\textpm0.32}(\textcolor{blue}{54.68})
& 32.26\textsubscript{\textpm0.48}(\textcolor{blue}{31.98})
& 55.14\textsubscript{\textpm1.06}(\textcolor{blue}{54.86})
& 2.68\textsubscript{\textpm0.56}(\textcolor{blue}{2.40})

\\
&
MIA
& 12.58\textsubscript{\textpm0.35}
& 53.50\textsubscript{\textpm0.81}(\textcolor{blue}{40.92})
& 43.00\textsubscript{\textpm0.97}(\textcolor{blue}{30.42})
& 5.98\textsubscript{\textpm0.96}(\textcolor{blue}{6.60})
& 0.78\textsubscript{\textpm0.12}(\textcolor{blue}{11.80})
& 0.42\textsubscript{\textpm0.19}(\textcolor{blue}{12.16})
& 0.00\textsubscript{\textpm0.00}(\textcolor{blue}{12.58})
& 0.32\textsubscript{\textpm0.21}(\textcolor{blue}{12.26})
& 9.36\textsubscript{\textpm0.85}(\textcolor{blue}{3.22})

\\
\arrayrulecolor{lightgray} 
\cmidrule[0.75pt]{2-11} \arrayrulecolor{black}

& Avg.Gap$\downarrow$ 
 & - 
& \textcolor{blue}{10.55}
& \textcolor{blue}{8.05}
& \textcolor{blue}{6.66}
& \textcolor{blue}{9.14}
& \textcolor{blue}{16.93}
& \textcolor{blue}{11.47}
& \textcolor{blue}{17.05}
& \textbf{\textcolor{blue}{1.75}}

\\ \midrule
%----------------------------------------------------------------------------------
{\multirow{5}{*}{\begin{tabular}[c]{@{}c@{}}TinyImageNet\\ (ResNet-34, 10\%)\end{tabular}}} & 
TA
& 66.83\textsubscript{\textpm0.20}
& 62.10\textsubscript{\textpm0.20}(\textcolor{blue}{4.73})
& 66.49\textsubscript{\textpm0.25}(\textcolor{blue}{0.34})
& 65.94\textsubscript{\textpm0.39}(\textcolor{blue}{0.89})
& 66.62\textsubscript{\textpm0.08}(\textcolor{blue}{0.21})
& 65.83\textsubscript{\textpm0.08}(\textcolor{blue}{1.01})
& 68.09\textsubscript{\textpm0.12}(\textcolor{blue}{1.26})
& 65.26\textsubscript{\textpm0.15}(\textcolor{blue}{1.57})
& 66.33\textsubscript{\textpm0.09}(\textcolor{blue}{0.51})

\\
&
RA
& 100.00\textsubscript{\textpm0.00}
& 99.97\textsubscript{\textpm0.01}(\textcolor{blue}{0.03})
& 99.99\textsubscript{\textpm0.00}(\textcolor{blue}{0.01})
& 100.00\textsubscript{\textpm0.00}(\textcolor{blue}{0.00})
& 99.97\textsubscript{\textpm0.01}(\textcolor{blue}{0.03})
& 100.00\textsubscript{\textpm0.00}(\textcolor{blue}{0.00})
& 99.99\textsubscript{\textpm0.00}(\textcolor{blue}{0.01})
& 100.00\textsubscript{\textpm0.00}(\textcolor{blue}{0.00})
& 99.17\textsubscript{\textpm0.03}(\textcolor{blue}{0.83})

\\
&
FA
& 1.91\textsubscript{\textpm0.12}
& 1.75\textsubscript{\textpm0.11}(\textcolor{blue}{0.17})
& 8.48\textsubscript{\textpm0.39}(\textcolor{blue}{6.56})
& 44.44\textsubscript{\textpm0.66}(\textcolor{blue}{42.53})
& 78.28\textsubscript{\textpm0.20}(\textcolor{blue}{76.37})
& 90.98\textsubscript{\textpm0.21}(\textcolor{blue}{89.06})
& 46.69\textsubscript{\textpm0.56}(\textcolor{blue}{44.78})
& 88.31\textsubscript{\textpm0.23}(\textcolor{blue}{86.40})
& 12.18\textsubscript{\textpm0.25}(\textcolor{blue}{10.27})

\\
&
MIA
& 9.15\textsubscript{\textpm0.24}
& 64.61\textsubscript{\textpm0.41}(\textcolor{blue}{55.46})
& 16.45\textsubscript{\textpm0.55}(\textcolor{blue}{7.30})
& 3.48\textsubscript{\textpm0.19}(\textcolor{blue}{5.67})
& 3.74\textsubscript{\textpm0.08}(\textcolor{blue}{5.41})
& 0.18\textsubscript{\textpm0.02}(\textcolor{blue}{8.97})
& 0.00\textsubscript{\textpm0.00}(\textcolor{blue}{9.15})
& 0.20\textsubscript{\textpm0.03}(\textcolor{blue}{8.95})
& 7.81\textsubscript{\textpm0.41}(\textcolor{blue}{1.34})

\\
\arrayrulecolor{lightgray} 
\cmidrule[0.75pt]{2-11} \arrayrulecolor{black}

& Avg.Gap$\downarrow$ 
 & - 
& \textcolor{blue}{15.10}
& \textcolor{blue}{3.55}
& \textcolor{blue}{12.27}
& \textcolor{blue}{20.50}
& \textcolor{blue}{24.76}
& \textcolor{blue}{13.80}
& \textcolor{blue}{24.23}
& \textbf{\textcolor{blue}{3.23}}

\\ \midrule

%----------------------------------------------------------------------------------

{\multirow{5}{*}{\begin{tabular}[c]{@{}c@{}}TinyImageNet\\ (ViT, 1\%)\end{tabular} }} & 
TA
& 70.86\textsubscript{\textpm0.30}
& 68.52\textsubscript{\textpm0.21}(\textcolor{blue}{2.34})
& 68.45\textsubscript{\textpm0.24}(\textcolor{blue}{2.41})
& 69.09\textsubscript{\textpm0.60}(\textcolor{blue}{1.77})
& 69.30\textsubscript{\textpm0.09}(\textcolor{blue}{1.56})
& 68.62\textsubscript{\textpm0.26}(\textcolor{blue}{2.24})
& 71.31\textsubscript{\textpm0.08}(\textcolor{blue}{0.45})
& 68.43\textsubscript{\textpm0.31}(\textcolor{blue}{2.43})
& 68.75\textsubscript{\textpm0.10}(\textcolor{blue}{2.11})

\\
&
RA
& 100.00\textsubscript{\textpm0.00}
& 100.00\textsubscript{\textpm0.00}(\textcolor{blue}{0.00})
& 99.99\textsubscript{\textpm0.00}(\textcolor{blue}{0.00})
& 99.99\textsubscript{\textpm0.00}(\textcolor{blue}{0.01})
& 99.98\textsubscript{\textpm0.02}(\textcolor{blue}{0.01})
& 99.96\textsubscript{\textpm0.02}(\textcolor{blue}{0.04})
& 99.99\textsubscript{\textpm0.00}(\textcolor{blue}{0.01})
& 99.98\textsubscript{\textpm0.04}(\textcolor{blue}{0.02})
& 99.96\textsubscript{\textpm0.00}(\textcolor{blue}{0.03})

\\
&
FA
& 2.00\textsubscript{\textpm0.31}
& 4.74\textsubscript{\textpm0.50}(\textcolor{blue}{2.74})
& 46.18\textsubscript{\textpm1.65}(\textcolor{blue}{44.18})
& 55.78\textsubscript{\textpm1.11}(\textcolor{blue}{53.78})
& 50.82\textsubscript{\textpm0.20}(\textcolor{blue}{48.82})
& 4.04\textsubscript{\textpm0.52}(\textcolor{blue}{2.04})
& 79.38\textsubscript{\textpm0.33}(\textcolor{blue}{77.38})
& 3.90\textsubscript{\textpm0.32}(\textcolor{blue}{1.90})
& 11.08\textsubscript{\textpm0.44}(\textcolor{blue}{9.08})

\\
&
MIA
& 16.40\textsubscript{\textpm0.39}
& 66.28\textsubscript{\textpm1.79}(\textcolor{blue}{49.88})
& 72.32\textsubscript{\textpm0.81}(\textcolor{blue}{55.92})
& 9.94\textsubscript{\textpm0.49}(\textcolor{blue}{6.46})
& 8.52\textsubscript{\textpm0.25}(\textcolor{blue}{7.88})
& 1.42\textsubscript{\textpm0.37}(\textcolor{blue}{14.98})
& 0.00\textsubscript{\textpm0.00}(\textcolor{blue}{16.40})
& 1.46\textsubscript{\textpm0.42}(\textcolor{blue}{14.94})
& 10.42\textsubscript{\textpm0.90}(\textcolor{blue}{5.98})

\\
\arrayrulecolor{lightgray} 
\cmidrule[0.75pt]{2-11} \arrayrulecolor{black}

& Avg.Gap$\downarrow$ 
 & - 
&\textcolor{blue}{13.74}
&\textcolor{blue}{25.63}
&\textcolor{blue}{15.50}
&\textcolor{blue}{14.57}
&\textcolor{blue}{4.82}
&\textcolor{blue}{23.56}
&\textcolor{blue}{4.82}
&\textcolor{blue}{\textbf{4.30}}

\\ \midrule

%----------------------------------------------------------------------------------

{\multirow{5}{*}{\begin{tabular}[c]{@{}c@{}}TinyImageNet\\ (ViT, 10\%)\end{tabular}}} & 
TA
& 70.90\textsubscript{\textpm0.26}
& 65.07\textsubscript{\textpm0.12}(\textcolor{blue}{5.83})
& 61.71\textsubscript{\textpm4.47}(\textcolor{blue}{9.19})
& 70.33\textsubscript{\textpm0.16}(\textcolor{blue}{0.57})
& 68.49\textsubscript{\textpm0.07}(\textcolor{blue}{2.41})
& 70.74\textsubscript{\textpm0.13}(\textcolor{blue}{0.16})
& 71.12\textsubscript{\textpm0.31}(\textcolor{blue}{0.22})
& 70.53\textsubscript{\textpm0.16}(\textcolor{blue}{0.37})
& 66.95\textsubscript{\textpm0.23}(\textcolor{blue}{3.95})

\\
&
RA
& 100.00\textsubscript{\textpm0.00}
& 100.00\textsubscript{\textpm0.00}(\textcolor{blue}{0.00})
& 91.06\textsubscript{\textpm10.14}(\textcolor{blue}{8.94})
& 99.99\textsubscript{\textpm0.00}(\textcolor{blue}{0.01})
& 99.52\textsubscript{\textpm0.01}(\textcolor{blue}{0.48})
& 99.98\textsubscript{\textpm0.00}(\textcolor{blue}{0.02})
& 99.99\textsubscript{\textpm0.00}(\textcolor{blue}{0.01})
& 99.99\textsubscript{\textpm0.00}(\textcolor{blue}{0.01})
& 99.92\textsubscript{\textpm0.01}(\textcolor{blue}{0.08})

\\
&
FA
& 7.95\textsubscript{\textpm0.33}
& 3.85\textsubscript{\textpm0.13}(\textcolor{blue}{4.10})
& 29.64\textsubscript{\textpm12.58}(\textcolor{blue}{21.68})
& 80.17\textsubscript{\textpm0.22}(\textcolor{blue}{72.22})
& 83.91\textsubscript{\textpm0.16}(\textcolor{blue}{75.96})
& 89.46\textsubscript{\textpm0.36}(\textcolor{blue}{81.51})
& 72.14\textsubscript{\textpm1.02}(\textcolor{blue}{64.19})
& 90.56\textsubscript{\textpm0.31}(\textcolor{blue}{82.61})
& 13.28\textsubscript{\textpm0.34}(\textcolor{blue}{5.32})

\\
&
MIA
& 13.08\textsubscript{\textpm0.10}
& 65.46\textsubscript{\textpm0.76}(\textcolor{blue}{52.38})
& 22.26\textsubscript{\textpm1.68}(\textcolor{blue}{9.17})
& 23.76\textsubscript{\textpm0.61}(\textcolor{blue}{10.67})
& 19.42\textsubscript{\textpm0.09}(\textcolor{blue}{6.33})
& 1.63\textsubscript{\textpm0.08}(\textcolor{blue}{11.46})
& 0.00\textsubscript{\textpm0.00}(\textcolor{blue}{13.08})
& 1.76\textsubscript{\textpm0.11}(\textcolor{blue}{11.33})
& 11.68\textsubscript{\textpm0.32}(\textcolor{blue}{1.40})

\\
\arrayrulecolor{lightgray} 
\cmidrule[0.75pt]{2-11} \arrayrulecolor{black}

& Avg.Gap$\downarrow$ 
 & - 
&\textcolor{blue}{15.58}
&\textcolor{blue}{12.25}
&\textcolor{blue}{20.87}
&\textcolor{blue}{21.29}
&\textcolor{blue}{23.28}
&\textcolor{blue}{19.38}
&\textcolor{blue}{23.58}
&\textcolor{blue}{\textbf{2.69}}

\\ \bottomrule[1.5pt]
\end{tabular}
}
\end{table*}

\subsection{ROC Curves using Shadow-model-based MIA}
In our main experiments, we employ a prediction-based membership inference attack (MIA)~\cite{mia}, which is widely used in many unlearning works~\cite{badteacher,ssd,salun,DBLP:conf/nips/HuangCZWHLH24}. This attack trains a binary classifier by treating the test data as non-members and the remaining training data as members. The classifier takes  DNN's prediction entropy as input and output the classification result. Because it does not rely on multiple shadow models, it is both computationally efficient and readily adaptable to different forgetting sets.

To offer a complementary evaluation, we also plot the ROC curve for LiRA~\cite{lira}, a shadow-model-based MIA. Specifically, using CIFAR-100 with a forgetting ratio of 1\% as an example, we train 64 shadow models: each model’s training set includes every forgetting sample with probability 0.5, and we randomly sample additional examples from the remaining training and test sets to reach a total of 50000 samples per model. During the attack phase, we use a held-out subset of 500 test samples to determine classification thresholds for various false positive rates (FPR), and we compute the true positive rate (TPR), i.e., the fraction of forgetting samples classified as members.

\begin{figure}[ht]
\centering
\includegraphics[width=0.45\textwidth]{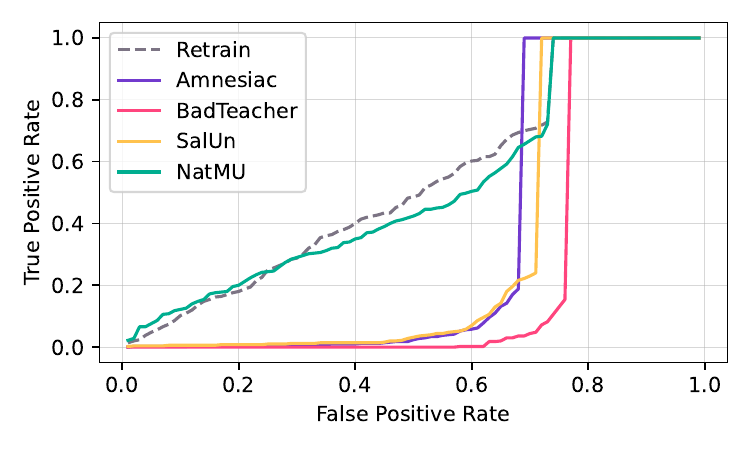}
\caption{ROC curves using the shadow-model-based MIA~\cite{lira}, evaluated on CIFAR-100 with ResNet-18 under a 1\% forgetting ratio. NatMU exhibits behavior similar to that of the retrained model, whereas the ROC curves of baseline methods significantly deviate from retraining.}
\label{fig:roc}
\end{figure}

The results in Fig~\ref{fig:roc} shows that, under the shadow-model-based MIA, the baseline methods exhibit substantially lower TPRs than Retrain, indicating greater deviation from the retrained model’s behavior. 
These findings also align with those from prediction-based MIA~\cite{mia}, indicating consistent conclusions across both attack types.
As discussed in \cite{fisherforgetting}, such low TPRs can cause a ``Streisand Effect'', where the forgetting samples are actually more noticeable. In contrast, NatMU achieves a TPR nearly identical to that of Retrain, demonstrating close alignment with the retrained model.

\subsection{Results with Different Masks}\label{appendix:other-masks}
In the design of NatMU, we employ four gradual  \mixup masks for each forgetting sample. To investigate the impact of using different masks, we compare the performance of gradual \mixup, non-gradual \mixup, and CutMix-based masks in Table~\ref{table:other-masks}. The non-gradual  \mixup masks are initialized with all values set to 0.5 and are scaled by $\delta$ to control the mixing strength. In contrast, the CutMix masks are initialized with patches in the four corners, where the edge length determines the mask strength.

The results show that, compared to non-gradual  \mixup, NatMU with gradual  \mixup achieves lower gaps in both TA and MIA, resulting in a reduced Avg.Gap. While the use of CutMix masks leads to better performance than non-gradual  \mixup, it still falls short of the performance achieved by gradual  \mixup. Notably, across all mask variants, NatMU consistently outperforms relabeling-based baselines, demonstrating the effectiveness of our approach based on injecting correct information at the input level.

\begin{table}[h]
\centering
\caption{Comparison of NatMU with different masking strategies and compared relabeling-based baselines, evaluated on CIFAR-100 with ResNet-18 under 10\% forgetting ratio.
NatMU with gradual  \mixup achieves the lowest Avg.Gap among all masking variants. Across all mask variants, NatMU consistently outperforms relabeling-based baselines, highlighting the effectiveness of ``injecting correct information''.}
\label{table:other-masks}
\resizebox{0.48\textwidth}{!}{
\begin{tabular}{ccccccc}
\toprule[1.5pt]
Method & TA& RA& FA& MIA & Avg.Gap$\downarrow$   \\
\midrule
Retrain
& 75.43
& 99.95
& 75.92
& 53.49
& - \\  \midrule

% \multirow{1}{*}{$\{(\boldsymbol{x}^f,  y_j^r) \}_{j=1}^{n=4}$} 

\begin{tabular}[c]{@{}c@{}} Amnesiac \\ 

\end{tabular}
& 
\begin{tabular}[c]{@{}c@{}} 71.23\\ 
(\textcolor{blue}{4.21})
\end{tabular}
& 
\begin{tabular}[c]{@{}c@{}} 99.92\\ 
(\textcolor{blue}{0.03})
\end{tabular}
& 
\begin{tabular}[c]{@{}c@{}} 94.90\\ 
(\textcolor{blue}{18.98})
\end{tabular}
& 
\begin{tabular}[c]{@{}c@{}} 20.66\\ 
(\textcolor{blue}{32.82})
\end{tabular}
& \multirow{1}{*}{\textcolor{blue}{14.01}}

\\ \midrule
\begin{tabular}[c]{@{}c@{}} BadTeacher \\ 

\end{tabular}
& 
\begin{tabular}[c]{@{}c@{}} 76.97\\ 
(\textcolor{blue}{1.54})
\end{tabular}
& 
\begin{tabular}[c]{@{}c@{}} 99.88\\ 
(\textcolor{blue}{0.06})
\end{tabular}
& 
\begin{tabular}[c]{@{}c@{}} 78.08\\ 
(\textcolor{blue}{2.16})
\end{tabular}
& 
\begin{tabular}[c]{@{}c@{}} 0.60\\ 
(\textcolor{blue}{52.89})
\end{tabular}
& \multirow{1}{*}{\textcolor{blue}{14.16}}
\\ \midrule
\begin{tabular}[c]{@{}c@{}} SalUn \\ 

\end{tabular}
& 
\begin{tabular}[c]{@{}c@{}} 71.57\\ 
(\textcolor{blue}{3.86})
\end{tabular}
& 
\begin{tabular}[c]{@{}c@{}} 99.88\\ 
(\textcolor{blue}{0.07})
\end{tabular}
& 
\begin{tabular}[c]{@{}c@{}} 94.30\\ 
(\textcolor{blue}{18.38})
\end{tabular}
& 
\begin{tabular}[c]{@{}c@{}}20.69\\ 
(\textcolor{blue}{32.80})
\end{tabular}
& \multirow{1}{*}{\textcolor{blue}{13.78}}
\\ \midrule

% \multirow{1}{*}{} 
\begin{tabular}[c]{@{}c@{}} NatMU \\ 
(Non-gradual \mixup)
\end{tabular}
& 
\begin{tabular}[c]{@{}c@{}} 72.54\\ 
(\textcolor{blue}{2.89})
\end{tabular}
& 
\begin{tabular}[c]{@{}c@{}} 97.66\\ 
(\textcolor{blue}{2.29})
\end{tabular}
& 
\begin{tabular}[c]{@{}c@{}} 75.62\\ 
(\textcolor{blue}{0.30})
\end{tabular}
& 
\begin{tabular}[c]{@{}c@{}} 40.68\\ 
(\textcolor{blue}{12.81})
\end{tabular}
& \multirow{1}{*}{\textcolor{blue}{4.57}}
\\ \midrule
% \multirow{1}{*}{$\{ (\mathcal{T}_j(\boldsymbol{x}^f,\boldsymbol{x}_j^r),  y_j^r) \}_{j=1}^{n=4}$} 
\begin{tabular}[c]{@{}c@{}} NatMU \\ 
(CutMix)
\end{tabular}
& 
\begin{tabular}[c]{@{}c@{}} 73.40\\ 
(\textcolor{blue}{2.03})
\end{tabular}
& 
\begin{tabular}[c]{@{}c@{}} 97.97\\ 
(\textcolor{blue}{1.98})
\end{tabular}
& 
\begin{tabular}[c]{@{}c@{}} 76.38\\ 
(\textcolor{blue}{0.46})
\end{tabular}
& 
\begin{tabular}[c]{@{}c@{}} 42.66\\ 
(\textcolor{blue}{10.83})
\end{tabular}
& \multirow{1}{*}{\textcolor{blue}{3.83}}
\\ \midrule

\begin{tabular}[c]{@{}c@{}} NatMU \\ 
(Gradual \mixup)
\end{tabular}
& 
\begin{tabular}[c]{@{}c@{}} 74.27\\ 
(\textcolor{blue}{1.16})
\end{tabular}
& 
\begin{tabular}[c]{@{}c@{}} 96.88\\ 
(\textcolor{blue}{3.07})
\end{tabular}
& 
\begin{tabular}[c]{@{}c@{}} 76.73\\ 
(\textcolor{blue}{0.82})
\end{tabular}
& 
\begin{tabular}[c]{@{}c@{}} 51.45\\ 
(\textcolor{blue}{2.04})
\end{tabular}
& \multirow{1}{*}{\textcolor{blue}{\textbf{1.77}}}
\\ 
\bottomrule[1.5pt]

\end{tabular}
}
\end{table}

%%%%%%%%%%%%%%%%%%%%%%%%%%%%%%%%%%%%%%%%%%%%%%%%%%%%%%%
%%%%%%%%%%%%%%% APPENDIX B
%%%%%%%%%%%%%%%%%%%%%%%%%%%%%%%%%%%%%%%%%%%%%%%%%%%%%%%%

\section{Experiment Details}\label{appendix:experiment-details}

\subsection{Training}
In class-wise unlearning, both the original and retrained models are trained on the CIFAR-100 and CIFAR-20 datasets for 100 epochs using ResNet-18, with a warm-up period of 2 epochs. The optimizer is SGD, and a step learning rate schedule is applied, where the learning rate decays by a factor of 0.1 at the 60th and 80th epochs. The initial learning rate is set to 0.1, weight decay to 0.0005, and the batch size to 128. Data augmentation techniques include RandomCrop(32, padding=4) and RandomHorizontalFlip(0.5).

In sample-wise unlearning, we train VGG16-BN on CIFAR-10, ResNet-18 on CIFAR-100, and ResNet-34 on TinyImageNet-200. The experimental setup is largely consistent with class-wise unlearning, with two main differences: ResNet-34 is trained on TinyImageNet-200 with a batch size of 256, and RandomCrop(64, padding=8) replaces RandomCrop(32, padding=4) in the data augmentation settings. 
For ViT model, we finetune a TinyViT~\cite{tinyvit} model pretrained by TIMM~\cite{timm} with AdamW optimizer for 100 epochs. The initial learning rate is set to $5e-4$ with a cosine scheduler, weight decay to 0.0005, and the batch size to 256. Images are first resized to $224\times 224$ and augmented by RandomCrop(224, padding=8) and RandomHorizontalFlip(0.5).

\textbf{Computer resources.} 
All experiments are conducted within a Docker container using Python 3.8.8 and PyTorch 1.12.0+cu113. The hardware includes an RTX 4090 GPU, a 16-core CPU, and 40 GB of memory. 

\subsection{Details of Baselines}

For each comparison method, we use Bayes Hyperparameter Search from Optuna~\cite{DBLP:conf/kdd/AkibaSYOK19} for efficient search. Unless otherwise specified, for optimization-based methods, including NatMU, we  use the AdamW optimizer with the batch size (bs) same as pretraining and search for the weight decay within the range of $[0, 0.001]$. The learning rate decays using  the cosine learning rate scheduler~\cite{cosine-lr} and training epoch is set to 5 (10 for ViT experiments). 
We compare NatMU with 8 baseline methods:

1. \textbf{SSD}~\cite{ssd} is an optimization-free unlearning method. It adopts the gradients on different data to represent its importance to specific data. It calculates the parameter gradients of forgetting data and remaining data respectively, and suppress the parameters which are more important for forgetting data. We search the hyperparameters $\alpha \in [0.1, 1000]$ and $\lambda \in [0.1, 10]$. To obtain a better gradient approximation, we use a bs of 500 for forgetting data (in class-wise unlearning and 1\% sample-wise unlearning on CIFAR, the forget set only contains 500 samples) and pretraining-bs*8 for remaining data.

2. \textbf{NegGrad+}~\cite{scrub} is a relabeling-free unlearning method proposed to serve as a strong baseline in SCRUB~\cite{scrub}. It combines gradient ascent on the forgetting data and gradient decent on the remaining data. For class-wise unlearning, we use a SGD optimizer and search the learning rate in $[0.005, 0.05]$; and for sample-wise unlearning, AdamW optimizer with learning rate in $[0.0001, 0.002]$. We use a coefficient to determine the weight of gradient ascent loss, and search it in $[0.1, 10]$.

3. \textbf{SCRUB}~\cite{scrub} is also a relabeling-free method which forces the unlearned model obey the teacher model selectively. On forgetting data, the unlearned model should disobey the teacher; on remaining data, the unlearned model should obey the teacher, and classify these samples into the original category. It also uses an alternating optimization technique for better optimization of the two objectives. We search learning rate in $[5e-5, 1e-3]$, the epoch of DO-MAX-EPOCH in \{1, 2, 3, 4, 5\}, $\alpha \in [0.01, 3]$ and $\gamma \in [0, 2]$. For TinyImageNet we use a forget set batch size of 400 for better performance and for other settings, the forget set bs and retain set bs are set to the pretraining bs. 

4. \textbf{Amnesiac}~\cite{amnesiac} relabels the forgetting sample with random different labels and minimizes the cross-entropy loss on the relabeled data and remaining data. We search the learning rate in $[5e-5, 1e-3]$. We initialize the linear layer weight in full-class unlearning for efficiency. 

5. \textbf{BadTeacher}~\cite{badteacher} relabels the forgetting sample with a randomly initialized model and the remaining sample with the pretrained model. Then it optimizes the unlearned model using a KL loss function with a temperature coefficient. 
We search the learning rate in $[1e-5, 3e-4]$ and temperature coefficient in $[0.1, 10]$. 

6. \textbf{SalUn}~\cite{salun} improves Amnesiac by only optimizing important parameters. It calculates the gradients on forgetting data and optimizes the parameters with larger gradient values. We search the learning rate in $[5e-5, 1e-3]$ and the mask ratio in $[0.01, 0.99]$. When calculating the gradient on forgetting data to select the important parameters, we use a forget set bs of 500. 

7. \textbf{$\delta\text{-targeted}^\lambda$}~\cite{targeted} improves Amnesiac from three aspects: 1) it relabels each forgetting samples with its top non-ground-truth prediction class. 2) It adds noise when relabeling the samples, controlled by the hyperparameter $\delta$. 3) It re-scales the cross entropy loss function used in unlearning, controlled by $\lambda$.  Following~\cite{targeted}, $\delta$ is set to zero as default. For class-wise unlearning, we found $\lambda=1$ works well and search the learning rate in $[4e-4, 1e-2]$ with SGD optimizer. For sample-wise unlearning, $\lambda=3$ performs better, and we search the learning rate in $[5e-5, 1e-3]$ with AdamW optimizer. 

8.~\textbf{AMUN}~\cite{amun} generates an adversarial sample $(\boldsymbol{x}^a, y^a)$ from a forgetting $(\boldsymbol{x^f}, y^f)$, with $\|\boldsymbol{x}^a-\boldsymbol{x}^f\| \leq \epsilon$ but $y^a \neq y^f$. Then it finetunes the model on unmodified remaining data, forgetting data and the generated adversarial data. Its basic idea is to shift the decision boundary of forgetting samples, since the adversarial samples are similar to forgetting ones at the input level, but have totally different labels. However, as AMUN still finetunes the model on original forgetting data, it may fail to achieve a low forgetting accuracy, especially in class-wise unlearning where FA is zero. Thus, we add a coefficient $\delta$ to control the weight of loss on forgetting data and search $\delta$ in $[0,1]$. Note that $\delta=1$ refers to the vanilla AMUN. We search the learning rate in $[5e-5, 1e-3]$ following the paper.

\subsection{Details of NatMU}

Following the impair-repair idea in \cite{impair}, we use a high learning rate in the initial stages of training to promote model forgetting.
In full-class and sub-class unlearning, we initialize the last linear layer before unlearning, as the implementation in Amnesiac. We also adopt data augmentation for remaining data to improve the generalization ability. Code is available at \url{https://github.com/ZhengbaoHe/NatMU}.

\textbf{Hyperparameters}. we search the learning rate in $[0.0002, 0.002]$, $\delta \in [-0.2, 0.1]$ and weight decay in $[0.0001,0.002]$. The hyperparameters used in different tasks are shown in Table~\ref{tab:hyperparameters}. We initialize the linear layer weight for full-class unlearning. 
Compared to other relabeling-based methods, NatMU only introduces one more hyperparameter $\delta$. Basically, for simpler classification tasks, we use a smaller $\delta$.

\begin{table}[h]
    \centering
    \caption{Hyperparameters of NatMU in different unlearning tasks.}
    \label{tab:hyperparameters}
    \begin{tabular}{cccc}
    \toprule 
        Unlearning Task & lr & wd & $\delta$ \\
    \midrule
        Full-class & 0.00028 & 0 & -0.053 \\
        Sub-class & 0.00089 & 0.0004 & -0.053 \\
        Sample-wise (CIFAR-10) & 0.00105 & 0.002 & -0.19 \\
        Sample-wise (CIFAR-100) & 0.00107 & 0.0005 & -0.031 \\
        Sample-wise (TinyImageNet, RN34) & 0.00083 & 0 & 0.045 \\
        Sample-wise (TinyImageNet, TinyViT) & 0.002 & 0.05 & 0.15  \\
        
    \bottomrule
    \end{tabular}
    
\end{table}

\begin{figure}[ht]
    \centering
  
    \begin{subfigure}[b]{0.22\textwidth}
        \centering
        \includegraphics[width=\textwidth]{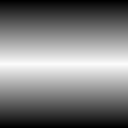}
        \caption{$m_1$}
        \label{fig:sub m_1}
    \end{subfigure}
    \hfill
    \begin{subfigure}[b]{0.22\textwidth}
        \centering
        \includegraphics[width=\textwidth]{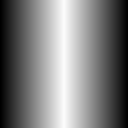}
        \caption{$m_2$}
        \label{fig:sub m_2}
    \end{subfigure}
    \hfill
    \begin{subfigure}[b]{0.22\textwidth}
        \centering
        \includegraphics[width=\textwidth]{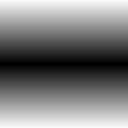}
        \caption{$m_3$}
        \label{fig:sub m_3}
    \end{subfigure}
    \hfill
    \begin{subfigure}[b]{0.22\textwidth}
        \centering
        \includegraphics[width=\textwidth]{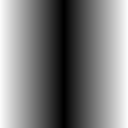}
        \caption{$m_4$}
        \label{fig:sub m_4}
    \end{subfigure}
  
    \caption{Visualization of weighting vectors. }
    \label{fig:mask-vectors}
\end{figure}

\textbf{Weighting mask visualization}. 
Fig.~\ref{fig:mask-vectors} visualizes the 4 weighting vectors to facilitate understanding. We can see that, a weighting mask captures the entire pixels of the forgetting samples.
Moreover, different weighting masks focus on different parts of the images. We also have tried other weighting masks, such constant weighting mask in  \mixup~\cite{mixup} or area weighting mask in CutMix~\cite{cutmix}. Experiments show that this used implementation has a better robustness across different unlearning scenarios. 

\textbf{Algorithm.} The algorithm for generating the fine-tuning dataset for NatMU is provided in Algorithm~\ref{algo:gererate dataset}. After generating the dataset, we can fine-tine the original model with it to obtain the unlearned model.
\begin{algorithm}[H]
    \caption{Construct Fine-tuning Dataset}\label{algo:gererate dataset}
    \textbf{Input:} remaining set $D^r$, forgetting set $D^f$,  original model $f_{\theta_o}$, weighting vectors $\{\boldsymbol{m}_j \}_{j=1}^n$, scaling factor $\delta$.
    
    \textbf{Output:} fine-tuning dataset $D_{\mathrm{Nat}}$
    \begin{algorithmic}[1]
    
    \For{$j \gets 1$ \textbf{to} $n$}
    \State Generate $\boldsymbol{m}_j^\mathrm{scaled}$ according Eq.~\ref{eq:scale}.
    \EndFor
    
    \State $D_{\mathrm{Nat}} \gets D^r$

    \For{$(\boldsymbol{x}_i, y_i) \in D^f$}
        \State Generate $\mathcal{D}^f_i$ according to Eq.~\ref{eq:generating}.
        \State $D_{\mathrm{Nat}}  \gets D_{\mathrm{Nat}} \cup \mathcal{D}^f_i$
    \EndFor
    
    \State \textbf{return} $D_{\mathrm{Nat}}$
    
    \end{algorithmic}
\end{algorithm}

\subsection{Selecting the Difficult-to-Learn Samples}

To identify the difficult-to-learn samples, we track whether each sample is correctly classified at the end of each training epoch, and record the number of epochs in which the sample is correctly classified (denoted as $\#\text{num}$). Then, we sort all training samples in ascending order based on $\#\text{num}$ and select the top $r\%$ samples as the forgetting samples, where $r\%$ represents the forgetting ratio.

\end{document}